\documentclass{article}

\usepackage{arxiv}

\usepackage{graphicx}
\usepackage[T1]{fontenc}
\usepackage{lmodern}

\usepackage{hyperref}
\usepackage{url}            
\usepackage{booktabs}       
\usepackage{amsfonts}       
\usepackage{nicefrac}       
\usepackage{microtype}      
\usepackage{xcolor}         

\usepackage{latexsym,amsfonts,amssymb} 
\usepackage{amsmath}
\usepackage{amsthm}
\usepackage{url}
\usepackage{subfigure}

\usepackage{algorithm}
\usepackage[noend]{algpseudocode}
\usepackage{stmaryrd}
\usepackage{bbm}
\usepackage{thmtools}
\usepackage{thm-restate}
\usepackage[T1]{fontenc}
\usepackage{enumitem}

\usepackage{xcolor}

\newtheorem{thm}{Theorem}
\newtheorem{example}{Example}
\newtheorem{remark}{Remark}
\newtheorem{theorem}{Theorem}
\newtheorem{lemma}[thm]{Lemma}
\newtheorem{proposition}[thm]{Proposition}
\newtheorem{corollary}[thm]{Corollary}
\newtheorem{definition}[thm]{Definition}

\newcommand{\exptd}{\mathbb{E}}
 
\newcommand{\argmax}{\operatornamewithlimits{argmax}}

\newcommand\R{\mathbb{R}}   
\newcommand\N{\mathbb{N}}   

\newcommand{\IND}[1]{\mathbbm{1}_{ {#1}  }}

\newcommand{\up}{\mathrm{up}}
\newcommand{\low}{\mathrm{down}}

\newcommand{\tautemp}{{\tau}}

\title{Multi-Armed Bandits with Censored\\ Consumption of Resources}

\renewcommand{\shorttitle}{Multi-Armed Bandits with Censored Consumption of Resources}

\author{ Viktor Bengs \\
	Institute of Informatics, LMU Munich, Germany  \\  
	viktor.bengs@lmu.de 
	\And  Eyke H\"ullermeier \\
	Institute of Informatics \& 
	Munich Center for Machine Learning, Munich, Germany\\
	eyke@lmu.de
	}

\hypersetup{
	pdftitle={Multi-Armed Bandits with Censored Consumption of Resources},
	pdfsubject={q-cs.LG, q-stat.ML},
	pdfauthor={Viktor Bengs, Eyke Huellermeier},
	pdfkeywords={Algorithm Configuration, Algorithm Selection, Bivariate Distributions, Censored Feedback, Exploration, Exploitation, UCB, Thompson Sampling},
}

\begin{document}
\maketitle              
\begin{abstract}
We consider a resource-aware variant of the classical multi-armed bandit problem: In each round, the learner selects an arm and determines a resource limit. It then observes a corresponding (random) reward, provided the (random) amount of consumed resources remains below the limit. Otherwise, the observation is censored, i.e., no reward is obtained. 
For this problem setting, we introduce a measure of regret, which incorporates both the actual amount of consumed resources of each learning round and the optimality of realizable rewards as well as the risk of exceeding the allocated resource limit.
Thus, to minimize regret, the learner needs to set a resource limit and choose an arm in such a way that the  chance to realize a high reward within the predefined resource limit is high, while the resource limit itself should be kept as low as possible.
We propose a UCB-inspired online learning algorithm, which we analyze theoretically in terms of its regret upper bound. 
In a simulation study, we show that our learning algorithm outperforms straightforward extensions of standard multi-armed bandit algorithms.
\end{abstract}

\section{Introduction} \label{sec:introdution}

\emph{Multi-armed bandit} (MAB) problems constitute an important branch of machine learning research. 
Their popularity largely stems from an appealing combination of theoretical tractability and practical relevance. In fact, MABs cover a wide range of real-world sequential decision problems, where an agent takes actions (metaphorically considered as ``pulling arms'') in order to optimize a specific evaluation criterion, simultaneously exploring the set of actions available and exploiting the feedback resulting from the actions taken. The latter typically comes in the form of (numerical) rewards, generated by the pulled arm according to an underlying probability distribution. 

In spite of its versatility, the complexity of real-world problems or the availability of additional side information may suggest further extensions of the basic MAB setting.  
Indeed, several variants of the basic setting have been developed in order to model specific real-world problem scenarios more appropriately, including $\mathcal{X}$-armed \cite{bubeck2011x}, linear  \cite{auer2002using,abe2003reinforcement}, dueling  \cite{yue2009interactively},  combinatorial  \cite{cesa2012combinatorial}, or threshold bandits \cite{abernethy2016threshold}, just to name a few\,---\,for a more detailed overview we refer to \cite{lattimore2020bandit}.
In this paper, we introduce yet another extension of the basic MAB problem, again motivated by practical considerations. More specifically, we consider applications in which the execution of an action requires resources, and will not be successful unless enough resources are provided. 
Thus, instead of observing a (noisy) reward in every round, the reward is only generated if the resources consumed by the pulled arm remain below a resource limit specified by the learner.
Consequently, the learner needs to make two choices in every round of the decision process: the arm to be pulled and the resources allocated to that arm. Since we assume that resources are costly, the value of an outcome produced as a result decreases with the resources consumed.
Additionally, the learner might be penalized for allocating a resource limit such that no reward is generated.

Our setting is largely (though not exclusively) motivated by the problem of algorithm selection \cite{kerschke2019automated}, which has gained increasing attention in the recent past. Here, the arms are algorithms that can be run on a specific problem instance, for example different solvers that can be applied to an optimization problem or different machine learning algorithms that can be run on a data set. Given a problem instance, the task of the learner is to choose an algorithm that appears most appropriate, and the reward depends on the quality of the result achieved, typically measured in terms of a performance metric (e.g., the generalization performance of a model trained on a data set). Even if this metric is of major concern, one should not overlook that different algorithms have different runtimes or different memory consumptions. For example, training a deep neural network is way more costly than training a decision tree. Depending on the application context, these resource requirements might be important, too. In automated machine learning, for example, many algorithms\,---\,or even complete ``machine learning pipelines''\,---\,are tried, one by one, before a final decision must be made after a certain cutoff time \cite{hutter2019automated}. The more costly the algorithms are, the less can be tried.

In cases like those just discussed, the learner needs to find the right balance between two competing targets: an as high as possible reward and an as low as possible consumption of resources.
As a consequence, the learner might be willing to sacrifice reward if it helps to keep the overall consumption of resources low, or the other way around, be willing to allocate more resources if this significantly increases the chance of realizing a high reward. 
In light of this, the underlying correlations between the reward distribution of an arm and the distribution of resource consumption need to be learned, in order to ascertain to which degree the target values conform to each other.
Moreover, the learner needs to cope with possibly censored feedback, in case the chosen arm did not return any reward under the allocated resource limit.

In this paper, we model sequential decision problems of the above kind formally and introduce a reasonable measure of regret (or loss) capable of capturing the additional trade-off between realizable reward and consumption of resources as well as the risk of overexciting the allocated resources (Section~\ref{sec_problem_intro}).
In Section~\ref{sec_algorithmic_solution},  we fist study this problem under the restriction that the possible resource limits can only be chosen within a fixed finite set and describe how the problem can be na\"ively tackled by a standard MAB learner.
Next, we define a suitable estimate for the target value in the considered problem, which extracts all available learning information from the possibly censored type of feedback. 
With this, we propose a UCB-inspired bandit learning algorithm, the Resource-censored Upper Confidence Bound $(\texttt{RCUCB})$ algorithm, for which we derive an upper bound on its cumulative regret.
Our result reveals in particular why $\texttt{RCUCB}$ is in general superior to straightforward modifications of well-established standard multi-armed bandit learning algorithms for the considered type of bandit problem.
By modifying the $\texttt{RCUCB}$ algorithm in a suitable way, leading to the $\texttt{z-RCUCB}$ algorithm, we show in Section \ref{sec_algorithmic_solution_part2} how one can deal with the case where the possible resource limits can be chosen as any value within a left-open interval.
Further, we experimentally confirm $\texttt{RCUCB}$'s superiority to the straightforward standard bandit reduction approaches in an experimental study (Section~\ref{sec_exp}).
Finally, we discuss other bandit problems related to ours (Section~\ref{sec_related_work}), prior to concluding the paper (Section~\ref{sec_conclusion}).
For the sake of convenience, we provide a list of symbols used in the paper in the supplementary material, where we also provide all proofs of the theoretical results.

\section{The Bandit Problem} \label{sec_problem_intro}
In the following, we specify the bandit problem described in Section~\ref{sec:introdution} in a formal way {  and motivate it using the example of algorithm selection, where the role of an arm is played by a concrete configuration of a learning algorithm, e.g., a neural network with a specific parametrization (network structure, weights, etc.).}
In particular, we provide a working example where we consider the scenario of a company which provides an on-the-fly machine learning service, where the customers can submit a learning task in form of a data set and some performance metric, for which a suitable machine learning model is returned. 
The payment agreement between the customer and the company provides for the customer to pay the company an amount of money depending on the performance of the returned machine learning model, while the company has a fast-track-promise and will pay the client some amount of money if a suitable machine learning model cannot be provided within a certain time.

\paragraph{Learning Process.} The learning process proceeds over $T$ many rounds, where $T \in \mathbb{N}$ is not necessarily known beforehand. For each round $t\in [T] := \{1,2,\ldots,T\}$, there is a maximal resource limit $\tau_{\mathrm{max}} \in \mathbb{R}_+,$ which is fixed and known beforehand. 
{  In (online) algorithm selection, for instance, each round corresponds to a time step, in which an incoming task specified by a data set comprising of a training and test data set and some performance metric needs to be solved. Each algorithm consumes resources for a given task, e.g., the energy consumption or simply the time for the training phase. Due to external constraints, the consumed resources should not exceed some specific limit, e.g., a maximal energy consumption level or a time limit.}

\paragraph{Arms.} We assume a finite number of $n$ arms, where $n \in \mathbb N$. 
For sake of convenience, we identify the arms by the set $ [n]= \{1,\ldots,n\}.$ 
Each arm $i \in [n]$ is associated with two distributions: 
\begin{itemize} 
	\item a \emph{reward distribution} $P_i^{(r)}$ with support\footnote{It is straightforward to extend our algorithmic solution to $\sigma$-sub-Gaussian reward distributions.} in $[0,1];$  
	\item a \emph{consumption of resources distribution} $P_i^{(c)}$ with support in $\mathbb{R}_+$ characterizedby the cumulative distribution function  $F_i^{(c)}.$ 
\end{itemize}
The joint distribution of an arm's reward and resource consumption is denoted by $P_i^{(r,c)}$ and is \emph{not} necessarily the product of $P_i^{(r)}$ and $P_i^{(c)}$, i.e., an arm's reward and consumption of resources are not assumed to be independent. 
In particular, this allows for stochastic dependencies between rewards and resource consumptions.

For instance, running a specific configuration of a learning algorithm on an incoming task generates a reward, e.g., the accuracy on the test data or a monetary conversion thereof, and consumes resources, e.g., the energy consumption or simply the time for the training phase. If the data set is generated by some unknown random mechanism (a random training/test split) both the reward and the resource consumption are random as well. Moreover, both observations are likely to be correlated, because the more complex the configuration of a learning algorithm is, e.g., a neural network with a large number of neurons and weights, the higher its accuracy (reward) in general, but also the higher its resource consumption due to its high complexity.

	\begin{example} \label{example_1}
		Coming back to the working example, let us assume for sake of simplicity that the company has three possible machine learning models available each representing an arm, so that $n=3.$
		Suppose for simplicity reasons that the payoff for the returned model is 1:1 to its performance,  so that the reward distribution is equivalent to the general performance distribution of a model on possible learning tasks.
		The only resources consumed is the wall-clock time for running the model. 
		In Figure \ref{fig:example_arm_distr} the reward, consumption of resources and their joint distribution of the three models are illustrated.
		
		\begin{figure}
			\centering
			\subfigure
			{
				\includegraphics[width=0.8\linewidth]{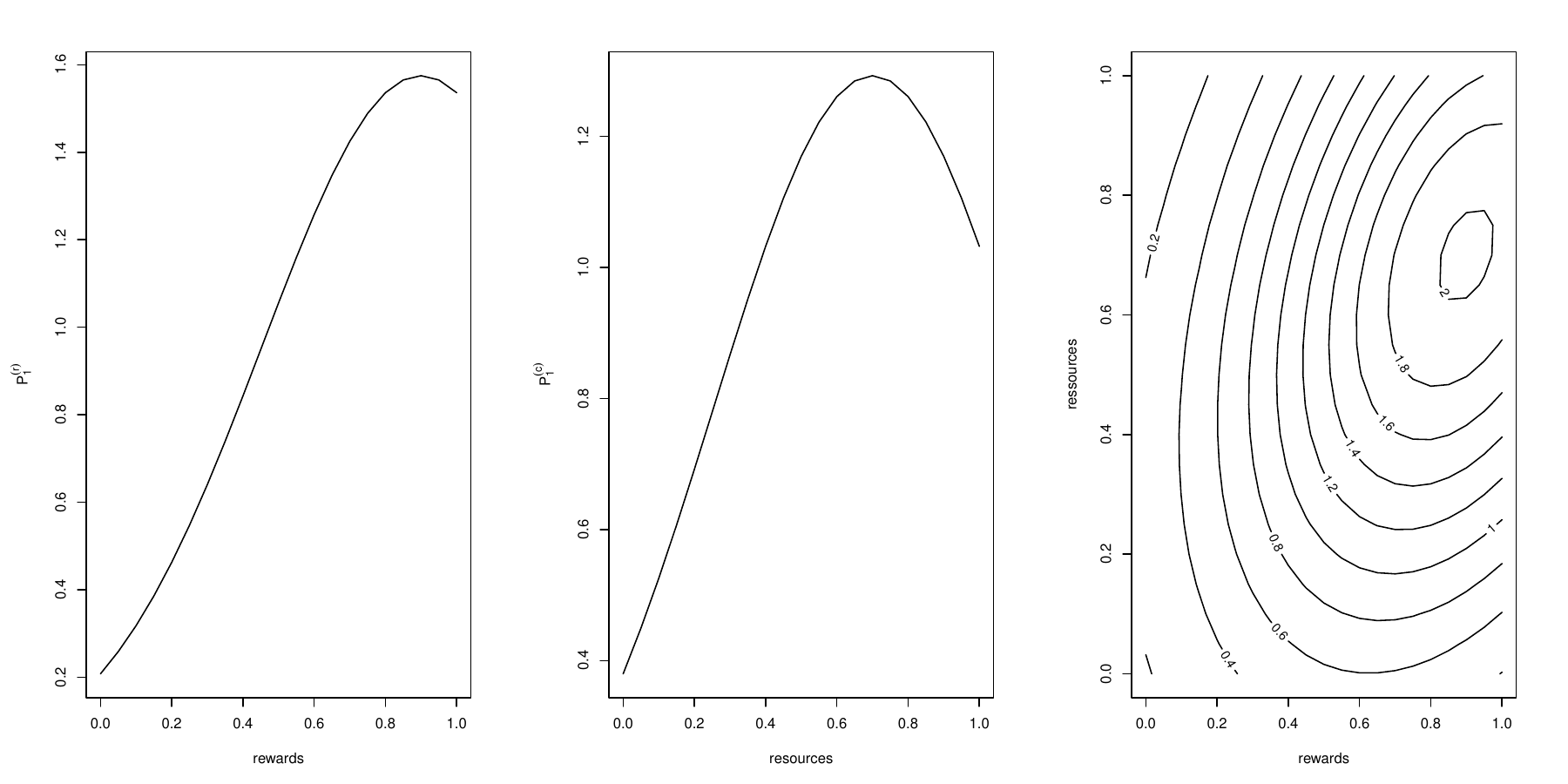}}
			\subfigure
			{  
				\includegraphics[width=0.8\linewidth]{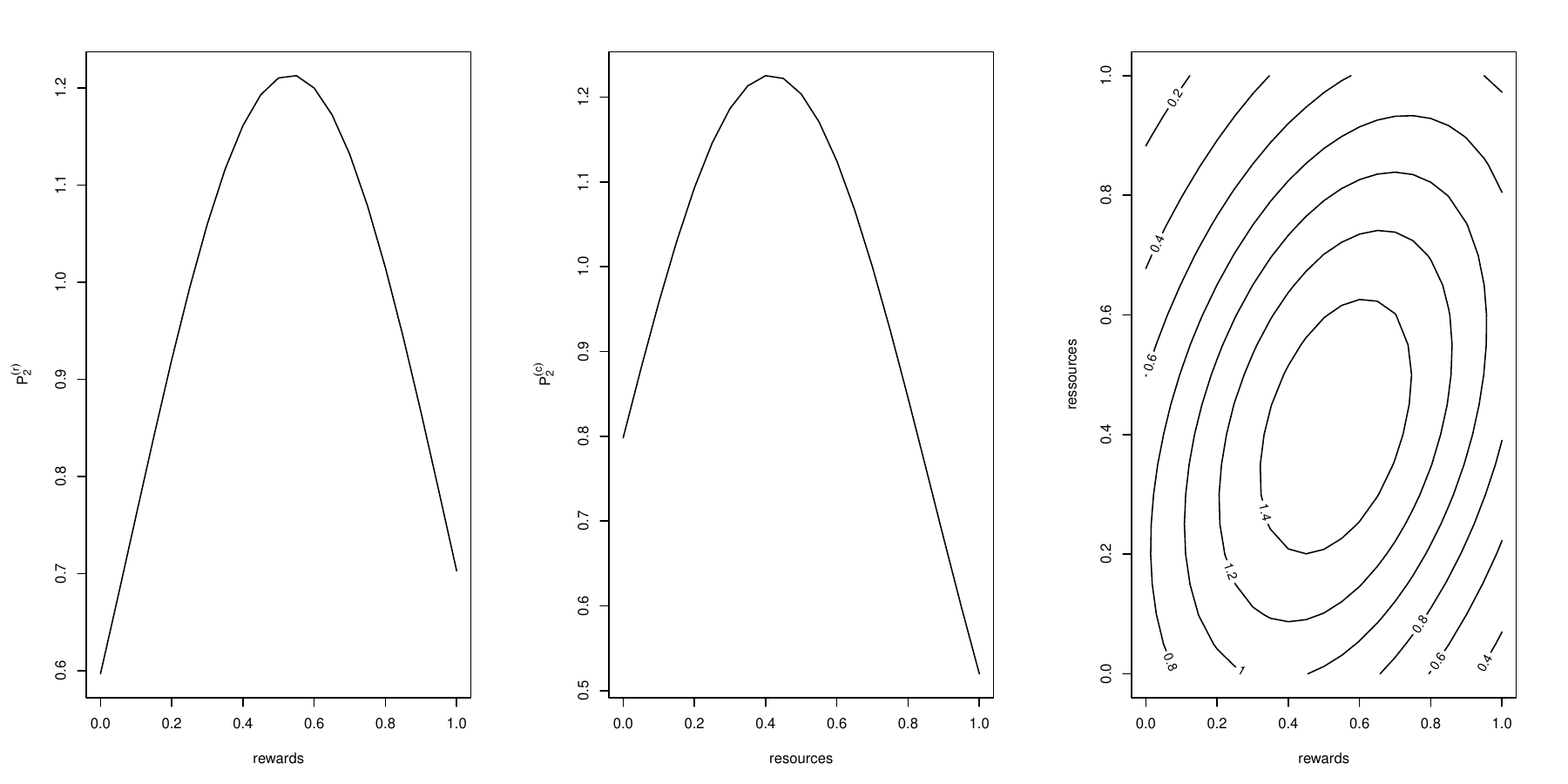}}
			\subfigure
			{  
				\includegraphics[width=0.8\linewidth]{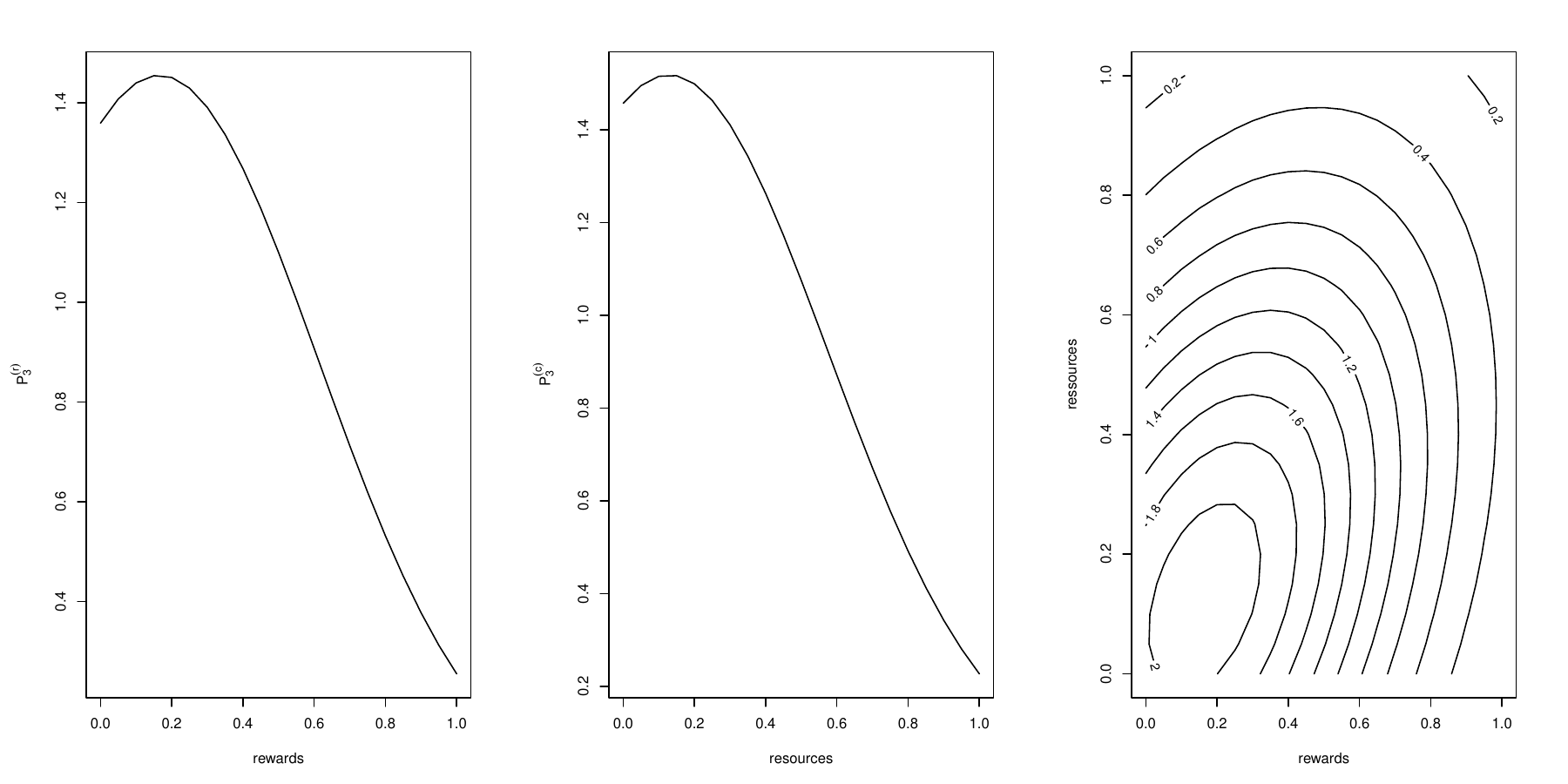}
			}
			\caption{Exemplary reward (left column), consumption of resource (middle column) and joint distribution for three arms ($i$th arm corresponds to $i$th row for $i=1,2,3$) representing the machine learning models in Example \ref{example_1}.}
			\label{fig:example_arm_distr}
		\end{figure}
		We see that the first model yields high rewards (general performance), but also has a high consumption of resources (running time), e.g.\ a very complex model such as a large deep neural network.
		The second model has mediocre rewards (general performance), while consuming fewer resources as the first one, e.g.\ a random forest.
		Finally, the third model has both low rewards and low resource use, e.g.\ a simple linear regression model.
		All three models show a positive correlation between rewards and resource consumption, which makes sense in this case due to the complexity of the models, because the higher the resource consumption, the higher the reward.
	\end{example}

\paragraph{Learner.} A learner (or bandit algorithm) in this setting is a possibly non-deterministic procedure, which, in each round $t\in [T]$, chooses an arm $I_t  \in [n]$ and a resource limit $\tau_t \in \mathcal M$ depending on the history of previously chosen arms, resource limits, and observed feedback (specified below).
Here,  $\mathcal M$ is a subset of $(0,\tau_{\mathrm{max}}]$  specifying the admissible resource range. 
In the usual algorithm selection setting, the learner is essentially a mechanism deciding on which algorithm to choose for the incoming task based on the history of observations seen so far. 
In our setting, on the other hand, the learner has a more challenging decision to make, as it needs to decide on the most suitable algorithm/resource-limit pair for the given task.

\paragraph{Feedback.} The feedback observed by the learner in round $t$, if $I_t$ is the chosen arm and $\tau_t$ the resource limit, is 
\begin{align}\label{def_feedback}
X_{I_t,t} = \begin{cases}
(R_{I_t,t}, C_{I_t,t}), & \mbox{if } C_{I_t,t}\leq \tau_t \\
(\varnothing , (\tau_t,\infty]  ), & \mbox{else}
\end{cases} \, ,
\end{align} 
where $(R_{I_t,t}, C_{I_t,t}) \sim P_{I_t}^{(r,c)}.$
In words, if the (noisy) consumption of resources $C_{I_t,t}$ of the chosen arm $I_t$ is within the scheduled resource limit $\tau_t$ of the learner, the corresponding reward of the arm $R_{I_t,t}$ is observed (realized), and the corresponding consumption of resources $C_{I_t,t}$ as well. 
Otherwise, neither the consumption of resources $C_{I_t,t}$ nor the corresponding arm reward $R_{I_t,t}$ is observed (or realized), which we represent by the left-open interval $(\tau_t,\infty]$ resp.\ $\varnothing$\footnote{Here, $\varnothing$ is interpreted as a symbol for a dummy variable indicating that no reward information was received.}.
Here, it is worth noting that although the learner does not observe direct feedback in the latter case, it still observes a valuable information in the form of censored feedback, namely that the consumption of resources $C_{I_t,t}$ exceeded the resource limit $\tau_t,$ i.e., the latter is an element in $(\tau_t,\infty].$ 

In our scenario, we assume that the observed feedback $X_{I_t,t}$ in round $t$ is independent of the past given $I_t$ and $\tau_t.$
This assumption is reasonable from a practical point of view, as, for instance, the run of one specific configuration of a learning algorithm on a randomly split training set is independent of the run of the same learning algorithm configuration on another randomly split training set.

\paragraph{{Profit and loss account.}}
The task of the learner is to select, in each round $t$, an arm as well as a resource limit such that in expectation an as high as possible reward can be realized within the specified resource limit, while simultaneously keeping the expected consumption of resources of the round as small as possible.
To this end, we assume that the learner is provided with two monotonic increasing functions, namely
\begin{itemize} 
\item  a \emph{cost function} $c:\R_+ \to [0,1],$ which specifies the cost generated by the consumed resources;
\item a \emph{penalty function} $\lambda: \mathcal{M} \to \R_+$ which specifies the penalty for exceeding the allocated resources.
\end{itemize}
The cost function is in the first place mapping the consumption of resources on the same scale as the rewards in order to make them comparable\footnote{In particular, the cost function is in fact a mapping $c:\R_+ \to \mathrm{supp}(P^{(r)})$, where $\mathrm{supp}(P^{(r)})$ is the common support of the reward distributions, which is assumed to be $[0,1]$ for sake of simplicity.}
The penalty function maps the allocated resources (in case of exceeding them) on the same scale as the rewards as well, but in addition gives the learner an incentive to choose resource limits smaller than the maximal possible resource limit. 
Leveraging the prevalent way of profit and loss accounting in economics, we define for each possible decision pair of a learner $(i,\tautemp) \in [n] \times \mathcal{M}$ its \emph{penalized expected gain} $\nu_{i,\tautemp}$ via 
\begin{align} \label{def_nu_i_tau}
\nu_{i,\tautemp} :=  \mathbb E_{P_i^{(r,c)}} \Big( \big( X_i^{(r)} - c(X_i^{(c)}) \big) \cdot \IND{ \{X_i^{(c)} \leq  \tau\}  }  \Big)   -  \mathbb E_{P_i^{(c)}} \Big( \lambda(\tau) \IND{  \{X_i^{(c)}>\tau\} } \Big) ,
%
\end{align}
where $ (X_i^{(r)} , X_i^{(c)} )^\top \sim  P_i^{(r,c)}.$
In words, the quality of a decision pair $(i,\tautemp) \in [n] \times \mathcal{M}$ is measured by means of its expected gain (first term in \eqref{def_nu_i_tau}), which counts the expected profit against the expected loss, while taking an expected ``fine'' or penalty for possibly exceeding the allocated resources into account.

With this, the task of the learner is to select in each learning round an arm/resource-limit pair having the maximal penalized expected gain, i.e.,
\begin{align} \label{def_max_problem}
(i^*,\tau^*) \in \argmax\limits_{(i, \tau ) \in [n]\times \mathcal M} \nu_{i,\tautemp}.  
\end{align}
The ``negative part'' or the expected cost part in \eqref{def_nu_i_tau}, i.e., 
$$\mathbb E_{P_i^{(c)}} \Big( c(X_i^{(c)})  \cdot \IND{ \{X_i^{(c)} \leq  \tau\}  }  + \lambda(\tau) \IND{  \{X_i^{(c)}>\tau\} } \Big) $$ 
allows one to recover common performance metrics considered for algorithm selection problems \cite{kerschke2019automated}, such as the so-called \emph{penalized average running times} if the consumption of resources correspond to runtimes of algorithms. 
For instance, the expected value of the popular PAR10 score corresponds to the choice of $c(x)=x$ and $\lambda(x)=10x$ \footnote{Here, we assume for sake of simplicity that the resource consumption (runtime) is already on the proper scale for comparing it with the rewards.}.

In general, one can note that depending on the concrete form of the functions, the learner can either be urged to focus on arms with a small consumption of resources, but possibly slightly smaller expected rewards ($c$ and/or $\lambda$ grow quickly), or to almost exhaust the available resources in a single round in order to realize the presumably high rewards of arms with high consumption of resources ($c$ and/or $\lambda$ grow slowly).

{ \color{black}  
	\begin{example}
		Recall that in our working example, the company has a fast-track-promise and pays a compensation, say $L,$ to the customer if no suitable model can be provided within a specific amount of time $\tilde \tau$.
		Suppose that the wall-clock time of running the model (consumption of resources) only generates energy costs, which correspond to one tenth of the wall-clock time, i.e., $c(x)=x/10.$
		Thus, the penalty function is  $\lambda(x)=c(x) \IND{\{x\leq \tilde \tau\} }  + L x \IND{\{x> \tilde \tau\}},$ where the first term accounts for the (energy) costs of running the model.
		For the three available models with distributions as in Example \ref{example_1} we illustrate in Figure \ref{fig:nuplot} the penalized expected gain for two cases, a strict fast-track-promise case with $L=10$ and $\tilde \tau = 0.5,$ and a soft fast-track-promise case with $L=5$ and $\tilde \tau = 0.99.$ 
		
		\begin{figure}[ht!]
			\centering
			\includegraphics[width=0.9\linewidth]{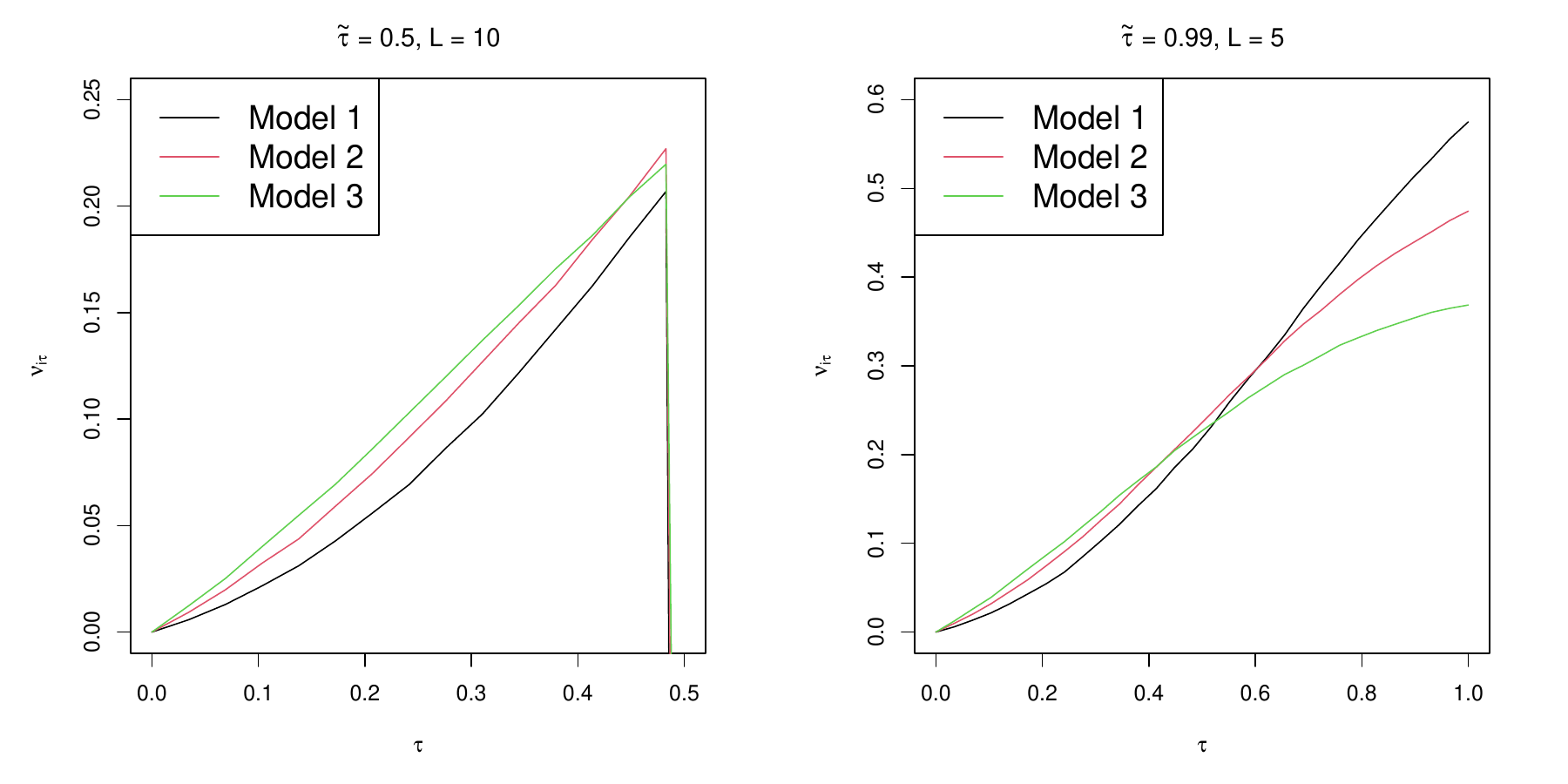}
			\caption{Penalized expected gain $\nu_{i,\tautemp}$ for cost function $c(x)=x/10$ and penalty function $\lambda(x)=c(x) \IND{\{x\leq \tilde \tau\} }  + L x \IND{\{x> \tilde \tau\}}$ for $\tilde \tau = 0.5, L=10$ (left) and $\tilde \tau = 0.99, L=5$ (right) for the three arms (machine learning models) of Example \ref{example_1}.}
			\label{fig:nuplot}
		\end{figure}
		
		We see that in the case of a strict fast-track-promise (left plot) the simple model is most lucrative for most choices of $\tau$ and being overtaken by the medium-complex model only near the $\tilde \tau$, while the complex model is completely unsuitable\footnote{We leave out the cases for which $\tau>0.5$ as $\nu_{i,\tautemp}$ is negative in this case.}. 
		For the soft fast-track-promise case (right plot), the most complex model attains the highest penalized expected gain, for which a larger choice of $\tau$ is necessary due to its high chance of returning a censored feedback by exceeding the allocated resources.
	\end{example}
}

\paragraph{{Quality of a learner.}} 
Having defined the quality of a decision pair, one can compare the decision made by a learner with the optimal decision to obtain a natural measure for the (sub-)optimality of the learner's decision in each round by means of a notion of regret.
Indeed, if the learner chooses the arm $I_t$ and the resource limit $\tau_t$ in round $t,$  define the \emph{instantaneous (pseudo-)  regret} as the difference between the optimal penalized expected gain and the penalized expected gain of the chosen pair, i.e., $r_t :=  	\nu^*  - \nu_{I_t,\tau_t},$
where $\nu^*$ is the maximum value in \eqref{def_max_problem}.
Hence, the cumulative (pseudo-) regret is given by
\begin{align}\label{regret_def} 
\mathcal R_T \, := \, \sum\nolimits_{t=1}^{T} r_t \, = \, \nu^*  \, T  - \sum\nolimits_{t=1}^{T} \, \mathbb E \,	\nu_{I_t,\tau_t},
\end{align}
where $(I_t,\tau_t)_{t=1}^T$  are the actions chosen by the learner during the $T$ rounds.
Note that, in general, it is possible to have multiple optimal pairs $(i,\tautemp) \in [n] \times \mathcal{M}$, such that the instantaneous regret $r_t$ vanishes.
However, without loss of generality, we subsequently assume that there is only one unique pair $(i^*,\tau^*)$ such that 
$\nu^* = \nu_{i^*,\tau^*}$ holds, as having multiple optimal pairs only makes the learning problem easier.
\begin{remark}
The considered bandit problem can recover the standard MAB problem by assuming that $\mathcal{M}=\{0\}$ and each arm's consumption of resources distribution is the Dirac measure on $\mathcal{M},$ while the cost and the penalty function are both the zero function.
Moreover, it is also possible to consider problem scenarios in which there is only one fixed resource limit, say $\tilde \tau,$ by setting $\mathcal{M}=\{\tilde \tau\}.$
\end{remark}

\section{Finite Number of Resource Limits} \label{sec_algorithmic_solution}

In this section, we assume $\mathcal M$ to be a finite set of grid points within the admissible resource range of each round $(0,\tau_{\mathrm{max}}]$.
In the following, we denote by $I_t$ the chosen arm and by $\tau_t$ the resource limit set by a learner in round $t\in [T],$ where the learner should be clear from the context.

\subsection{Reduction to Classical MAB Problem} \label{sec_reduction_MAB}

For any pair  $(i,\tautemp) \in [n] \times \mathcal{M}$, define the sub-optimality gap of this pair by means of 
\begin{align} \label{def_sub_opt_gap}
\Delta_{i,\tautemp} := \nu^* - \nu_{i,\tautemp}.
\end{align}
With this, it is straightforward to show that the cumulative (pseudo-) regret in (\ref{regret_def}) admits a regret decomposition similar to the pseudo-regret in the classical MAB problem (see Lemma 4.5 in \cite{lattimore2020bandit}):
\begin{align} \label{regret_decomposition}
\mathcal R_T = \sum\nolimits_{(i, \tau ) \in [n]\times \mathcal M} \Delta_{i,\tautemp} \, \mathbb{E} ( T_{i,\tautemp}(T+1)),
\end{align}
where $$T_{i,\tautemp}(t) = \sum\nolimits_{s=1}^{t-1} \IND{ \{ I_s=i \, \wedge \, \tau_s = \tautemp \} }$$ is the number of times the pair  $(i,\tautemp) \in [n] \times \mathcal{M}$ has been chosen till round $t \in \{1,\ldots,T\}.$
In light of this, one might be tempted to cast the considered bandit problem  into a classical MAB problem by considering each pair  $(i,\tautemp) \in [n] \times \mathcal{M}$ as an arm (``virtual arm'') which generates the ``reward'' sequence 
\begin{align} \label{def_rewards_sequence}
\big(   ( R_{i,t}   - c(C_{i,t})  ) \IND{\{C_{i,t}\leq \tau \} } - \lambda(\tau) \IND{\{C_{i,t}> \tau \}  }\big)_{t=1,\ldots,T} \enspace ,
\end{align}
each having expected value $\nu_{i,\tau}.$
Thus, the bandit problem at hand can be na\"ively considered as an unstructured class of (classical) multi-armed bandits $\mathcal E = \times_{i \in[n],\tautemp \in \mathcal{M}} \mathcal{P}_{i,\tautemp}$ (see Section 4.3 in \cite{lattimore2020bandit}), where $\mathcal{P}_{i,\tautemp}$ is a set of bivariate probability distributions on $[0,1]\times \R_+.$

As indicated by the reduction, the considered bandit problem can in principle be tackled by any bandit algorithm for the classical MAB problem by means of interpreting each pair $(i,\tautemp) \in [n] \times \mathcal{M}$ as an (virtual) arm.
However, as we shall see in the following section, exemplified on the basis of UCB \cite{auer2002finite}, this straightforward reduction seems to be sub-optimal, as the available information of the possibly censored type of feedback is not incorporated in an appropriate way.
This is in fact not surprising, because the problem at hand is actually not an unstructured bandit problem, as for each fixed arm $i\in[n],$ there is a relationship between the probability distributions $(\mathcal{P}_{i,\tautemp})_{\tautemp \in \mathcal{M}}$ due to the joint reward and resource consumption distribution $P_i^{(r,c)}.$ This relationship is lost by the reduction.

\subsection{Penalized Expected Gain Estimates} \label{sec_estimates}

Considering the desired value $\nu^*$, which arises from (\ref{def_max_problem}), one certainly needs to estimate the penalized expected gain $\nu_{i,\tautemp}$ in \eqref{def_nu_i_tau} in a suitable way.
Regarding their form, one needs for each pair  $(i,\tautemp) \in[n] \times \mathcal{M}$ suitable estimates for both the expected gain 
$$ g_{i,\tau} :=	\mathbb E_{P_i^{(r,c)}} \Big( \big( X_i^{(r)} - c(X_i^{(c)}) \big) \cdot \IND{ \{X_i^{(c)} \leq  \tau\}  }  \Big)  	$$
and the expected penalty term
$$	\Lambda_{i,\tau} := \mathbb E_{P_i^{(c)}} \Big( \lambda(\tau) \IND{  \{X_i^{(c)}>\tau\} } \Big) = \lambda(\tau) \, (1-P_i^{(c)}(\tau)) .	$$
For the expected gain we define for any round $t\in [T]$ the estimate 
\begin{align} 
\label{def_emp_gain_estimate}
\begin{split}
\hat g_{i,\tautemp}(t) &=  \frac{\sum_{s=1}^{t-1} ( R_{i,s} - c(C_{i,s})) \cdot \IND{\{ C_{i,s}\leq \tautemp  \} } \cdot \IND{\{   I_s=i \, \wedge \, \tau_s \geq \tautemp \}	} }{ N_{i,\tautemp}(t)} ,   
\end{split}
\end{align}
where $$	N_{i,\tautemp}(t) = \sum_{s=1}^{t-1} \IND{\{  I_s = i \, \wedge \, \tau_s \geq \tautemp \}	}.$$ 
The expected penalty term in round $t\in [T]$ can be estimated via
\begin{align} 
\label{def_emp_penalty_estimate}
\begin{split}
\hat \Lambda_{i,\tautemp}(t) &=  \lambda(\tau) \, \frac{\sum_{s=1}^{t-1} \IND{\{ C_{i,s} > \tautemp  \} } \cdot \IND{\{   I_s=i\}	} }{ N_{i,0}(t)},
\end{split}
\end{align}
which is simply the empirical survival function estimate.
Thus, combining \eqref{def_emp_gain_estimate} and \eqref{def_emp_penalty_estimate}, our suggested penalized expected gain estimate for a pair  $(i,\tautemp) \in[n] \times \mathcal{M}$ is
\begin{align} \label{def_exp_gain_estimate}
\hat \nu_{i,\tautemp}(t) =  \hat g_{i,\tautemp}(t) - \hat \Lambda_{i,\tautemp}(t).
\end{align}
These estimates admit a simple update rule for the chosen arm $I_t \in [n]$ in round $t.$ 
Indeed, it holds that $\hat \Lambda_{I_t,\tautemp}(t+1) = \frac{\lambda(\tau)}{N_{i,0}(t)+1} \big( \frac{N_{i,0}(t) \, \hat \Lambda_{I_t,\tautemp}(t)}{\lambda(\tau)} +  \IND{\{ C_{I_t,t} > \tautemp  \} }   \big) $ and
\begin{itemize} 
%
\item if $\tautemp > \tau_t,$  then $\hat g_{I_t,\tautemp}(t+1)=\hat g_{I_t,\tautemp}(t),$
\item if $\tautemp \leq  \tau_t,$ then 
$  \hat g_{I_t,\tautemp}(t+1) =	{ \begin{cases}
\frac{ N_{I_t,\tautemp}(t) \cdot \hat g_{I_t,\tautemp}(t)  + (R_{I_t,t} - c(C_{I_t,t}))  }{ N_{I_t,\tautemp}(t)+1}, & {\small C_{I_t,t} }\leq \tautemp, \\
\frac{ N_{I_t,\tautemp}(t) \cdot \hat g_{I_t,\tautemp}(t) }{ N_{I_t,\tautemp}(t)+1}, & C_{I_t,t} > \tautemp,
\end{cases} }$ \\
as well as $N_{I_t,\tautemp}(t+1) = N_{I_t,\tautemp}(t) +1.$
\end{itemize}
Note that this update has a complexity of $O(|\mathcal{M}|)$. Moreover, $\hat g_{I_t,\tautemp}$ is updated for all resource limits $\tautemp$ below the currently chosen one (i.e., $\tau_t$), even though the feedback was possibly censored, i.e., in the case where $C_{I_t,t} > \tautemp$.
This is in particular advantageous compared to a standard plug-in estimate (i.e., see \eqref{def_standard_estimate} in the appendix), which does not adapt the estimate value for censored observations.

Besides their appealing property of extracting all available feedback information, these estimates also allow for deriving suitable confidence intervals by exploiting results from the theory of martingales and using a peeling argument.
Indeed, for confidence lengths defined for each pair  $(i,\tautemp) \in[n] \times \mathcal{M}$ in round $t\in [T]$ by
\begin{align*}
\mathfrak c_{i,\tautemp}(t;\alpha) 
&= \mathfrak c_{i,\tautemp}^{(g)}(t;\alpha) + \mathfrak c_{i,\tautemp}^{(\Lambda)}(t;\alpha)  \\
&= \sqrt{ (2 \alpha \log(t))/ N_{i,\tautemp}(t)}   + \lambda(\tau) \sqrt{ (2 \alpha \log(t))/ N_{i,0}(t)}, \quad \alpha>1, 
\end{align*}
we obtain the following result (cf.\ Section~\ref{sec_appendix_proof_concentration} for the proof).
\begin{proposition} \label{prop_nu_estimates}
Let $(i,\tautemp) \in [n]\times \mathcal{M}$ and $\alpha>1.$
Then, for any round $t \in [T]$, it holds that 
	\begin{align*}
		\mathbb{P}\big( \hat \nu_{i,\tautemp}(t) -  \nu_{i,\tautemp} > \mathfrak c_{i,\tautemp}(t;\alpha)  \big) 
		&\leq  2 \left(1+\frac{\log(t)}{\log(\frac{\alpha+1}{2})} \right) t^{-\frac{2\alpha}{\alpha+1}}, 
	\end{align*}
and the right-hand side is also an upper bound for  $\mathbb{P}\big( \hat \nu_{i,\tautemp}(t) -  \nu_{i,\tautemp} < - \mathfrak c_{i,\tautemp}(t;\alpha) \big).$
\end{proposition}

\subsection{Resource-censored Upper Confidence Bound} \label{sec_RCUCB}

Another appealing property of the estimates introduced in the previous section and especially the underlying counter variables $N_{i,\tautemp}$ is the possibility to refine the regret decomposition in \eqref{regret_decomposition}, which in turn will provide insights into the question why a learner revolving around $\hat \nu_{i,\tautemp}$ will in general improve upon a na\"ive reduction to the standard MAB problem.
More specifically, for $\tau\neq \tau_{\mathrm{max}}$ let $$\up(\tautemp) = \min\nolimits_{\tilde \tau \in \mathcal{M}\backslash\{\tau_{\mathrm{max}}\} }\{ \tilde \tau > \tautemp\}$$ and $T^+ = T+1$. Then, we can write the cumulative regret $\mathcal R_T $ as
\begin{align} \label{eq_regret_decomp_refined}
\begin{split}
\mathcal R_T 
&= \sum\limits_{\stackrel{\tautemp \in  \mathcal{M}\backslash\{\tau_{\mathrm{max}}\} }{i \in [n]} } \Delta_{i,\tautemp} \big(   \mathbb E(N_{i,\tautemp}(T^+))  - \mathbb E(N_{i,\up(\tautemp)}(T^+))  \big)  \\
&\quad + \sum_{i \in [n] } \Delta_{i,\tau_{\mathrm{max}}} \mathbb E(T_{i,\tau_{\mathrm{max}}}(T^+)),
\end{split}
\end{align}
where we used that $ T_{i,\tautemp}(T^+) = N_{i,\tautemp}(T^+)  -  N_{i,\up(\tautemp)}(T^+)$ for any $(i,\tautemp)$ with $\tautemp \neq \tau_{\mathrm{max}}.$
Thus, to keep the number of sub-optimal arm pulls of a specific arm/resource-limit pair low (i.e., $T_{i,\tautemp}$), one can play the same arm but with the next larger resource limit (i.e., increase $N_{i,\up(\tautemp)}$), which in turn will increase $ N_{i,\tilde \tau}$ for all $\tilde \tau \leq \tautemp,$ but simultaneously improve the estimation accuracy of $(\hat \nu_{i,\tilde \tau})_{\tilde \tau \leq \tautemp}.$ 
In some sense, this consideration suggests that a certain generosity regarding the choice of the resource limit might be favorable.

Inspired by these insights, we define the Resource-censored Upper Confidence Bound ($\texttt{RCUCB}$) algorithm: In the first $n$ rounds, each arm is chosen once with the maximal available resource limit, i.e., $(I_t,\tau_t) = (t,\tau_{\mathrm{max}})$ for $t\in [n].$
Then, in each subsequent round $t\in\{n+1,\ldots,T\}$, the arm and the resource limit are chosen as follows:
\begin{align} \label{RCUCB_choice}
\left( I_t  ,	\tau_t  \right) \in  \argmax_{ (i,   \tautemp) \in [n] \times \mathcal{M}} \big(  \hat \nu_{i,\tautemp}(t) + \mathfrak c_{i,\tautemp}(t;\alpha)\big),
\end{align}
where ties are broken arbitrarily and $\alpha>0$ is a fixed parameter of choice.

Note that unless the penalty function $\lambda$ increases too drastically from $\tautemp$ to $\up(\tautemp)$, the confidence length $\mathfrak c_{i,\tautemp}$ of any arm $i\in[n]$ is likely to be smaller than  $\mathfrak c_{i,\up(\tautemp)}$  in cases where $N_{i,\tautemp} >  N_{i,\up(\tautemp)}$ holds\footnote{Recall that $N_{i,\tautemp} \geq  N_{i,\up(\tautemp)}$ always holds.}. 
Thus, $\texttt{RCUCB}$'s exploration behavior tends to be biased towards higher resource limits, which in turn is preferable regarding the discussion above\footnote{The exploration behavior of UCB-type algorithms is mainly driven by the confidence intervals.}.

Note that the main novelty of $\texttt{RCUCB}$ lies primarily in the composition of the underlying exploitation term $\nu_{i,\tau}$, since it consists of two components that are already complex terms in themselves. Indeed, the first component, the expected gain estimate $\hat g_{i,\tau}$, is designed such that information from the potentially censored feedback is still extracted while ensuring the construction of valid confidence intervals. The second component, the estimator of the expected penalty term $\hat{\Lambda}_{i,\tau}$, is an empirical survival function estimate and correspondingly more complex than a classical empirical mean. 

We obtain the following upper bound on the cumulative regret of  $\texttt{RCUCB}$ (see Section~\ref{sec_appendix_proof_regret_upper_bound} for the proof).

\begin{theorem}\label{theorem_regret_upper_bound}
Let $\alpha>1$ in \eqref{RCUCB_choice}. 
Then, for any number of rounds $T,$ $\varepsilon\in(0,1),$ and any $\delta \in(0,1/2)$ such that $ (n|\mathcal{M}|-1)^{1-\delta} T^{2\delta} \leq T$,  it holds that
$$\mathcal R_T^{\texttt{RCUCB}} \leq \sum\limits_{(i,   \tautemp) \in [n] \times \mathcal{M}} \Delta_{i,\tau} \, u_{i,\tautemp}(T,\alpha) - \mathbb{P}(A_\varepsilon) \sum\limits_{(i,\tau) \in [n]\times \mathcal{M}\backslash\{\tau_{\mathrm{max}}\} } \Delta_{i,\tau} \, l_{i,\up(\tautemp)}(T,\alpha),$$ 
where 
\begin{align*}
u_{i,\tautemp}(T,\alpha) &:= \frac{8 \alpha  (1+\lambda(\tautemp))^2 \log(T)}{\Delta_{i,\tautemp}^2 } 
+ 1 + \frac{8}{\log\left(\frac{\alpha+1}{2}\right)} \left( \frac{\alpha+1}{\alpha-1} \right)^2, 
\\
l_{i,\up(\tautemp)}(T,\alpha)  &:= \frac{  \alpha \, \varepsilon \, \delta \,  \log(T)}{   H_{i,\up(\tautemp)}(\alpha) }, \\
A_\varepsilon&:= \bigcap\limits_{ (i,   \tautemp) \in [n] \times \mathcal{M}, \ t \in [T]} A_{i,\tau,t,\varepsilon},\\
A_{i,\tau,t,\varepsilon} &:=  \{	\hat \nu_{i,\tautemp}(t) + (1-\varepsilon) \,  \mathfrak c_{i,\tautemp}(t;\alpha) \geq   \nu_{i,\tautemp} \geq \hat \nu_{i,\tautemp}(t) -   \mathfrak c_{i,\tautemp}(t;\alpha) \},
\end{align*}
and $H_{i,\tau}(\alpha) := \max_{(j, \tau' ) \in [n]\times \mathcal M: j \neq i \vee \tau' \neq \tau} \Big(	\frac{8(1+\lambda(\tau')^2 \alpha)}{\Delta_{i,\tau}-\Delta_{j,\tau'}}		+1 \Big)^2  (\Delta_{i,\tau}-\Delta_{j,\tau'})^2.$
\end{theorem}
Theorem \ref{theorem_regret_upper_bound} reveals why $\texttt{RCUCB}$ is in general superior to the straightforward mapping to the standard MAB problem. 
The terms $u_{i,\tautemp}$ correspond (up to multiplicative constants) to the upper bounds on the expected sub-optimal arm pulls of the na\"ive UCB variant (cf.\ Corollary \ref{coroallary_regret_upper_bound_ucb}), from which $l_{i,\up(\tautemp)}$ is subtracted (with a probabilistic weight $\mathbb{P}(A_\varepsilon)$). 
The term $l_{i,\up(\tautemp)}$ is a lower bound for the expected number of sub-optimal arm pulls of an (virtual) arm's ``higher resource neighbor'', i.e., the (virtual) arm corresponding to $(i,\up(\tautemp)).$ 
\begin{remark}
The term $H_{i,\tau}$ occurring in the lower bound for the expected number of sub-optimal arm pulls is essentially the largest difference between the sub-optimality gap of the corresponding arm/resource-limit pair and any other sub-optimality gap of an arm/resource-limit pair. 
In particular, this term is small (such that  $l_{i,\up(\tautemp)}$ is large) if the sub-optimality gaps of the arm/resource-limit pairs are similar, i.e., the learning problem is difficult.
The event $A_\varepsilon$ is the anytime concentration (up to some $\varepsilon$ relaxation for the upper deviation) of the estimate's confidence bounds around the ground-truth value. Using a union bound and Hoeffding's inequality it is straightforward to show that $A_\epsilon$ has strictly positive mass.
\end{remark}

\begin{remark}
Note that the terms $\varepsilon$ and $\delta$ could be chosen appropriately such that the second term in the regret bound is as large as possible.
Furthermore, it is worth mentioning that the term $(1+\lambda(\tautemp))$ occurring in the upper as well as lower bound terms is due to the crude estimate $1\leq N_{i,0}(T^+)$ used to bound $(1+\lambda(\tautemp)/N_{i,0}(T^+)).$ 
We conjecture that this bound can be refined to $(1+ \tilde C\, \lambda(\tautemp)/\log(T)),$ for an appropriate constant $\tilde C>0.$
\end{remark}
Regarding the update complexity of $\texttt{RCUCB}$, we can derive the following result, which is proven in Section \ref{sec_proof_updates}.
\begin{proposition} \label{prop_update_complexity_RCUCB}
	$\texttt{RCUCB}$ has a worst case update complexity of order $O(n|\mathcal{M}|).$
\end{proposition}
Note that the update complexity for most of the state-of-art bandit algorithms combined with the reduction in Section \ref{sec_reduction_MAB} is of order $O(n|\mathcal{M}|)$ as well, since these are usually linear in the number of arms, which are $n|\mathcal{M}|$ many in light of the reduction.

\section{Arbitrary Resource Limits} \label{sec_algorithmic_solution_part2}

We now turn to the case in which $\mathcal{M}$ equals $(0,\tau_{\mathrm{max}}]$.
Obviously, the challenge in this variant is to cope with the infinite size of the decision set $[n] \times \mathcal{M}.$ 
To this end, we will follow the ideas of the \emph{zooming} algorithm \cite{kleinberg2008multi} or the StoOO algorithm \cite{munos2014bandits} and maintain finite subsets of $\mathcal{M},$ one for each arm, which will be refined successively in order to include resource limits $\tau,$ where the penalized expected gain of an arm is believed to be large.
The exploration-exploitation behavior will be up to an additional bias-correction guided by the upper confidence term considered in \texttt{RCUCB}.

\subsection{Zooming-$\texttt{RCUCB}$}

A zooming algorithm seeks to find a good approximation of the optimum of an unknown stochastic function $f:\mathcal{X} \to \mathbb{R}$ over a (semi-)metric space $\mathcal{X}$ with (semi-)metric $d$ using a numerical budget $T$ for the maximal number of function evaluations. 
For this purpose, a zooming algorithm constructs a hierarchical partitioning of $\mathcal{X}$ into nested subsets in an online manner. 
Each subset is associated with a specific point, usually the center of the subset, at which the function $f$ may be evaluated. 
For each subset the algorithm maintains an estimate of the function value at the center point as well as an confidence interval representing the uncertainty in the estimation. 
Further, by assuming structural properties on the expected value of $f$ such as Lipschitz continuity (w.r.t.\ $d$) the algorithm maintains a bias correction term for the function value at the center point depending on the size of the associated subset. 

In each time step the algorithm chooses to evaluate the function at the center point with the highest potential to be an optimal point by taking the confidence interval and the bias correction into account. 
Once the width of the confidence interval is smaller than the bias term of $f$ at a center point, the corresponding subset is refined into smaller subsets. 
This adds new center points to the set of considered center points, at which the estimate of $f$, as well as the confidence width and potential bias is computed, while the center point responsible for the refinement is left out of the consideration (or is used as the center point for a smaller subset). 
The rationale behind this approach is to ``zoom'' successively into regions of $\mathcal{X},$ where the optimum of $f$ is located.

\begin{figure}
	\centering
	{\includegraphics[width=0.9\linewidth]{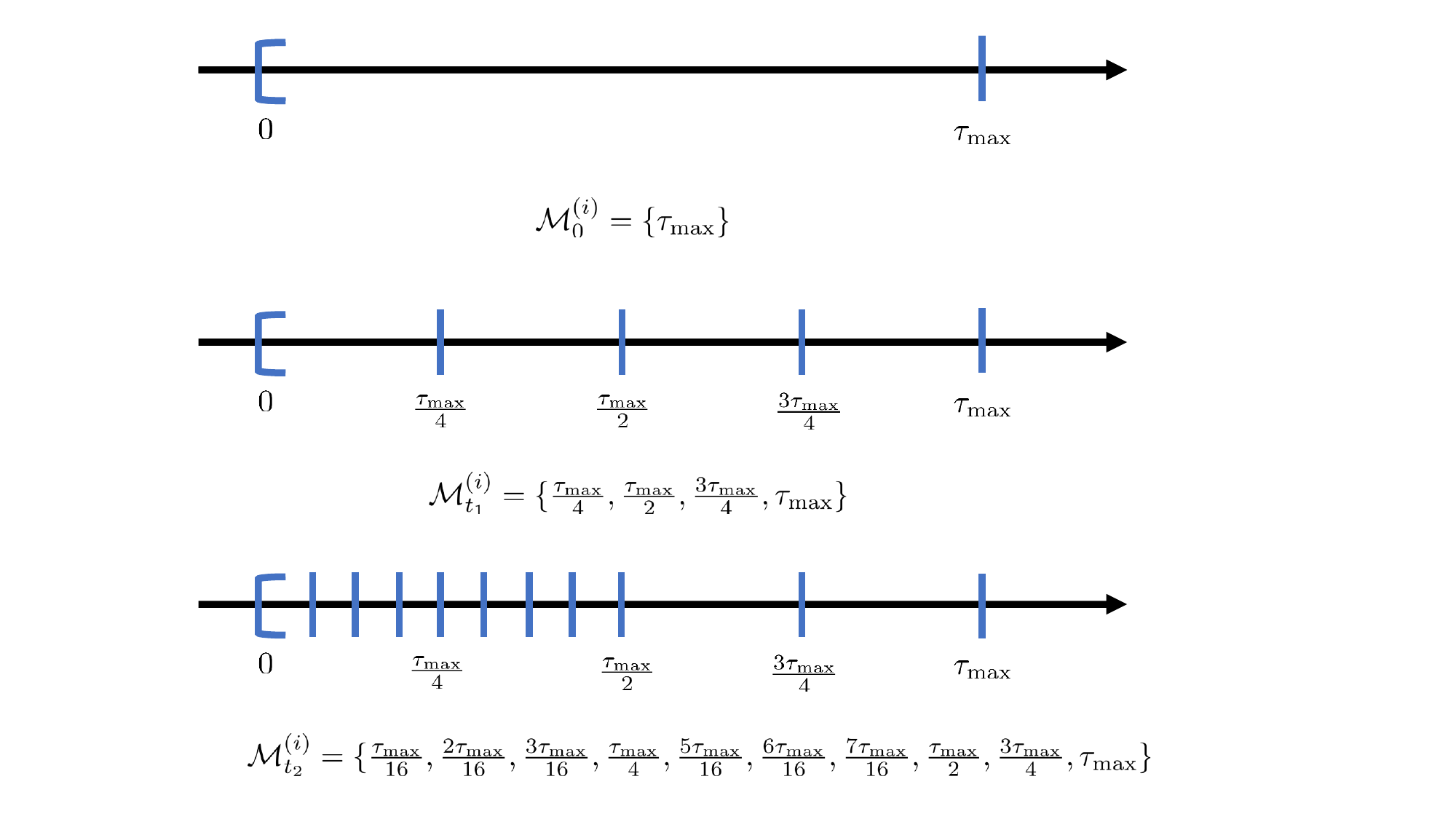}}
	\caption{Initial grid point set $\mathcal{M}^{(i)}_0$ (top plot). Extension of $\mathcal{M}^{(i)}_0$ at the point $\tau_{\mathrm{max}}$ at some time step $t_1$ for $m=4$ (middle plot). Multiple extension of $\mathcal{M}^{(i)}_{t_1}$ at the points $\tau_{\mathrm{max}}/4$ and $\tau_{\mathrm{max}}/2$ at some time step $t_2>t_1$ for $m=4$ (bottom plot). }
	\label{fig:gridpoints}
\end{figure}

Following the idea of zooming algorithms, we maintain for any arm $i\in[n]$ a time-dependent grid set $\mathcal{M}^{(i)}_t,$ where $\mathcal{M}^{(i)}_0 = \{ \tau_{\mathrm{max}}\}.$ 
For any arm $i\in[n]$ and $l=1,\ldots,|\mathcal{M}^{(i)}_t|$ denote by $\tau_l^{(i)}(t)$ the grid points in $\mathcal{M}^{(i)}_t.$ 
Each point is representing the left-open interval $(\low_i(\tau_l^{(i)}(t)),\tau_l^{(i)}(t)],$ where
$$\low_i(\tautemp) = \max\nolimits_{\tilde \tau \in \mathcal{M}^{(i)}_t }\{ \tilde \tau \leq \tautemp\}$$ 
is the next smallest grid point to $\tautemp$ in $\mathcal{M}^{(i)}_t$, and we set $\low_i(\tautemp) =0$ if the set is empty.
We say that $\mathcal{M}^{(i)}_t$ is extended at some point $\tau_l^{(i)} \in \mathcal{M}^{(i)}_t$ if the grid points of the equidistant decomposition\footnote{All grid points are in the interior of the interval.} of $[\low_i(\tau_l^{(i)}(t)),\tau_l^{(i)}(t)]$ with size $m-1$ is added to $\mathcal{M}^{(i)}_t.$ 
{  For an illustration of the grid point sets consider Figure \ref{fig:gridpoints}, where the initial grid point set $\mathcal{M}^{(i)}_0$ is illustrated in the top plot, while the middle plot shows the extension of $\mathcal{M}^{(i)}_0$ at the point $\tau_{\mathrm{max}}$ at some time step $t_1$ for $m=4.$ Here, for instance, $\low_i(\tau_{\mathrm{max}})=\frac{3\tau_{\mathrm{max}}}{4}$ and $\low_i(\frac{\tau_{\mathrm{max}}}{4})=0,$ so that $\tau_{\mathrm{max}}$ represents the left-open interval $(\frac{3\tau_{\mathrm{max}}}{4},\tau_{\mathrm{max}}],$ while $\frac{\tau_{\mathrm{max}}}{4}$ represents $(0,\frac{\tau_{\mathrm{max}}}{4}].$ }
Note that the size $m\geq 2$ is fixed and specified by the learner and the same holds true for the criterion leading to an extension of a grid set. 

\begin{algorithm}[H]
	\textbf{Input:}  $\alpha>1,$ $m\geq 2,$ $\delta\in(0,1),$ $T\in \N$\\
	\textbf{Initialization:}  $\mathcal{M}^{(i)}_t = \{ \tau_{\mathrm{max}}\}$ $\forall i \in [n],$ $\hat \nu_{i,\tautemp} = 0,$ $\mathfrak c_{i,\tautemp} = \infty$ $\forall  (i,   \tautemp) \in [n] \times \mathcal{M}^{(i)}_t$
	\begin{algorithmic}[1]
		\For {$t\in \{1,2,\ldots,T\}$}
		\State  Choose $\left( I_t  ,	\tau_t  \right)$ according to \eqref{zRCUCB_choice}
		\State Observe $X_{I_t,t}$  \hfill (cf.\ \eqref{def_feedback})
		\State Update $\hat \nu_{I_t,\tautemp},$ $\mathfrak c_{I_t,\tautemp}$ $\forall \tau \in  \mathcal{M}^{(I_t)}_t$
		\For {$\tau \in \mathcal{M}^{(I_t)}_t$}
		\If { $(\tau-\low_{I_t}(\tau))   \geq  \mathfrak c_{I_t,\tautemp}$ } 
		\State $\mathcal{M}^{(I_t)}_t \leftarrow $ \texttt{extend}($\mathcal{M}^{(I_t)}_t,m,\tau$) 
		\State Initialize $\hat \nu_{I_t,\tautemp}$ $\mathfrak c_{I_t,\tautemp}$ for all new grid points $\tau$ in $\mathcal{M}^{(I_t)}_t$
		\EndIf
		\EndFor
		\EndFor						
		\State  Let $(\hat i, \hat \tau)$ be the arm/resource-limit pair with the finest grid level among $(\mathcal{M}^{(i)}_t)_{i\in [n]}$ \\
		\Return $(\hat i, \hat \tau)$
	\end{algorithmic}
	\caption{The  $\texttt{z-RCUCB}$ algorithm}
	\label{algo_z_RCUCB}
\end{algorithm}

In light of this, we suggest the $\texttt{z-RCUCB}$ (zooming-$\texttt{RCUCB},$ Algorithm \ref{algo_z_RCUCB}), which adapts the choice criterion \eqref{RCUCB_choice} of \texttt{RCUCB} and additionally refines the finite grid points by ``zooming'' into subsets of $\mathcal{M},$ where the penalized expected gain of an arm seems to be large.
More precisely, for some $\delta\in(0,1)$ the arm/resource limit pair in time step $t$ chosen by  $\texttt{z-RCUCB}$ is
\begin{align} \label{zRCUCB_choice}
\left( I_t  ,	\tau_t  \right) \in  \argmax_{ (i,   \tautemp) \in [n] \times \mathcal{M}^{(i)}_t} \big(  \hat \nu_{i,\tautemp}(t) + \mathfrak c_{i,\tautemp}(n\, T^2/\delta;\alpha)   + (\tau-\low_i(\tau))  \big).
\end{align}
The set of grid points $\mathcal{M}^{(I_t)}_t$ is extended at some point $\tau \in \mathcal{M}^{(I_t)}_t$ if 
\begin{align} \label{def_ext_crit}
    (\tau-\low_{I_t}(\tau))   \geq  \mathfrak c_{I_t,\tautemp}(n\, T^2/\delta;\alpha) 
\end{align}
holds.
Roughly speaking, the extension criterion is used in the case where the discretization bias (represented by the left-hand side in \eqref{def_ext_crit}) for the left-open interval $(\low_{I_t}(\tau),\tau]$  is larger than the uncertainty of its representing grid point $\tau$ (represented by the confidence interval $\mathfrak c_{I_t,\tautemp}$).
Note that as all counter variables $ N_{I_t,\tilde \tau}$ such that $\tilde \tau \in \mathcal{M}_t^{(I_t)}$ and $\tilde \tau \leq \tau_t$ hold are incremented in round $t,$ it could be that more than one grid point in $\mathcal{M}^{(I_t)}_t$ is extended in one round $t.$ Such a multiple extension is illustrated in the bottom plot of Figure \ref{fig:gridpoints}.

In comparison to \eqref{RCUCB_choice}, the choice criterion in \eqref{zRCUCB_choice} incorporates an additional term $(\tau-\low_i(\tau)),$  which can be interpreted as a bias-correction due to the discretization of $\mathcal{M}$ by  $\mathcal{M}^{(i)}_t.$
Moreover, the ($\alpha$ root of the) confidence level is set to $n\, T^2/\delta$, which requires the knowledge of $T.$
In case the number of learning rounds $T$ is not known beforehand, one can use the well-known \emph{doubling trick} \cite{cesa2006prediction} to obtain an algorithm that preserves the theoretical guarantees of an algorithm that needs to know the number of learning rounds.  
Note that the parameters of $\texttt{z-RCUCB}$ are the exploration constant $\alpha>1,$ the confidence level $\delta\in(0,1),$ the grid refinement size $m\geq 2$ and the total number of learning rounds $T.$

\subsection{Theoretical Guarantees}
Similarly as for the infinite multi-armed bandit case or $\mathcal{X}$-armed bandits \cite{bubeck2011x,munos2014bandits}, we will focus the theoretical analysis  on the loss after $T$ many rounds (simple regret) given by
$  L_T:= \nu^* -  \nu_{\hat i(T), \hat \tau(T)},$ where $(\hat i(T), \hat \tau(T))$ is the arm/resource-limit pair with the finest grid level among $(\mathcal{M}^{(i)}_T)_{i\in [n]}.$
For this purpose, we will make the following assumption on the local smoothness of the optimal penalized expected gain (interpreted as a function of $\tau$):
$$  \nu_{i^*,\tau^*}  - \nu_{i^*,\tau} \leq |\tau^* - \tau |, \qquad \forall \tau \in \mathcal{M}. $$
Such an assumption is common in $\mathcal{X}$-armed bandits and one of the weakest assumptions in this regard \cite{grill2015black}.

Next, we need the notion of $\eta$-near-optimality dimension to capture the possible rate of convergence of the resulting estimates using the successive discretization process via the finite grid points above for the problem at hand.  
\begin{definition} \label{defi_near_opt}
The $\eta$-near-optimality dimension is the smallest $d\geq0$ such that there exists a constant $C>0$ (the $\eta$-near-optimality constant) such that for all $\varepsilon>0,$ the maximal number of disjoint sets of the form
$$ \{j\}\times(a_j,b_j], \qquad j\in[n], 0\leq a_j<b_j \leq \tau_{\mathrm{max}} $$
such that  $|b_j -a_j| \leq \eta \varepsilon$ and $(j,b_j)$ is an element of $\{	(i,\tau) \in [n]\times \mathcal{M} \, | \,     \nu_{i,\tau} \geq \nu_{i^*,\tau^*} -	\epsilon \},$ is less than $C \varepsilon^{-d}.$
\end{definition}
Finally, we introduce for any $l\geq 1$  the equidistant grid points with granularity $m^{-l}$ via 
\begin{align*} 
%
\tau_{(l,j)} = \frac{o_j}{m} (\tau_{(l-1,s_j)} - \tau_{(l-1,s_j-1)} ), \quad j\in\{1,\ldots,m^l \}
\end{align*}
where $ s_j = \lfloor j/m \rfloor +1$ and 
$$o_j=\begin{cases}
j \mod m, & \mbox{if $j \mod m \neq 0$} \\
m, & \mbox{else}.
\end{cases}$$
Here, we set $\tau_{(l,0)}=0$ and
$\tau_{(l,m^l)}=\tau_{\mathrm{max}}$ for any $l.$ 
Note that for any $l \geq 0$ and any $j,k \in\{0,\ldots,m^l\}$ such that $|j-k|\leq 1$ it holds   
$|\tau_{(l,j)} - \tau_{(l,k)}| \leq m^{-l}.$
With this, and assuming the local smoothness, we obtain the following result for the loss of $\texttt{z-RCUCB}.$
\begin{theorem} \label{theorem_regret_upper_bound_zRCUCB}
Let $d$ be the $1/3$-near-optimality of 
$ \{	(i,\tau) \in [n]\times \mathcal{M} \, | \,     \nu_{i,\tau} \geq \nu_{i^*,\tau^*} -	\epsilon \},$ 
i.e., the set of all $\epsilon$-best arm/resource-limit pairs,  with corresponding  near optimality constant $C>0.$ 
Then, for any $\delta\in(0,1), \alpha>1$ it holds with probability at least $1-\delta$ that
$$		L_T^{\texttt{z-RCUCB}}   \leq 	\tilde C \big(	\nicefrac{ \log(T^2/\delta)}{T}	\big)^{\frac{1}{d+2}},	$$
where 
\begin{align*}
    \tilde C &= \left( \frac{4\alpha  \, C \, m^2 \, (H_T + (1+\lambda(\tau_{\mathrm{max}}))^2)}{3^d (1- m^{-(d+2)})} \right)^{1/(d+2)}, \\
	H_T &= \max_{l\geq 0, 1\leq j \leq m^l-1}  \left( (1+\lambda(\tau_{(l,j)}))^2  - \frac{1}{m^2} \left(1+\lambda(\tau_{(l-1,s_j)})\, c_T   \right)^2 \right), \\
	c_T &= \sqrt{ \frac{2\alpha \log(nT^2/\delta) \left(1+\lambda(\tau_{\mathrm{max}}) \right)^2  }{\tau_{\mathrm{max}}^2 T}}.
\end{align*}
Moreover, setting $\delta=1/T$ yields 
$ \exptd [	L_T^{\texttt{z-RCUCB}}] = O\big(\big(	\nicefrac{\log(T)}{T}	\big)^{\frac{1}{d+2}}\big).  $
\end{theorem}
Note that the speed of convergence is basically the same as the one obtained for the $\mathcal{X}$-armed bandit setting.
The difference is only regarding the constant $\tilde C$, which leads to an improvement over a straightforward application of an $\mathcal{X}$-armed bandit algorithms on $\mathcal{X}=[n]\times\mathcal{M}.$  
Indeed, using a straightforward application of some $\mathcal{X}$-armed bandit algorithm such as StoOO \cite{munos2014bandits} on $\mathcal{X}=[n]\times\mathcal{M},$ one can derive a theoretical guarantee on the loss of StoOO as in Theorem \ref{defi_near_opt}  with 
$$\tilde H_T = \max_{l\geq 0, 1\leq j \leq m^l-1} (1+\lambda(\tau_{(l,j)}))^2 $$ 
replacing $H_T.$ 
However, it obviously holds that $\tilde H_T \geq H_T$ and the gap between $\tilde H_T$ and $H_T$ can be large depending on the underlying penalty function $\lambda.$

Regarding the update complexity of $\texttt{z-RCUCB}$, we can derive the following result, which is proven in Section \ref{sec_proof_updates}.
	\begin{proposition} \label{prop_update_complexity_zRCUCB}
		$\texttt{z-RCUCB}$'s  update complexity is in $O(|\mathcal{M}^{(I_t)}_t|+(m-1)).$
	\end{proposition}

\section{Experimental Study} \label{sec_exp}
In this section, we present experimental results for our learning algorithm and compare it with variants of the Upper Confidence Bound algorithm ($\texttt{UCB}$) and Thompson Sampling ($\texttt{TS}$), adapted to the considered type of bandit problem in the spirit of Section~\ref{sec_reduction_MAB}.
For further details see Section~\ref{sec_exp_appendix}.

\subsection{Synthetic Data} \label{subsec_syn_exp}

We consider three different problem instances \textbf{PosCorr}, \textbf{NegCorr}, and \textbf{Indep}, each consisting of $n=10$ arms, where the correlation between the reward and resource consumption distribution of the arms is positive, negative, and zero\footnote{In fact, the distributions are even independent for \textbf{Indep}.}, respectively. 
The arm distributions for \textbf{PosCorr} are similar as in Figure \ref{fig:example_arm_distr} in the sense that the correlation level is the same for all arms and only the arm's means are different, while for \textbf{NegCorr} the correlation structure of \textbf{PosCorr} is simply reversed and for \textbf{Indep} no correlation structure is present at all. 
The explicit choice of the distributions is detailed in Section~\ref{sec_exp_appendix}.
For all problem instances we consider the admissible resource range $(0,1],$ i.e., $\tau_{\mathrm{max}}=1$ and  an equidistant grid point set $\mathcal{M}$ for the admissible resource range $(0,1]$ of size $10.$ 
We also consider varying grid point sizes as well as varying numbers of arms in Section~\ref{sec_exp_part_two}.
In light of our running example in Section \ref{sec_problem_intro}, we use for the cost function $c(x)=x/10$ and for the penalty function $\lambda(x)=c(x)\IND{\{x\leq 0.5\} }  + 10x \IND{\{x> 0.5\}}. $
All considered learning algorithms proceed over a total number of  $T=100,000$ rounds.

\begin{figure}[ht!]
\centering
\subfigure
{
\includegraphics[width=0.31\linewidth]{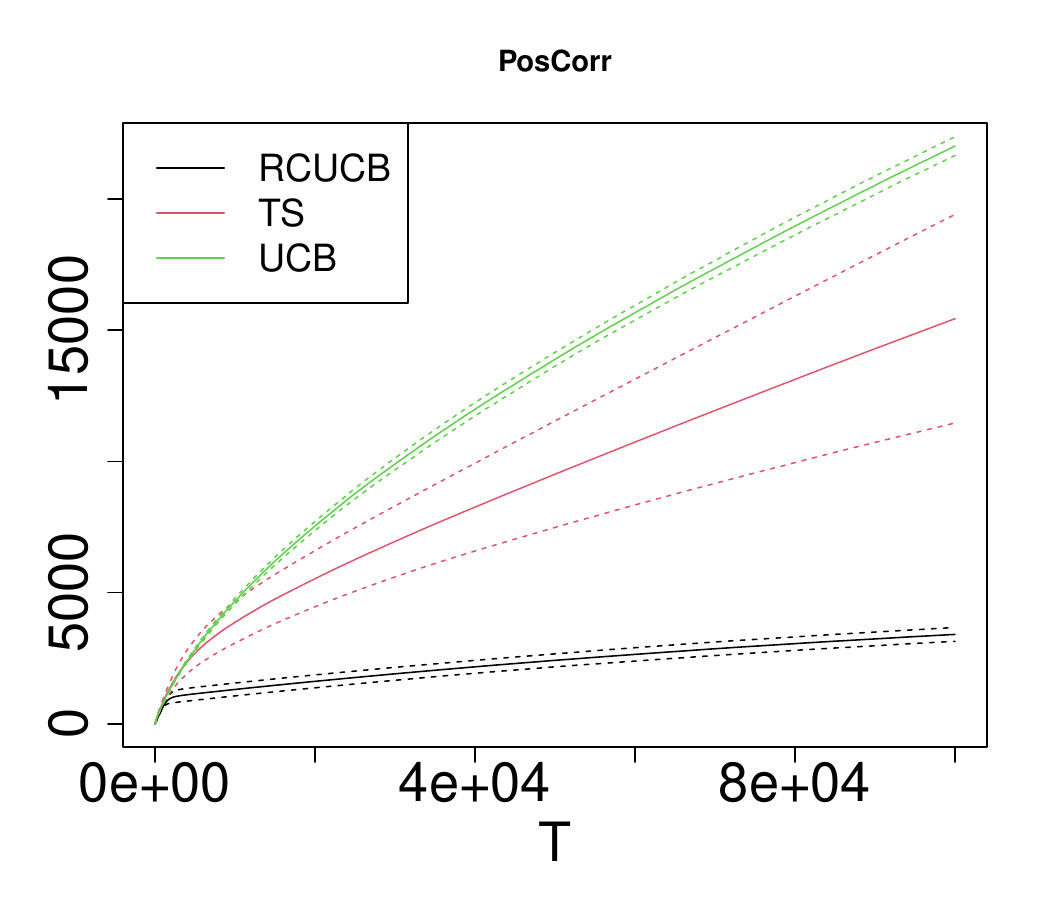}}
\subfigure
{  
\includegraphics[width=0.31\linewidth]{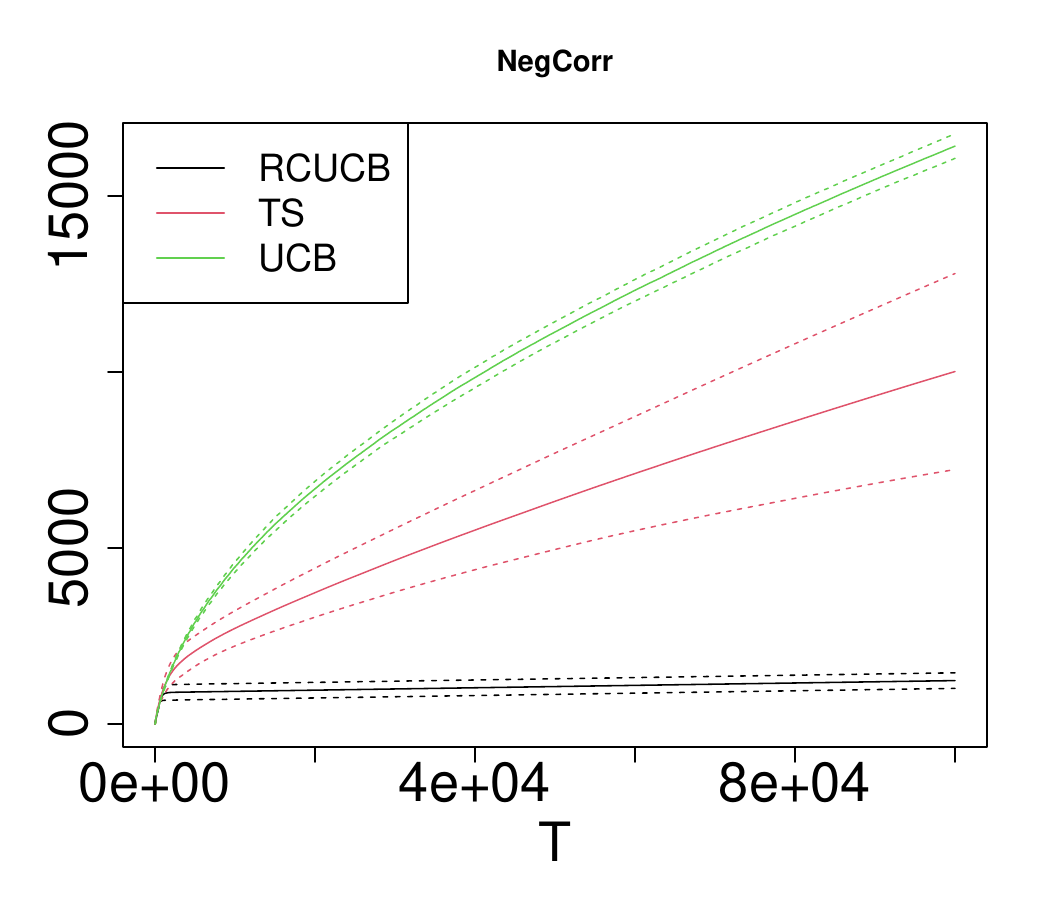}}
\subfigure
{  
\includegraphics[width=0.31\linewidth]{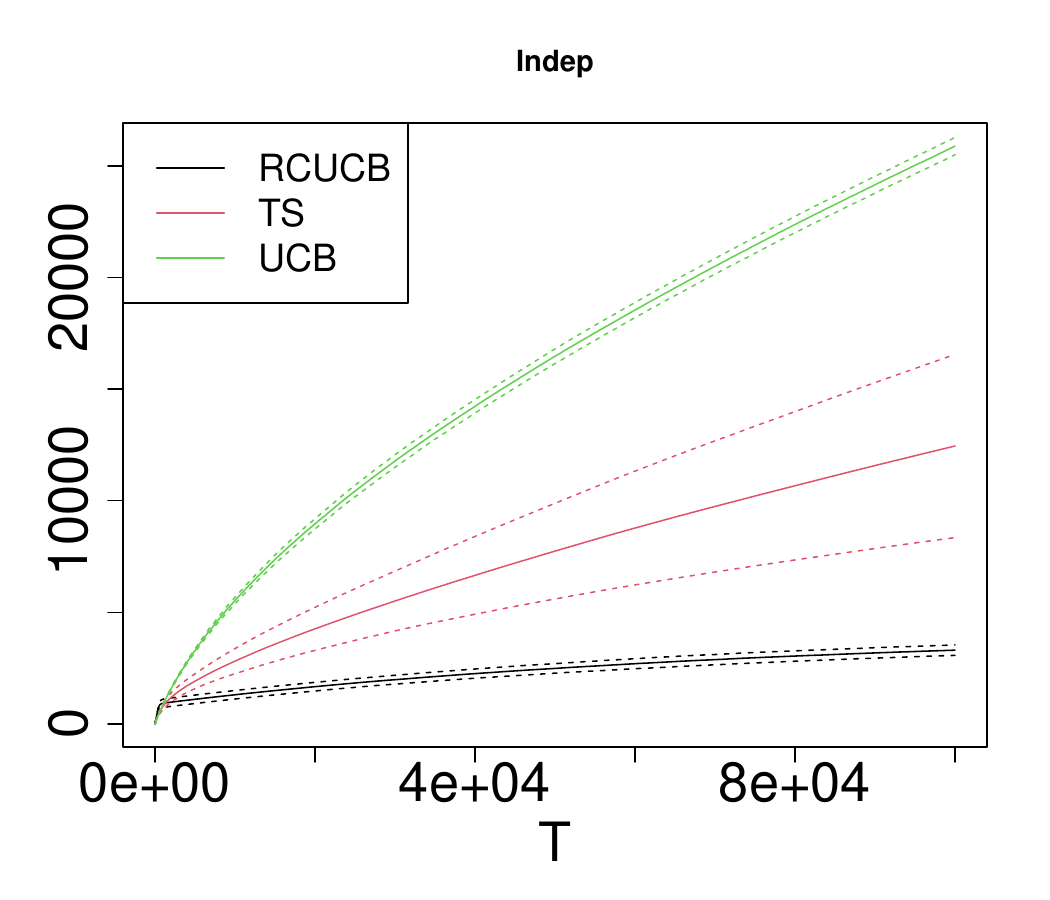}}
\caption{Mean cumulative regret (solid lines) for UCB ($\alpha=1$), TS and RCUCB ($\alpha=1$) for the \textbf{PosCorr}, \textbf{NegCorr}, and \textbf{Indep} problem instances. The dashed lines depict the empirical confidence intervals, using the standard error.}
\label{fig:regretanalysisproblem1}
\end{figure}

\noindent Figure \ref{fig:regretanalysisproblem1} illustrates the mean cumulative regret over 100 repetitions for these problem scenarios.
It is clearly visible that $\texttt{RCUCB}$ distinctly outperforms both $\texttt{UCB}$ and $\texttt{TS}$ on each problem instance.
Regarding the impact of the correlation on the performance, we see that $\texttt{RCUCB}$ reveals a much better performance, if there is a correlation - either negative or positive - present between the reward and the consumption of resource distribution compared to the considered baselines. Indeed, in Figure \ref{fig:regretanalysisproblem1}, we see that the relative gap between $\texttt{RCUCB}$ and its competitors is larger for the correlated problem settings than for the uncorrelated one. Thus, the learning behavior of $\texttt{RCUCB}$ seems to profit from available correlations of the two distributions.

Finally, the following table reports the (mean) proportion of censored rounds, i.e., where the resource limit was exceeded or equivalently no reward was observed, as well as the probability of observing a censored observation for the optimal arm/resource-limit pair $(i^*,\tau^*)$.
\begin{center}
\begin{tabular}{c|c|c|c||c}
	& $\texttt{RCUCB}$ & $\texttt{TS}$ & $\texttt{UCB}$ & $1-P_{i^*}^{(c)}(\tau^*)$ \\
	\hline
	\textbf{PosCorr}  & 0.6399 & 0.7516 & 0.7993 & 0.6116 \\
	\hline
	\textbf{NegCorr}  & 0.7700 & 0.8510 & 0.8894 & 0.7631 \\
	\hline
	\textbf{Indep}    & 0.4462 & 0.5749 & 0.5455 & 0.4404 \\
\end{tabular}
\end{center}
As expected, both $\texttt{UCB}$ and $\texttt{TS}$ seemingly fail to process the censored feedback in an appropriate way, as the proportion of rounds with censored rewards is much higher than the actual ground truth probability of obtaining censored rewards for the optimal arm/limit pair.
Thus, $\texttt{RCUCB}$ is again preferable over the two competing algorithms.

\subsection{Algorithm Configuration} \label{subsec_algo_config}
%
%
%
We consider the problem of configuring a Random Forest for regression over a variety of tuning parameters (arms) in an on-the-fly manner, within a reasonable time for the training (resource) on a specific data set.
To this end, we consider the \emph{AmesHousing} data set\footnote{https://cran.r-project.org/package=AmesHousing}, which is  randomly split into a 70:30 training/test set in each learning round, and each learner chooses a Random Forest parameter configuration as well as a time limit for the training.
As reward for the learner, we use $1$ minus the (normalized) root-mean-squared error on the test data, provided the learner's predefined limit for the training is not exceeded. 
Otherwise, the learner obtains a reward of zero, i.e., the feedback is censored.
The considered set of possible parameters of the Random Forest consists of
\begin{itemize}
%
\item the number of trees: $ \{100,200,\ldots,700\},$
\item the number of variables to randomly sample as candidates at each split: $ \{20,22,\ldots,30\},$
\item the minimal node size: $\{3,5,7,9 \},$
\item the fraction of training samples for bagging: $\{0.55, 0.632, 0.75, 0.8\}.$
\end{itemize}
The remaining parameters of the Random Forest are set as the default parameters as specified in the R-package `ranger'\footnote{ \url{https://cran.r-project.org/web/packages/ranger/index.html}}.
Each combination is treated as an arm, resulting in $n=672$ arms in total.

For the admissible range of time limits for training, we have used an equidistant grid of size $m=10$ of the interquartile range of the obtained training times if each possible configuration is run once.
Motivated by the PAR10 loss in algorithm configuration problems \cite{kerschke2019automated}, we use $c(x)=x$ for the cost function, while the penalty function is $\lambda(x)=10x.$ 
The total number of rounds is set to $T=2 \, n\, m$, and the number of repetitions to 10. 
All these experiments were conducted on a machine featuring Intel(R) Core(TM) i7-8550U@1.80 GHz CPUs with 4 cores and 16 GB of RAM.

	\begin{figure}
		\centering{  
			\includegraphics[width=0.7\linewidth]{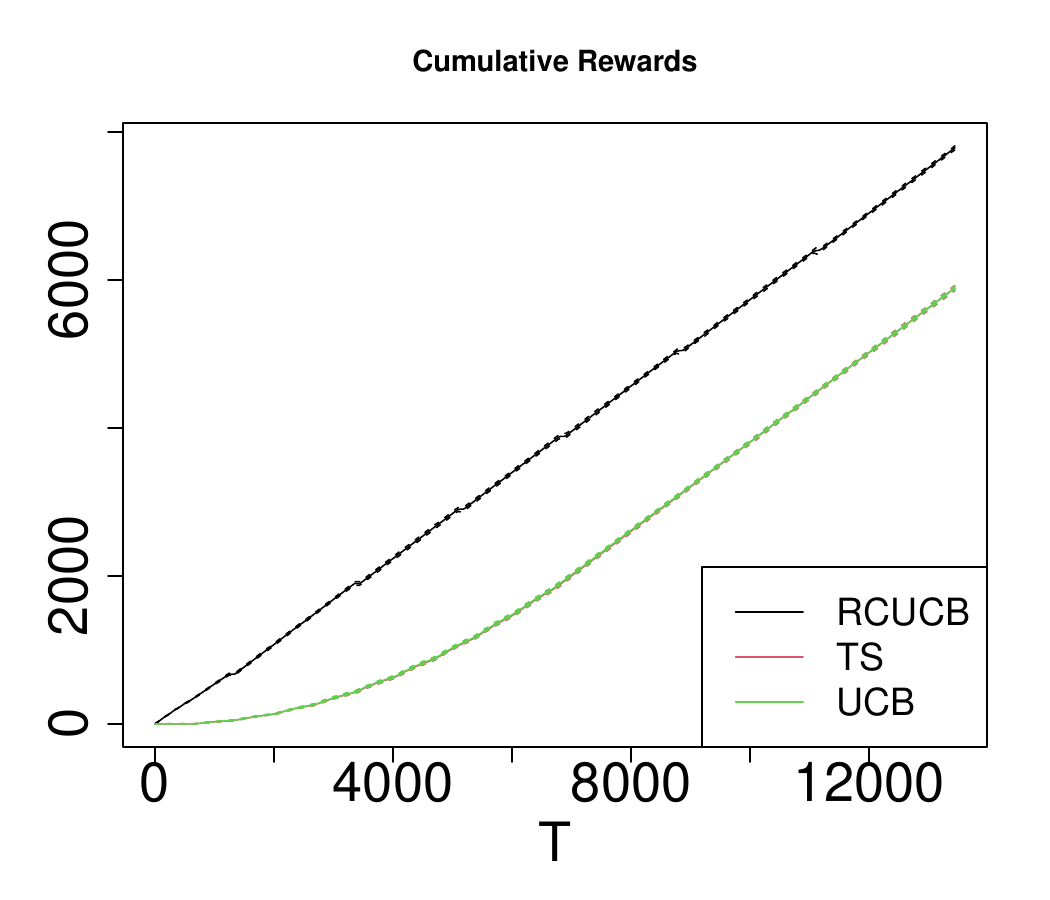}
		}
		\caption{Mean cumulative rewards for the algorithm configuration task of Random Forest on the AmesHousing data set. }
	\label{fig:RF_simu}
\end{figure}

\noindent The mean cumulative rewards over 10 repetitions of the algorithms is shown in Figure \ref{fig:RF_simu}, in which we see that the cumulative reward of $\texttt{RCUCB}$ exceeds the cumulative rewards of $\texttt{UCB}$ and $\texttt{TS}$ throughout.  
Note that both $\texttt{UCB}$ and $\texttt{TS}$ have almost the same mean cumulative rewards, so they are barely distinguishable in the plot.
The proportion of rounds, where the time limit for the training was exceeded, is $0.0589$ for $\texttt{RCUCB}$,  $0.2843$ for  $\texttt{TS}$, and $0.2853$ for $\texttt{UCB}.$ 
Again the obtained results are in favor of $\texttt{RCUCB}.$

\section{Related Work} \label{sec_related_work}

Various authors have considered bandit problems in which each arm is equipped with a multivariate distribution, i.e., in which the learner receives potentially vectorial type of feedback.
In the \emph{bandit problem with delayed feedback} \cite{joulani2013online,mandel2015queue,vernade2017stochastic,pike2018bandits}, each arm possesses a reward and a reward-time generation distribution, and rewards of previously chosen arms can be observed in a later round.
This is fundamentally different from our setting, where an arm's reward is only observable in the round it is played.
Moreover, the two distributions occurring in the bandits problem with delayed feedback are usually assumed to be independent, while we do not make an independence assumption on the reward and consumption of resources distribution of the arms.

Multivariate feedback of a played arm is at the core of the \emph{multi-objective multi-armed bandit} (MOMAB) problem.
Due to the possibly competing objectives encoded in the vectorial payoffs, different approaches have been considered to specify an optimal decision in the MOMAB problem.
Each objective is considered as a different multi-armed bandit problem in \cite{gabillon2011multi}, and the aim is to find the optimal arm for each objective separately.
In the majority of works, the Pareto front with respect to the mean vector is used to determine the optimality of an arm (see \cite{auer2016pareto} or \cite{drugan2019covariance} as well as references therein).
Finally, by aggregating the vectorial payoff by means of the generalized Gini index, a single objective to be optimized can again be obtained \cite{busa2017multi}.

In the \emph{bandits with knapsacks problem} \cite{badanidiyuru2013bandits,slivkins2019introduction,cayci2020budget}, each arm is associated with a reward and a (multivariate) resource consumption distribution as well. 
Although the original problem does not involve censoring rewards, variations of this problem have recently been considered in which the learner also has the option of setting a round-by-round limit on an arm's resource consumption that leads to censored rewards if exceeded \cite{cayci2019learning,sharoff2020farewell}.  
Nonetheless, the learning process in both the original bandits with knapsacks problem as well as the censored variant is substantially different from the one considered in this work: There is a predefined overall resource budget, which once exhausted leads to a termination of the entire learning process.  
This in turn leads to a different notion of cumulative regret and consequently different approaches regarding its theoretical analysis.

%
Censored feedback due to thresholding has been considered in \cite{abernethy2016threshold}, \cite{jain2018firing} and \cite{verma2019censored} within a bandit learning setting as well, albeit without multivariate distributions of the arms.
Also, the threshold values are either specified in each round by the environment or unknown but fixed among the arms, whereas in our setting, the learner chooses the threshold itself.
Resource allocation in a combinatorial bandit scenario has been the subject of research in \cite{lattimore2014optimal} and \cite{dagan2018better} as well as for a contextual variant in \cite{lattimore2015linear}.
However, in all these scenarios, the reward distributions are Bernoulli with a specific shape of the success probability depending on the allocated resources.

Bidding in online auctions \cite{cesa2014regret} is also concerned with choosing a suitable resource limit (reserve price for auctions) and obtaining possibly censored feedback from bidders.
This scenario is different from ours, as all available bidders are involved in a learning round (auction), while in our setting the learning algorithm has to pick one of the bidders in a metaphorical sense.

Ephemeral resource-constrained optimization problems (ERCOPs) are dealing with a dynamic constrained  optimization problem, in which both the objective function and the set of feasible solutions is static and certain time-dependent constraints may exist such that certain solutions may be temporarily unavailable for evaluation \cite{allmendinger2010line,allmendinger2011policy,allmendinger2013handling,allmendinger2015ephemeral}. 
These dynamic constraints are referred to as ephemeral resource constraints and account for a possible temporary unavailability of the resources needed to evaluate a solution. This results in an online learning problem similar to ours, where evaluations of solutions (an \emph{arm-resource pair} in our terminology) may be incomplete (or fail).
However, existing work on ERCOPs explicitly considers only scenarios where it can be checked a priori whether a solution will be incomplete (\emph{censored} in our terminology) without actually trying it. 
This leads to entirely different learning/optimization approaches than ours, as the evaluability of a solution is stochastic in our setting and consequently excludes such prechecks. 
Moreover, in our case, the focus is on a specific performance measure (regret) to evaluate an optimization strategy over time, which has significant implications for the design of appropriate strategies. 
For example, one optimization strategy for ERCOPs is to wait until a certain solution is available (or evaluable) again; a strategy that seems to be questionable with regard to cumulative regret minimization.

Finally, as we take the online algorithm selection problem as a running example for our setting, it is worth mentioning that bandit-based approaches have been already considered for this problem \cite{gagliolo2007learning,gagliolo2010algorithm,degroote2017online,online_as_main_degrooteCBK18,tornede2022machine}. 
However, these focus on certain algorithmic problem classes, such as the boolean satisfiability problem (SAT) or the quantified boolean formula problem (QBF). 
In particular, these works consider binary reward signals (the solver has solved/not solved the problem instance) and the runtimes of their respective solvers as the consumption of resources.    
Extending these approaches to more general frameworks like ours with continuous reward signals or other types of resource consumptions is far from a given.

\section{Conclusion and Future Work}        \label{sec_conclusion}

In this paper, we have introduced another variant of the classical multi-armed bandit problem, where attention is paid not only to the rewards themselves, but also to the resources consumed by an arm necessary to generate the rewards within each round of the sequential decision process.
The learner (bandit policy) is equipped with the ability to determine the resource limit of one round and might be willing to sacrifice optimality regarding the obtained rewards in order to keep the overall consumption of resources low, as this generates costs diminishing the overall gain.
As a consequence, the learner needs to find a good compromise between the two possibly conflicting targets, namely an as high as possible realizable reward on the one side, and an as low as possible consumption of resources for the reward generation on the other side.

To this end, we proposed a regret measure, which, by virtue of a cost and a penalty function, takes these two targets into account and allows for a suitable assessment of the expected gain with respect to the allocated resources.
By defining a suitable estimate of an arm's expected gain and its probability to exceed the allocated resources, we proposed optimistic bandit strategies for dealing with a finite or an infinite subset of available resource limits.

For future work, it would be interesting to extend the considered bandit problem to a combinatorial bandit setting, in which it is possible to choose a subset of arms in each round. 
Moreover, the very idea of incorporating resource constraints for the feedback generation process is not restricted to feedback of numerical nature, but could also be of interest for related bandit scenarios with other types of feedback, such as the preference-based multi-armed bandit problem \cite{busa2018preference}.
Last but not least, as the motivation of the considered type of bandit problem stems from practical applications, it would be of interest to investigate our algorithm for a variety of real-world problems, such as a more extensive simulation study on algorithm configuration \cite{schede2022AC}.

\medskip
	
\bibliographystyle{splncs04}
\bibliography{literature_shortened}

\begin{thebibliography}{10}
\providecommand{\url}[1]{\texttt{#1}}
\providecommand{\urlprefix}{URL }
\providecommand{\doi}[1]{https://doi.org/#1}

\bibitem{abe2003reinforcement}
Abe, N., Biermann, A., Long, P.: Reinforcement learning with immediate rewards
  and linear hypotheses. Algorithmica  \textbf{37}(4),  263--293 (2003)

\bibitem{abernethy2016threshold}
Abernethy, J., Amin, K., Zhu, R.: Threshold bandit, with and without censored
  feedback. In: {NeurIPS}. pp. 4896--4904 (2016)

\bibitem{agrawal2012analysis}
Agrawal, S., Goyal, N.: Analysis of {T}hompson sampling for the multi-armed
  bandit problem. In: {COLT}. pp. 1--39 (2012)

\bibitem{allmendinger2010line}
Allmendinger, R., Knowles, J.: On-line purchasing strategies for an
  evolutionary algorithm performing resource-constrained optimization. In:
  International Conference on Parallel Problem Solving from Nature. pp.
  161--170. Springer (2010)

\bibitem{allmendinger2011policy}
Allmendinger, R., Knowles, J.: Policy learning in resource-constrained
  optimization. In: {GECCO}. pp. 1971--1978 (2011)

\bibitem{allmendinger2013handling}
Allmendinger, R., Knowles, J.: On handling ephemeral resource constraints in
  evolutionary search. Evolutionary computation  \textbf{21}(3),  497--531
  (2013)

\bibitem{allmendinger2015ephemeral}
Allmendinger, R., Knowles, J.: Ephemeral resource constraints in optimization.
  In: Evolutionary Constrained Optimization, pp. 95--134. Springer (2015)

\bibitem{auer2002using}
Auer, P.: Using confidence bounds for exploitation-exploration trade-offs.
  Journal of Machine Learning Research  \textbf{3}(Nov),  397--422 (2002)

\bibitem{auer2002finite}
Auer, P., Cesa-Bianchi, N., Fischer, P.: Finite-time analysis of the multiarmed
  bandit problem. Machine Learning  \textbf{47}(2-3),  235--256 (2002)

\bibitem{auer2016pareto}
Auer, P., Chiang, C.K., Ortner, R., Drugan, M.: Pareto front identification
  from stochastic bandit feedback. In: {AISTATS}. pp. 939--947 (2016)

\bibitem{badanidiyuru2013bandits}
Badanidiyuru, A., Kleinberg, R., Slivkins, A.: Bandits with knapsacks. In:
  Annual Symposium on Foundations of Computer Science. pp. 207--216. IEEE
  (2013)

\bibitem{busa2018preference}
Bengs, V., Busa-Fekete, R., El~Mesaoudi-Paul, A., H\"{u}llermeier, E.:
  Preference-based online learning with dueling bandits: {A} survey. Journal of
  Machine Learning Research  \textbf{22}(7),  1--108 (2021)

\bibitem{bubeck2010bandits}
Bubeck, S.: Bandits Games and Clustering Foundations. Ph.D. thesis,
  Universit{\'e} des Sciences et Technologie de Lille-Lille I (2010)

\bibitem{bubeck2011x}
Bubeck, S., Munos, R., Stoltz, G., Szepesv{\'a}ri, C.: X-armed bandits. Journal
  of Machine Learning Research  \textbf{12}(5),  1655--1695 (2011)

\bibitem{busa2017multi}
Busa-Fekete, R., Sz{\"o}r{\'e}nyi, B., Weng, P., Mannor, S.: Multi-objective
  bandits: Optimizing the generalized {G}ini index. In: {ICML}. pp. 625--634
  (2017)

\bibitem{cayci2019learning}
Cayci, S., Eryilmaz, A., Srikant, R.: Learning to control renewal processes
  with bandit feedback. Proceedings of the ACM on Measurement and Analysis of
  Computing Systems  \textbf{3}(2),  1--32 (2019)

\bibitem{cayci2020budget}
Cayci, S., Eryilmaz, A., Srikant, R.: Budget-constrained bandits over general
  cost and reward distributions. In: {AISTATS}. pp. 4388--4398 (2020)

\bibitem{cesa2014regret}
Cesa-Bianchi, N., Gentile, C., Mansour, Y.: Regret minimization for reserve
  prices in second-price auctions. IEEE Transactions on Information Theory
  \textbf{61}(1),  549--564 (2014)

\bibitem{cesa2006prediction}
Cesa-Bianchi, N., Lugosi, G.: Prediction, Learning, and Games. Cambridge
  University Press (2006)

\bibitem{cesa2012combinatorial}
Cesa-Bianchi, N., Lugosi, G.: Combinatorial bandits. Journal of Computer and
  System Sciences  \textbf{78}(5),  1404--1422 (2012)

\bibitem{dagan2018better}
Dagan, Y., Koby, C.: A better resource allocation algorithm with semi-bandit
  feedback. In: {ALT}. pp. 268--320 (2018)

\bibitem{degroote2017online}
Degroote, H.: Online algorithm selection. In: {IJCAI}. pp. 5173--5174 (2017)

\bibitem{online_as_main_degrooteCBK18}
Degroote, H., Causmaecker, P.D., Bischl, B., Kotthoff, L.: A regression-based
  methodology for online algorithm selection. In: Proceedings of the Eleventh
  International Symposium on Combinatorial Search, {SOCS} 2018. pp. 37--45
  (2018)

\bibitem{drugan2019covariance}
Drugan, M.: Covariance matrix adaptation for multiobjective multiarmed bandits.
  IEEE Transactions on Neural Networks and Learning Systems  \textbf{30}(8),
  2493--2502 (2019)

\bibitem{gabillon2011multi}
Gabillon, V., Ghavamzadeh, M., Lazaric, A., Bubeck, S.: Multi-bandit best arm
  identification. In: {NeurIPS}. pp. 2222--2230 (2011)

\bibitem{gagliolo2007learning}
Gagliolo, M., Schmidhuber, J.: Learning restart strategies. In: {IJCAI}. pp.
  792--797 (2007)

\bibitem{gagliolo2010algorithm}
Gagliolo, M., Schmidhuber, J.: Algorithm selection as a bandit problem with
  unbounded losses. In: International Conference on Learning and Intelligent
  Optimization {(LION)}. pp. 82--96. Springer (2010)

\bibitem{grill2015black}
Grill, J.B., Valko, M., Munos, R.: Black-box optimization of noisy functions
  with unknown smoothness. In: {NeurIPS}. pp. 667--675 (2015)

\bibitem{hutter2019automated}
Hutter, F., Kotthoff, L., Vanschoren, J.: Automated Machine Learning: Methods,
  Systems, Challenges. Springer Nature (2019)

\bibitem{jain2018firing}
Jain, L., Jamieson, K.: Firing bandits: Optimizing crowdfunding. In: {ICML}.
  pp. 2206--2214 (2018)

\bibitem{joulani2013online}
Joulani, P., Gy{\"o}rgy, A., Szepesv{\'a}ri, C.: Online learning under delayed
  feedback. In: {ICML}. pp. 1453--1461 (2013)

\bibitem{kerschke2019automated}
Kerschke, P., Hoos, H., Neumann, F., Trautmann, H.: Automated algorithm
  selection: Survey and perspectives. Evolutionary Computation  \textbf{27}(1),
   3--45 (2019)

\bibitem{kleinberg2008multi}
Kleinberg, R., Slivkins, A., Upfal, E.: Multi-armed bandits in metric spaces.
  In: Proceedings of the fortieth annual ACM symposium on Theory of computing.
  pp. 681--690 (2008)

\bibitem{lattimore2014optimal}
Lattimore, T., Crammer, K., Szepesv{\'a}ri, C.: Optimal resource allocation
  with semi-bandit feedback. In: {UAI}. pp. 477--486 (2014)

\bibitem{lattimore2015linear}
Lattimore, T., Crammer, K., Szepesv{\'a}ri, C.: Linear multi-resource
  allocation with semi-bandit feedback. In: {NeurIPS}. pp. 964--972 (2015)

\bibitem{lattimore2020bandit}
Lattimore, T., Szepesv{\'a}ri, C.: Bandit Algorithms. Cambridge University
  Press (2020)

\bibitem{mandel2015queue}
Mandel, T., Liu, Y.E., Brunskill, E., Popovi{\'c}, Z.: The queue method:
  Handling delay, heuristics, prior data, and evaluation in bandits. In:
  {AAAI}. pp. 2849--2856 (2015)

\bibitem{munos2014bandits}
Munos, R.: From bandits to {Monte-Carlo} tree search: {T}he optimistic
  principle applied to optimization and planning. Foundations and
  Trends{\textregistered} in Machine Learning  \textbf{7}(1),  1--129 (2014)

\bibitem{pike2018bandits}
Pike-Burke, C., Agrawal, S., Szepesvari, C., Grunewalder, S.: Bandits with
  delayed, aggregated anonymous feedback. In: {ICML}. pp. 4105--4113 (2018)

\bibitem{schede2022AC}
Schede, E., Brandt, J., Tornede, A., Wever, M., Bengs, V., H{\"{u}}llermeier,
  E., Tierney, K.: A survey of methods for automated algorithm configuration.
  Journal of Artificial Intelligence Research  \textbf{75} (2022)

\bibitem{sharoff2020farewell}
Sharoff, P., Mehta, N., Ganti, R.: A farewell to arms: Sequential reward
  maximization on a budget with a giving up option. In: {AISTATS}. pp.
  3707--3716 (2020)

\bibitem{slivkins2019introduction}
Slivkins, A.: Introduction to multi-armed bandits. Foundations and
  Trends{\textregistered} in Machine Learning  \textbf{12}(1-2),  1--286 (2019)

\bibitem{tornede2022machine}
Tornede, A., Bengs, V., H{\"u}llermeier, E.: Machine learning for online
  algorithm selection under censored feedback. In: {AAAI}. vol.~36, pp.
  10370--10380 (2022)

\bibitem{traca2021regulating}
Trac{\`a}, S., Rudin, C.: Regulating greed over time in multi-armed bandits.
  Journal of Machine Learning Research  \textbf{22}(3),  1--99 (2021)

\bibitem{verma2019censored}
Verma, A., Hanawal, M., Rajkumar, A., Sankaran, R.: Censored semi-bandits: A
  framework for resource allocation with censored feedback. In: {NeurIPS}. pp.
  14526--14536 (2019)

\bibitem{vernade2017stochastic}
Vernade, C., Capp{\'e}, O., Perchet, V.: Stochastic bandit models for delayed
  conversions. In: {UAI} (2017)

\bibitem{yue2009interactively}
Yue, Y., Joachims, T.: Interactively optimizing information retrieval systems
  as a dueling bandits problem. In: {ICML}. pp. 1201--1208 (2009)

\end{thebibliography}

\newpage

\newpage
\appendix

\thispagestyle{empty}
\begin{center}
{\Large\textbf{Supplementary material to \\"Multi-Armed Bandits with Censored Consumption of Resources''}  }
\end{center}

\section*{List of Symbols}
The following table contains a list of symbols that are frequently used in the main paper. \\ \medskip

\small
\begin{tabular}{l|l}
$\mathbb N$ & set of natural numbers $1,2,\dots$ without $0$\\
$\mathbb N_0$ & set of natural numbers $1,2,\dots$ including $0$\\
$\IND{A}$ & indicator function on the set $A$\\	
$T$ & (unknown) number of overall rounds (element of $\mathbb N$)\\
$T^+$ & (unknown) number of overall rounds incremented by one (i.e., $T+1$)\\
$\tau_{\mathrm{max}}$ & maximal possible resource limit \\
$\mathcal{M}$ & a subset of the admissible resource range of a round $(0,\tau_{\mathrm{max}}]$  \\
$\mathcal{M}^{(i)}_t$ & set of grid points in $(0,\tau_{\mathrm{max}}]$ maintained by $\texttt{z-RCUCB}$ for arm $i\in[n]$ in round $t$ \\
$\tau$ & resource limit of a round (element of $\mathcal{M}$)\\
$\up(\tautemp)$ & next larger resource limit of a resource limit $\tau \neq \tau_{\mathrm{max}}$ in $\mathcal{M}_t^{(i)}$ \\
$\low_i(\tautemp)$ & next smaller resource limit of a resource limit in $\mathcal{M}_t^{(i)}$  \\
$[n]$ & the set of arms $\{1,2,\dots,n\}$  ($n$ is an element of $\mathbb{N}$ greater than 1) \\
$P_i^{(r)},P_i^{(c)}$ & reward and resource consumption distribution of an arm $i\in[n]$ \\
$P_i^{(r,c)}$ & joint distribution of an arm's reward and resource consumption \\
$F_i^{(c)}$ & cumulative distribution function of $P_i^{(c)}$ \\
$R_{i,t},C_{i,t}$ & reward and resource consumption of arm $i\in[n]$ in round $t\in \{1,\ldots,T\}$  \\
$I_t$ & chosen arm in round $t\in \{1,\ldots,T\}$ (element of $[n]$)	\\
$\tau_t$ & chosen resource limit in round $t\in \{1,\ldots,T\}$ (element of $\mathcal{M}$) \\
$c(\cdot)$ & cost function ($c:\R_+ \to [0,1]$) specifying the cost generated by the consumed resources \\
$\lambda(\cdot)$ & penalty function ($\lambda: \mathcal{M} \to \R_+$)  specifying the  penalty for exceeding the allocated resources \\
$\nu_{i,\tautemp}$ & penalized expected gain of the pair $(i,\tautemp)\in[n]\times \mathcal{M}$ (see \eqref{def_nu_i_tau}) \\
$\nu^*$ &  optimal penalized expected gain \\
$(i^*,\tau^*)$ &  optimal arm/resource-limit pair (see \eqref{def_max_problem}) \\
$\Delta_{i,\tautemp}$ & sub-optimality gap of the pair $(i,\tautemp)\in[n]\times \mathcal{M}$ (see \eqref{def_sub_opt_gap}) \\
$r_t $ & instantaneous regret in round $t$ (see line above \eqref{regret_def}) \\ 
$\mathcal R_T$ & cumulative regret over $T$ number of rounds (see \eqref{regret_def})\\
$ T_{i,\tautemp}(t)$ & no.\ of rounds till $t,$ in which $(i,\tautemp)$ was chosen \\
$ N_{i,\tautemp}(t)$ & no.\ of rounds till $t,$ in which $\tautemp$ was below the predefined resource limit, while $i$ was chosen \\
$g_{i,\tautemp}$ &  expected gain of the pair $(i,\tautemp)\in[n]\times \mathcal{M}$ in round $t$  (see two displays above \eqref{def_emp_gain_estimate}) \\
$\Lambda_{i,\tautemp}$ &  expected penalty term of the pair $(i,\tautemp)\in[n]\times \mathcal{M}$ in round $t$ (see display above  \eqref{def_emp_gain_estimate}) \\
$\hat g_{i,\tautemp}$ &  expected gain estimate of the pair $(i,\tautemp)\in[n]\times \mathcal{M}$ in round $t$  (see \eqref{def_emp_gain_estimate}) \\
$\hat \Lambda_{i,\tautemp}$ &  expected penalty term estimate of the pair $(i,\tautemp)\in[n]\times \mathcal{M}$ in round $t$ (see \eqref{def_emp_penalty_estimate}) \\
$\hat \nu_{i,\tautemp}(t)$ & penalized expected gain estimate of the pair $(i,\tautemp)\in[n]\times \mathcal{M}$ in round $t$ (see \eqref{def_exp_gain_estimate}) \\
$\mathfrak c_{i,\tautemp}^{(g)}(t;\alpha) $ &  confidence interval length of $\hat g_{i,\tautemp}(t)$  (see display above Proposition \ref{prop_nu_estimates})  \\
$\mathfrak c_{i,\tautemp}^{(\Lambda)}(t;\alpha) $ &  confidence interval length of $\hat \Lambda_{i,\tautemp}(t)$  (see display above  Proposition \ref{prop_nu_estimates})  \\
$\mathfrak c_{i,\tautemp}(t;\alpha)$ &  confidence interval length of $\hat \nu_{i,\tautemp}(t)$  (see display above  Proposition \ref{prop_nu_estimates})  \\
\end{tabular}
\normalsize

\newpage

\section{Proof of Proposition \ref{prop_nu_estimates}: Concentration Inequalities for the Penalized Expected Gain Estimates} \label{sec_appendix_proof_concentration}

A key result for the proof of Proposition \ref{prop_nu_estimates} is the following lemma.

\begin{lemma}\label{lemma_concentration_ineq}
Let $(i,\tautemp) \in [n]\times \mathcal{M}.$ For any round  $t \in [T],$ the following holds for any $N \in \{0,1,\ldots,t\}$ and any $\varepsilon>0:$
\begin{equation*}
\begin{split} 
%
&\mathbb{P}\big( \exists s \in \{0,1,\ldots,N\}: Z_{i,\tautemp}^{(g)}(t) > \varepsilon , \, N_{i,\tautemp}(t) = s \big) 
\leq \exp\left(  -  \varepsilon^2/ (2N) \right), \\
&\mathbb{P}\big(  \exists s \in \{0,1,\ldots,N\}:  Z_{i,\tautemp}^{(g)}(t) < - \varepsilon , \, N_{i,\tautemp}(t) = s \big)
\leq \exp\left(  -  \varepsilon^2/ (2N) \right), 
\end{split} 
%
\end{equation*} 
where $ Z_{i,\tautemp}^{(g)}(t) = N_{i,\tautemp}(t) (\hat g_{i,\tautemp}(t) -  g_{i,\tautemp}).$

Moreover, for any round  $t \in [T],$ the following holds for any $N \in \{0,1,\ldots,t\}$ and any $\varepsilon>0:$
\begin{equation*}
\begin{split} 
%
&\mathbb{P}\big( \exists s \in \{0,1,\ldots,N\}: Z_{i,\tautemp}^{(\Lambda)}(t) > \varepsilon , \, N_{i,0}(t) = s \big) 
\leq \exp\left(  -  (\lambda(\tautemp) \, \varepsilon)^2/ (2N) \right), \\
%
&\mathbb{P}\big(  \exists s \in \{0,1,\ldots,N\}:  Z_{i,\tautemp}^{(\Lambda)}(t) < - \varepsilon , \, N_{i,0}(t) = s \big)
\leq \exp\left(  -   (\lambda(\tautemp) \, \varepsilon)^2/ (2N) \right), 
\end{split} 
%
\end{equation*} 
where $ Z_{i,\tautemp}^{(\Lambda)}(t) = N_{i,0}(t) (\hat \Lambda_{i,\tautemp}(t) -  \Lambda_{i,\tautemp}).$

\end{lemma}

\begin{proof}
We start by showing the claim for $Z_{i,\tautemp}^{(g)}(t).$
Fix some round $t\in \{1,\ldots,T\}$ and denote by $t_j$ the round at which for the $j$-th time $I_s=i$ and $\tau_s \geq \tautemp$ was observed, where $j \in \{0,1,\ldots,N_{i,\tautemp}(t)\}.$
Next, define the martingale sequence 
\begin{align*}
Y_0 &:= 0, \\ 
Y_j &:=\sum_{k=1}^j \Big( (  R_{I_{t_k},t_k}  - c(  C_{I_{t_k},t_k}) ) \cdot \IND{\{ C_{I_{t_k},t_k}\leq \tautemp  \} } \cdot \IND{\{  I_{t_k}=i \, \wedge \, \tau_{t_k} \geq \tautemp \}	}   - g_{i,\tautemp} \Big),  
\end{align*}
where $ j \in \{1,\ldots,N_{i,\tautemp}(t)\}.$
This is indeed a martingale sequence, as for any $j \in \{1,\ldots,N_{i,\tautemp}(t)\}$ we obtain by abbreviating $Y_0,\ldots,Y_{j-1}$ via $Y_{0:j-1}$ that
\begin{align*}
\mathbb{E} &\big(		Y_j	\, \big|  \, Y_{0:j-1} \big) \\
&= \mathbb{E}\big(		Y_j	\, \big|   Y_{0:j-1} \big) \\
&= \mathbb{E}\big(		Y_{j-1}    +  ( R_{I_{t_j},t_j} - c( C_{I_{t_j},t_j})) \cdot \IND{\{ C_{I_{t_j},t_j}\leq \tautemp  \} } \cdot \IND{\{  I_{t_j}=i \, \wedge \, \tau_{t_j} \geq \tautemp \}	} - g_{i,\tautemp}  	\, \big|   Y_{0:j-1}  \big) \\
&= Y_{j-1}    +   \mathbb{E}\big(	( R_{I_{t_j},t_j} - c( C_{I_{t_j},t_j})) \cdot \IND{\{ C_{i,t_j}\leq \tautemp  \} }  	\, \big|   Y_{0:j-1}  \big) - g_{i,\tautemp}  \\
&= Y_{j-1}    +   \mathbb E_{ (X_i^{(r)} , X_i^{(c)} )\sim  P_i^{(r,c)} } \big( (X_i^{(r)}  -c(X_i^{(c)} ))\cdot \IND{ \{X_i^{(c)} \leq \tautemp\}}  	\, \big|  Y_{0:j-1}  \big) - g_{i,\tautemp}  \\
&= Y_{j-1},     
\end{align*}
where we used that by definition of $t_j$ it holds that $ \IND{ \{  I_{t_j}=i \, \wedge \, \tau_{t_j} \geq \tautemp \} } =1$ and for the last equality that  $(R_{i,s},C_{i,s})_{s=1,\ldots,t}$ is an iid sequence.
Further, it holds for any $j \in \{1,\ldots,N_{i,\tautemp}(t)\}$ that $|Y_j - Y_{j-1}|\leq 1,$ since $R_{i,t_j},c(C_{i,t_j}),g_{i,\tautemp}\in[0,1]$ as well as $\IND{\{ C_{i,t_j}\leq \tautemp  \} } \in \{0,1\}.$
An application of Hoeffding's maximal inequality (see Lemma A.7 in \cite{cesa2006prediction}) leads to
\begin{align*}
&\mathbb{P}( \exists s \in \{0,1,2,\ldots, N\} : Y_{N_{i,\tautemp}(t)} - Y_0 > \varepsilon \, , \, N_{i,\tautemp}(t) = s) \leq \exp\left( -  \frac{\varepsilon^2}{2N} \right), \\
&\mathbb{P}( \exists s \in \{0,1,2,\ldots, N\} : Y_{N_{i,\tautemp}(t)}  -Y_0 < - \varepsilon \, , \, N_{i,\tautemp}(t) = s) \leq \exp\left(  - \frac{\varepsilon^2}{2N} \right).
\end{align*}
Since $ Y_{N_{i,\tautemp}(t)}  - Y_0 = Z_{i,\tautemp}^{(g)}(t)$ we can conclude the first part of the lemma.
%
%
For the second part, we now define the martingale sequence 
\begin{align*}
Y_0 &:= 0, \\ 
Y_j &:=\sum_{k=1}^j \Big( \IND{\{ C_{I_{t_k},t_k} > \tautemp  \} } \cdot \IND{\{  I_{t_k}=i \}	}   -  (1-P_i^{(c)}(\tau)) \Big),   \quad j \in \{1,\ldots,N_{i,0}(t)\}.
\end{align*}
Here, $t_k$ now denotes the round at which for the $k$-th time $I_s=i$  was observed, where $k \in \{0,1,\ldots,N_{i,0}(t)\}.$
One can easily verify that this is indeed a martingale sequence with $|Y_j - Y_{j-1}|\leq 1$ similarly as in the first part.
Noting that  $ \lambda(\tautemp) (Y_{N_{i,0}(t)}  - Y_0) = Z_{i,\tautemp}^{(\Lambda)}(t)$ and using once again  Hoeffding's maximal inequality leads to the second part of the lemma.
\end{proof}
For the sake of convenience, we restate Proposition \ref{prop_nu_estimates} prior to giving its proof.
\setcounter{proposition}{0}
\begin{proposition}  
%
%
Let $(i,\tautemp) \in [n]\times \mathcal{M}$ and $\alpha>1.$
Then, for any round $t \geq n+1$, it holds that
\begin{align*}
\mathbb{P}\big( \hat \nu_{i,\tautemp}(t) -  \nu_{i,\tautemp} > \mathfrak c_{i,\tautemp}(t;\alpha)  \big) 
&\leq 2 \left(1+\frac{\log(t)}{\log(\frac{\alpha+1}{2})} \right) t^{-\frac{2\alpha}{\alpha+1}}, 
\end{align*}
and 	
the right-hand side is also an upper bound for  $\mathbb{P}\big( \hat \nu_{i,\tautemp}(t) -  \nu_{i,\tautemp} < - \mathfrak c_{i,\tautemp}(t;\alpha) \big),$ where 
\begin{align*}
\mathfrak c_{i,\tautemp}(t;\alpha) 
&= \mathfrak c_{i,\tautemp}^{(g)}(t;\alpha) + \mathfrak c_{i,\tautemp}^{(\Lambda)}(t;\alpha) \\
&= \sqrt{ (2 \alpha \log(t))/ N_{i,\tautemp}(t)}   + \lambda(\tau) \sqrt{ (2 \alpha \log(t))/ N_{i,0}(t)}. 
\end{align*}
\end{proposition}

\begin{proof}
Let $Z_{i,\tautemp}^{(g)}(t)$ and  $Z_{i,\tautemp}^{(\Lambda)}(t)$ be as in Lemma \ref{lemma_concentration_ineq}, then
%
\begin{align*}
\mathbb{P}\big( &\hat \nu_{i,\tautemp}(t) -  \nu_{i,\tautemp} > \mathfrak c_{i,\tautemp}(t;\alpha) \big) \\
&\leq 	\mathbb{P}\Big( \hat g_{i,\tautemp}(t) -  g_{i,\tautemp} > \sqrt{ \frac{2 \alpha \log(t)}{N_{i,\tautemp}(t)}  }  \Big) 
+ 	\mathbb{P}\Big( \hat \Lambda_{i,\tautemp}(t) -  \Lambda_{i,\tautemp} < - \lambda(\tau) \sqrt{ \frac{2 \alpha \log(t) }{N_{i,0}(t)} } \Big) \\
&\leq 	\mathbb{P}\big( \exists s \in \{1,\ldots,t\} : Z_{i,\tautemp}^{(g)}(t)  >   \sqrt{2 \alpha  N_{i,\tautemp}(t) \log(t) 	 } \, , \, N_{i,\tautemp}(t)=s  \big) \\
&\quad + \mathbb{P}\big( \exists s \in \{1,\ldots,t\} : Z_{i,\tautemp}^{(\Lambda)}(t)  < - \lambda(\tautemp)   \sqrt{2 \alpha  N_{i,0}(t) \log(t) 	 } \, , \, N_{i,0}(t)=s  \big) \\
&= \mathbb{P}\big( \exists s \in \{1,\ldots,t\} :   Z_{i,\tautemp}^{(g)}(t)   >   \sqrt{ 2  \alpha s \log(t)  } \, , \, N_{i,\tautemp}(t)=s  \big) \\
&\quad + \mathbb{P}\big( \exists s \in \{1,\ldots,t\} : Z_{i,\tautemp}^{(\Lambda)}(t)  < - \lambda(\tautemp)   \sqrt{2 \alpha s \log(t) 	 } \, , \, N_{i,0}(t)=s  \big).
\end{align*}
Next, we use a peeling argument to show that the first term on the right-hand side of the latter display is bounded by $\left(1+\frac{\log(t)}{\log(\frac{\alpha+1}{2})} \right) t^{-\frac{2\alpha}{\alpha+1}}.$
Define $\beta = \frac{2}{\alpha+1}$ and note that $\beta \in(0,1),$ since $\alpha>1.$
For each $s\in\{1,\ldots,t\}$ there exists some $j \in \{0,1,\ldots,D_{t,\beta} \}$ with $\beta^{j+1} t < s \leq \beta^j t,$ where $D_{\beta,t} = \frac{\log(t)}{\log(1/\beta)}.$
Thus,
\begin{align*}
\mathbb{P}\big( &\exists s \in \{1,\ldots,t\} :   Z_{i,\tautemp}^{(g)}(t)   >   \sqrt{ 2  \alpha s \log(t)  } \, , \, N_{i,\tautemp}(t)=s  \big)	\\
&\leq  \sum_{j=0}^{D_{\beta,t}} \mathbb{P}\big( \exists s \in (\beta^{j+1} t , \beta^j t ] \cap \mathbb{N}:  Z_{i,\tautemp}^{(g)}(t)   >   \sqrt{ 2  \alpha s \log(t)  } \, , \, N_{i,\tautemp}(t)=s \big)	\\
&\leq  \sum_{j=0}^{D_{\beta,t}} \mathbb{P}\big( \exists s \in (\beta^{j+1} t , \beta^j t ] \cap \mathbb{N}:   \ Z_{i,\tautemp}^{(g)}(t)    >   \sqrt{ 2  \alpha \beta^{j+1} t \log(t)  } \, , \, N_{i,\tautemp}(t)=s \big)	\\
&\leq  \sum_{j=0}^{D_{\beta,t}} \exp\left( - \alpha \beta \log(t)  \right) 
= \left(1+\frac{\log(t)}{\log(\frac{\alpha+1}{2})} \right) t^{-\frac{2\alpha}{\alpha+1}}, 
\end{align*}    
where we used Lemma \ref{lemma_concentration_ineq} for the last inequality with $N= \lfloor \beta^j t \rfloor$ and $\varepsilon =  \sqrt{ 2  \alpha \beta^{j+1} t \log(t)  }$.
An analogous argumentation using the concentration inequalities for $Z_{i,\tautemp}^{(\Lambda)}$ in Lemma \ref{lemma_concentration_ineq}  shows that
\begin{align*}
\mathbb{P}\big( \exists s \in \{1,\ldots,t\} : Z_{i,\tautemp}^{(\Lambda)}(t) &< - \lambda(\tautemp)   \sqrt{2 \alpha s \log(t) 	 } \, , \, N_{i,0}(t)=s  \big) 	\\
&\leq \left(1+\frac{\log(t)}{\log(\frac{\alpha+1}{2})} \right) t^{-\frac{2\alpha}{\alpha+1}}.
\end{align*}
Finally, the inequality for $\mathbb{P}\big( \hat \nu_{i,\tautemp}(t) -  \nu_{i,\tautemp} < - \mathfrak c_{i,\tautemp}(t;\alpha) \big)$ can be proved similarly by using the two inequalities in Lemma \ref{lemma_concentration_ineq} that have not been used so far.
\end{proof}

\section{Proof of Theorem \ref{theorem_regret_upper_bound}: Regret Bound of \texttt{RCUCB}} \label{sec_appendix_proof_regret_upper_bound}

For sake of convenience, we restate Theorem \ref{theorem_regret_upper_bound} here again, but with the explicit form of the constant $C_\alpha.$

\setcounter{theorem}{0}
\begin{theorem}
Let $\alpha>1$ in \eqref{RCUCB_choice}. 
Then, for any number of rounds $T,$ $\varepsilon\in(0,1),$ and any $\delta \in(0,1/2)$ such that $ (n|\mathcal{M}|-1)^{1-\delta} T^{2\delta} \leq T$,  it holds that
$$\mathcal R_T^{\texttt{RCUCB}} \leq \sum\limits_{(i,   \tautemp) \in [n] \times \mathcal{M}} \Delta_{i,\tau} \, u_{i,\tautemp}(T,\alpha) - \mathbb{P}(A_\varepsilon) \sum\limits_{(i,\tau) \in [n]\times \mathcal{M}\backslash\{\tau_{\mathrm{max}}\} } \Delta_{i,\tau} \, l_{i,\up(\tautemp)}(T,\alpha),$$ 
where 
\begin{align*}
u_{i,\tautemp}(T,\alpha) &:= \frac{8 \alpha  (1+\lambda(\tautemp))^2 \log(T)}{\Delta_{i,\tautemp}^2 } 
+ 1 + \frac{8}{\log\left(\frac{\alpha+1}{2}\right)} \left( \frac{\alpha+1}{\alpha-1} \right)^2, 
%
\\
l_{i,\up(\tautemp)}(T,\alpha)  &:= \frac{  \alpha \, \varepsilon \, \delta \,  \log(T)}{   H_{i,\up(\tautemp)}(\alpha) }, \\
A_\varepsilon&:= \bigcap\limits_{ (i,   \tautemp) \in [n] \times \mathcal{M}, \ t \in [T]} A_{i,\tau,t,\varepsilon},\\
A_{i,\tau,t,\varepsilon} &=  \{	\hat \nu_{i,\tautemp}(t) + (1-\varepsilon) \,  \mathfrak c_{i,\tautemp}(t;\alpha) \geq   \nu_{i,\tautemp} \geq \hat \nu_{i,\tautemp}(t) -   \mathfrak c_{i,\tautemp}(t;\alpha) \},
\end{align*}
and $H_{i,\tau}(\alpha) := \max_{(j, \tau' ) \in [n]\times \mathcal M: j \neq i \vee \tau' \neq \tau} \Big(	\frac{8(1+\lambda(\tau')^2 \alpha)}{\Delta_{i,\tau}-\Delta_{j,\tau'}}		+1 \Big)^2  (\Delta_{i,\tau}-\Delta_{j,\tau'})^2.$	
\end{theorem}

\begin{proof}
We consider the refined regret decomposition in \eqref{eq_regret_decomp_refined}, i.e.,
\begin{align*}
\mathcal R_T 
&= \sum_{\stackrel{\tautemp \in  \mathcal{M}\backslash\{\tau_{\mathrm{max}}\} }{i \in [n]} } \Delta_{i,\tautemp} \big(   \mathbb E(N_{i,\tautemp}(T^+))  - \mathbb E(N_{i,\up(\tautemp)}(T^+))  \big) 
\\
&\quad + \sum\nolimits_{i \in [n] } \Delta_{i,\tau_{\mathrm{max}}} \mathbb E(T_{i,\tau_{\mathrm{max}}}(T^+)),
\end{align*}
and split the proof into two parts: In the first part we show that 
\begin{align*}
&\mathbb E(N_{i,\tautemp}(T^+)) \leq u_{i,\tautemp}(T,\alpha) \qquad   \forall (i,\tautemp) \in  [n] \times \mathcal{M}\backslash\{\tau_{\mathrm{max}}\}, \mbox{and} \\
&\mathbb E(T_{i,\tau_{\mathrm{max}}}(T^+)) \leq u_{i,\tau_{\mathrm{max}}}(T,\alpha)
\end{align*}
In the second part, we show that 
$$ \mathbb E(N_{i,\up(\tautemp)}(T^+))  \geq \mathbb{P}(A_{\epsilon}) \, l_{i,\up(\tautemp)}(T,\alpha), \quad \forall (i,\tautemp) \in  [n] \times \mathcal{M}\backslash\{\tau_{\mathrm{max}}\}, $$
where our approach for the second part is inspired by \cite{traca2021regulating}. 
Combining both parts in the refined refined regret decomposition leads to the claim of this theorem.

\noindent\textbf{First part: Upper bound by the $u_{i,\tautemp}$ terms}

\noindent The pair $(i,\tautemp) \in [n] \times \mathcal{M}$ is chosen in round $t\in\{n+1,\ldots,T\}$ if
\begin{align} \label{ineq_regret_upper_aux}
(i,\tautemp) \in \argmax_{ (j,   \tautemp') \in [n] \times \mathcal{M}} \hat \nu_{j,\tautemp'}(t) + \mathfrak c_{j,\tautemp'}(t;\alpha).
%
\end{align}
Let us define the following ``bad'' events:
\begin{align*}
B_{t,1} := \{  \hat \nu_{i,\tautemp}(t) - \mathfrak c_{i,\tautemp}(t;\alpha)  > \nu_{i,\tautemp}    \}, \quad
B_{t,2} := \{  \hat \nu_{i^*,\tau^*}(t) + \mathfrak c_{i^*,\tau^*}(t;\alpha)  < \nu_{i^*,\tau^*}    \},
\end{align*}
where $(i^*,\tau^*)$ is the optimal pair (cf.\ discussion below \eqref{regret_def}).
Now, if $B_{t,1}^\complement$ holds, then \eqref{ineq_regret_upper_aux} implies
\begin{align*}
\nu_{i,\tautemp} + 2 \, \mathfrak c_{i,\tautemp}(t;\alpha) \geq \hat \nu_{i^*,\tau^*}(t) + \mathfrak c_{i^*,\tau^*}(t;\alpha).
\end{align*}
If additionally $B_{t,2}^\complement$ holds, then the latter implies
\begin{align*}
\nu_{i,\tautemp}  + 2 \, \mathfrak c_{i,\tautemp}(t;\alpha) \geq  \nu_{i^*,\tau^*}.
\end{align*}
This in turn  implies $N_{i,\tautemp}(t) \leq \frac{8 \alpha  (1+\lambda(\tautemp))^2 \log(t)  }{\Delta_{i,\tautemp}^2}.$
Thus, if $i\neq i^*$ and $\tautemp \neq \tau^*$ as well as $B_{t,1}^\complement \cap B_{t,2}^\complement$ holds, then it follows that
$N_{i,\tautemp}(T) \leq \frac{8 \alpha  (1+\lambda(\tautemp))^2 \log(T)  }{\Delta_{i,\tautemp}^2}.$
Next, by Proposition \ref{prop_nu_estimates} and the union bound we obtain
$$\mathbb{P}(	B_{t,1} \cup B_{t,2}) \leq  4 \left(1+\frac{\log(t)}{\log(\frac{\alpha+1}{2})} \right) t^{-\frac{2\alpha}{\alpha+1}}.$$
With these considerations, we can infer that
\begin{align*}
\mathbb{E}&(N_{i,\tautemp}(T^+)) \\
&\leq \frac{8 \alpha  (1+\lambda(\tautemp))^2 \log(T)  }{\Delta_{i,\tautemp}^2} +1 + \sum_{t = \lceil  \frac{8 \alpha  (1+\lambda(\tautemp))^2 \log(T)  }{\Delta_{i,\tautemp}^2} \rceil}^{T} \mathbb{P} (	B_{t,1} \cup B_{t,2}) \\
&\leq \frac{8 \alpha  (1+\lambda(\tautemp))^2 \log(T)  }{\Delta_{i,\tautemp}^2} +1 + \frac{4}{\log(\frac{\alpha+1}{2})}  \sum_{t = 2}^{T} \left( \log\left(\frac{\alpha+1}{2}\right) +\log(t)\right) t^{-\frac{2\alpha}{\alpha+1}}\\
&\leq \frac{8 \alpha  (1+\lambda(\tautemp))^2 \log(T)  }{\Delta_{i,\tautemp}^2} +1 + \frac{4}{\log(\frac{\alpha+1}{2})}  \int_{1}^\infty \left( \log\left(\frac{\alpha+1}{2}\right) +\log(x)\right) x^{-\frac{2\alpha}{\alpha+1}} \, \mathrm{d}x \\
&\leq \frac{8 \alpha  (1+\lambda(\tautemp))^2 \log(T)  }{\Delta_{i,\tautemp}^2} + 1 + \frac{8}{\log\left(\frac{\alpha+1}{2}\right)} \left( \frac{\alpha+1}{\alpha-1} \right)^2 = u_{i,\tautemp}(T,\alpha),
\end{align*}		
where we used in the second last line that $\sum_{t=2}^{T} \log(t)/t^{c} \leq \int_{1}^{\infty} \log(x)/ x^{c} \, \mathrm{d}x$ for any $c>1,$ and for the last line that
$$  \int_{1}^{\infty} \frac{\log(x)}{x^{c}}  \, \mathrm{d}x = \frac{1}{(c-1)^2},$$
which can be seen by integration by parts.
Note that $T_{i,\tau_{\mathrm{max}}}(\cdot)  =N_{i,\tau_{\mathrm{max}}}(\cdot)$ for any $i\in[n],$ so we have also from the above that $\mathbb E(T_{i,\tau_{\mathrm{max}}}(T^+)) \leq u_{i,\tau_{\mathrm{max}}}(T,\alpha).$\\

\noindent \textbf{Second part: Lower bound by the $l_{i,\up(\tautemp)}$ terms}

\noindent For the sake of convenience, we define for any 	$(i,\tau)\in [n] \times \mathcal{M}$ the set
$$  \mathcal{M}_{n}(i,\tau):= \{	(j,\tau')\in [n] \times \mathcal{M} \, | \, j\neq i \vee \tau' \neq \tau  \}.	$$
For some arbitrary round $t\in[T,]$ and conditioned on the event $A_\varepsilon,$ we have that 
\begin{align} \label{key_ineq}
\begin{split}
\Delta_{j,\tau'}	-\Delta_{i,\tau} +  &\varepsilon \sqrt{\frac{2 \alpha \log(t)}{N_{i,\tau}(t)}  }  \\
& > 2 \sqrt{2 \alpha \log(t)} \Big( \frac{1}{\sqrt{N_{j,\tau'}(t)}} + \frac{\lambda(\tau')}{ \sqrt{ N_{j,0}(t)} } \Big)	, \quad \forall (j,\tau') \in \mathcal{M}_{n}(i,\tau)
\end{split}
\end{align}	
implies 
$$	\nu_{i,\tau} +  \mathfrak c_{i,\tautemp}(t;\alpha) 
>\nu_{j,\tautemp'} + \mathfrak c_{j,\tautemp'}(t;\alpha) , \quad \forall (j,\tau') \in \mathcal{M}_{n}(i,\tau).	$$
In other words, \eqref{key_ineq} implies that $(i,\tau)$ is chosen in round $t.$
So for the derivation of the lower bound, it suffices to answer the question: How many rounds does $\texttt{RCUCB}$ need to choose $(j,\tau') \in \mathcal{M}_{n}(i,\tau)$ such that \eqref{key_ineq} holds?
For this purpose, define 
$$	\rho( \Delta_{i,\tau}, \Delta_{j,\tau'} , N_{i,\tau}(t)  ) = \min_{s} \left\{	\varepsilon  \sqrt{\frac{2 \alpha \log(s)}{N_{i,\tau}(t)}  } \geq	\Delta_{i,\tau} - \Delta_{j,\tau'}	\right\}.$$
Note that \eqref{key_ineq} can only be fulfilled if $t > \rho( \Delta_{i,\tau} , \Delta_{j,\tau'}, N_{i,\tau}(t)  ),$ as otherwise the left-hand side is non-positive, while the right-hand side is positive for any round.
For such rounds it must hold that
\begin{align*}
N_{j,\tau'}(t) > \frac{ 8 (1+\lambda(\tau'))^2 \, \alpha \log(t)}{ \big(\Delta_{i,\tau} - \Delta_{j,\tau'}- \varepsilon  \sqrt{\frac{2 \alpha \log(t)}{N_{i,\tau}(t)}  } \big)^2} , \quad \forall (j,\tau') \in \mathcal{M}_{n}(i,\tau),
\end{align*}
since $N_{j,\tau'}(t) \leq N_{j,0}(t).$ 
Next, define 
\begin{align} \label{defi_gamma}
\begin{split}
\gamma( \Delta_{i,\tau} , \Delta_{j,\tau'}, N_{i,\tau}(t)  )	  
=  \inf_{s }   \Bigg\{		&s >   \rho( \Delta_{i,\tau} , \Delta_{j,\tau'}, N_{i,\tau}(t)  ) \,  \wedge  \\
& s > 	\frac{ 8 (1+\lambda(\tau'))^2 \, \alpha \log(s)}{ \big(\Delta_{i,\tau} - \Delta_{j,\tau'} - \varepsilon \sqrt{\frac{2 \alpha \log(s)}{N_{i,\tau}(t)}  } \big)^2} \Bigg\}. 
\end{split}
\end{align}

\noindent \textbf{Claim:} On the event $A,$ it holds that $\texttt{RCUCB}$ chooses some pair $(i,\tau) \in [n]\times \mathcal{M}$ for the $(N_{i,\tau}(t)+1)$-st time not later than after
$$	\sum_{(j,\tau') \in \mathcal{M}_{n}(i,\tau)} \gamma( \Delta_{i,\tau} , \Delta_{j,\tau'}, N_{i,\tau}(t)  ) 	$$
rounds.

Assume that in round $t$ the pair $(i,\tau)$ is chosen for the $N_{i,\tau}(t)$-th time and will not be chosen thereafter.
But then there will be a round $t_1>t$ such that for some pair $(j^{(1)},\tau^{(1)}) \in  \mathcal{M}_{n}(i,\tau)$ it holds that
$$	 N_{j^{(1)},\tau^{(1)}}(t_1) =	\gamma( \Delta_{i,\tau} , \Delta_{j^{(1)},\tau^{(1)}}, N_{i,\tau}(t)  )	$$
which by construction of $\gamma$ implies that
$$	\nu_{i,\tau} +  \mathfrak c_{i,\tautemp}(t;\alpha) 
>\nu_{j^{(1)},\tau^{(1)}} + \mathfrak c_{j^{(1)},\tau^{(1)}}(t;\alpha) . 	$$
By assumption the pair $(i,\tau)$	 is not chosen anymore, so that due to the latter inequality, we can infer that for all rounds after $t_1$ the pair $(j^{(1)},\tau^{(1)})$ is never chosen again either.
Thus, there will be a round $t_2$ such that for some pair $(j^{(2)},\tau^{(2)}) \in \mathcal{M}_{n}(i,\tau) \cap \mathcal{M}_{n}(j^{(1)},\tau^{(1)}) $ it holds that
$$	 N_{j^{(2)},\tau^{(2)}}(t_2) =	\gamma( \Delta_{i,\tau} , \Delta_{j^{(2)},\tau^{(2)}}, N_{i,\tau}(t)  )	$$
which implies that
$$	\nu_{i,\tau} +  \mathfrak c_{i,\tautemp}(t;\alpha) 
>\nu_{j^{(2)},\tau^{(2)}} + \mathfrak c_{j^{(2)},\tau^{(2)}}(t;\alpha) . 	$$
Similarly as for the pair $(j^{(1)},\tau^{(1)})$ we can infer that for all rounds after $t_2$ the pair $(j^{(2)},\tau^{(2)})$ is never chosen again.
Proceeding this argumentation iteratively we conclude that
$$	N_{j^{(l)},\tau^{(l)}}(t_l) =	\gamma( \Delta_{i,\tau} , \Delta_{j^{(l)},\tau^{(l)}}, N_{i,\tau}(t)  )		$$
for $l=1,\ldots, n\times |\mathcal{M}|-1$ and the pairs $(j^{(l)},\tau^{(l)})_l$ are all distinct.
As $(i,\tau)$	 is not chosen anymore for any round after $t,$ it holds that $ N_{i,\tau}(t) = N_{i,\tau}(t_l)$ for any $l=1,\ldots, n\times |\mathcal{M}|-1.$
Hence, after 
\begin{align*}
\begin{split}
\sum_{l=1}^{ n\times |\mathcal{M}|-1} N_{j^{(l)},\tau^{(l)}}(t_l) 
&= \sum_{l=1}^{ n\times |\mathcal{M}|-1} \gamma( \Delta_{i,\tau} , \Delta_{j^{(l)},\tau^{(l)}}, N_{i,\tau}(t)  )	 \\
&= \sum_{(j,\tau') \in \mathcal{M}_{n}(i,\tau)} \gamma( \Delta_{i,\tau} , \Delta_{j,\tau'}, N_{i,\tau}(t)  ) 
\end{split}
\end{align*}
rounds, the pair $(i,\tau)$ will be chosen by the design of $\texttt{RCUCB},$ which contradicts the assumption and implies the claim.

\noindent In summary, we have shown that
$$	N_{i,\tau}(T^+) \geq \max\Big(  s \in \N \, | \,  \sum\nolimits_{(j,\tau') \in \mathcal{M}_{n}(i,\tau)} \gamma( \Delta_{i,\tau} , \Delta_{j,\tau'}, s  ) \leq T \Big) +1 .		$$
Next, we need to show that the latter maximum is at least $ l_{i,\tautemp}(T,\alpha).$
For this purpose, define 
\begin{align*}
\tilde t &= \left\lfloor  \frac{\varepsilon^2 \, \alpha \log(T)}{ \widetilde{H}_T(i,\tautemp)}  \right\rfloor, \\
c_{(i,\tau),(j,\tau')} &= \Big(	\frac{8(1+\lambda(\tau')^2 \alpha)}{\Delta_{i,\tau}-\Delta_{j,\tau'}}		+1 \Big)^2, \\
\widetilde{H}_T(i,\tautemp) &= \frac{  \max_{(j, \tau' ) \in \mathcal{M}_{n}(i,\tau)} \, c_{(i,\tau),(j,\tau')}  \, (\Delta_{i,\tau}-\Delta_{j,\tau'})^2 }{\delta \log_T(d_T) + 2\delta},  \\
d_T 	&=	\frac{1}{T} \left\lfloor \frac{T}{n|\mathcal{M}|-1} \right\rfloor \in (1/T,1),
\end{align*}
where $\delta\in(0,1/2)$ is such that $ (n|\mathcal{M}|-1)^{1-\delta} T^{2\delta} \leq T.  $
Note that $ \tilde t +1 \geq l_{i,\tautemp}(T,\alpha),$ since $\log_T(d_T)\in(-1,0)$ and consequently $\widetilde{H}_T(i,\tautemp) \leq \frac{c_{i,\tau} \, \Delta_{i,\tautemp}^2}{\delta}.$ 
Therefore, we will show in the following that 
\begin{align} \label{ineq_conclusion}
\sup_{ s } \Big\{ \sum\nolimits_{(j,\tau') \in \mathcal{M}_{n}(i,\tau)} \gamma( \Delta_{i,\tau} , \Delta_{j,\tau'}, s  ) \leq T \Big\} +1 \geq \tilde t +1.
\end{align}
By the choice of $\tilde t$ we obtain
\begin{align*}
\rho( \Delta_{i,\tau}, \Delta_{j,\tau'} , \tilde t   ) 
&= \inf_{s} \left\{	\varepsilon  \sqrt{\frac{2 \alpha \log(s)}{ \tilde t}  } \geq	\Delta_{i,\tau} - \Delta_{j,\tau'}	\right\} \\
&\leq \inf_{s} \left\{	 \sqrt{\frac{\widetilde{H}_T(i,\tautemp) \log(s)}{ \log(T) }  } \geq		\Delta_{i,\tau} - \Delta_{j,\tau'}	\right\}.
\end{align*}
Thus, 
\begin{align} \label{aux_ineq_rho}
\rho( \Delta_{i,\tau}, \Delta_{j,\tau'} , \tilde t   )  \leq \left\lceil T^{\frac{(	\Delta_{i,\tau} - \Delta_{j,\tau'})^2}{\widetilde{H}_T(i,\tautemp)}}  \right\rceil, 
\end{align}

which can be seen by rearranging the inequality within the infimum with respect to $s.$
Also it holds that $\gamma( \Delta_{i,\tau} , \Delta_{j,\tau'},  \tilde t   ) \leq \left\lceil T^{\frac{c_{(i,\tau),(j,\tau')} \, (	\Delta_{i,\tau} - \Delta_{j,\tau'})^2}{\widetilde{H}_T(i,\tautemp)}}  \right\rceil.$
Indeed, for sake of convenience define 
$$\tilde s_{(i,\tau),(j,\tau')} := \left\lceil T^{\frac{c_{(i,\tau),(j,\tau')} \, (	\Delta_{i,\tau} - \Delta_{j,\tau'})^2}{\widetilde{H}_T(i,\tautemp)}}  \right\rceil,$$
 then it holds that 

\begin{enumerate}[noitemsep,topsep=0pt,leftmargin=3.5mm]
\item 	 $\tilde s_{(i,\tau),(j,\tau')} \geq \rho( \Delta_{i,\tau}, \Delta_{j,\tau'} , \tilde t   ),$ by means of \eqref{aux_ineq_rho} and since $c_{(i,\tau),(j,\tau')}>1;$ 
\item  By choice of $c_{(i,\tau),(j,\tau')},$ we have $\big(\Delta_{i,\tau} - \Delta_{j,\tau'} - \sqrt{\frac{2 \alpha \log( \tilde s)}{ \tilde t  }  } \big)^2 > 8 (1+\lambda(\tau'))^2 \, \alpha,$ so that
\begin{align*}
\frac{ 8 (1+\lambda(\tau'))^2 \, \alpha \log( \tilde s_{(i,\tau),(j,\tau')})}{ \big(\Delta_{i,\tau} - \Delta_{j,\tau'} - \sqrt{\frac{2 \alpha \log( \tilde s_{(i,\tau),(j,\tau')})}{ \tilde t  }  } \big)^2} 
< \log( \tilde s_{(i,\tau),(j,\tau')}) \leq \tilde s_{(i,\tau),(j,\tau')}.
\end{align*}
\end{enumerate}
Regarding the definition of $	\gamma	$ in \eqref{defi_gamma}, we obtain $\gamma( \Delta_{i,\tau} , \Delta_{j,\tau'},  \tilde t   ) \leq \tilde s_{(i,\tau),(j,\tau')}.$
It remains to show \eqref{ineq_conclusion}, for which it suffices to verify that
$$	\sum\nolimits_{(j,\tau') \in \mathcal{M}_{n}(i,\tau)} \gamma( \Delta_{i,\tau} , \Delta_{j,\tau'}, \tilde t ) \leq T.	$$
This can be seen as follows,
\begin{align*}
\sum\nolimits_{(j,\tau') \in \mathcal{M}_{n}(i,\tau)} \gamma( \Delta_{i,\tau} , \Delta_{j,\tau'}, \tilde t ) 
&\leq 	\sum\nolimits_{(j,\tau') \in \mathcal{M}_{n}(i,\tau)} \tilde s_{(i,\tau),(j,\tau')} \\
&= 	\sum\nolimits_{(j,\tau') \in \mathcal{M}_{n}(i,\tau)} \left\lceil T^{\frac{c_{(i,\tau),(j,\tau')} \, (	\Delta_{i,\tau} - \Delta_{j,\tau'})^2}{\widetilde{H}_T(i,\tautemp)}}  \right\rceil \\
&\stackrel{(a)}{\leq} 	\sum\nolimits_{(j,\tau') \in \mathcal{M}_{n}(i,\tau)} \left\lceil T^{  \delta \log_T(d_T) + 2\delta }  \right\rceil \\
&= 	 (n|\mathcal{M}|-1) \left\lceil  d_T^{\delta} \, T^{2\delta}  \right\rceil \\
&= 	 (n|\mathcal{M}|-1) \left\lceil  \left(\left\lfloor \frac{T}{n|\mathcal{M}|-1} \right\rfloor \right)^{\delta} \, T^{\delta}  \right\rceil \\
&\stackrel{(b)}{\leq}    T,
\end{align*}
where we used for $(a)$ that $(\delta \log_T(d_T) + 2\delta) \,\widetilde{H}_T(i,\tautemp) \geq c_{(i,\tau),(j,\tau')} \, (	\Delta_{i,\tau} - \Delta_{j,\tau'})^2,$ while $(b)$ follows by the choice of $\delta.$ 
\end{proof}

\section{Proof of Theorem \ref{theorem_regret_upper_bound_zRCUCB}: Loss Bound of \texttt{z-RCUCB}} \label{sec_appendix_proof_regret_upper_bound_z_RCUCB}

We start by recalling the definition of $\eta$-near-optimality of  the $\epsilon$-best arm/resource-limit pairs.

\setcounter{definition}{0}
\begin{definition} 
The $\eta$-near-optimality dimension is the smallest $d\geq0$ such that there exists a constant $C>0$ (the $\eta$-near-optimality constant) such that for all $\varepsilon>0,$ the maximal number of disjoint sets of the form
$$ \{j\}\times(a_j,b_j], \qquad j\in[n], 0\leq a_j<b_j \leq \tau_{\mathrm{max}} $$
such that  $|b_j -a_j| \leq \eta \varepsilon$ and $(j,b_j)$ is element of $\{	(i,\tau) \in [n]\times \mathcal{M} \, | \,     \nu_{i,\tau} \geq \nu_{i^*,\tau^*} -	\epsilon \},$ is less than $C \varepsilon^{-d}.$
\end{definition}
Further, recall that $m\geq 2$ is the grid refinement size of $\texttt{z-RCUCB}.$
Define 
$$\mathcal{M}^* = \bigcup_{l \in \N_0} \bigcup_{1 \leq j  \leq m^l } \{\tau_{(l,j)}\}, $$
where 
$\tau_{(l,0)}=0$ and
$\tau_{(l,m^l)}=\tau_{\mathrm{max}}$ for any $l$ and for any $l\geq 1,$ $j\in \{1,\ldots,m^l\}$ it holds
\begin{align} \label{defi_grid_points}
\tau_{(l,j)} = \frac{o_j}{m} (\tau_{(l-1,s_j)} - \tau_{(l-1,s_j-1)} ),
\end{align}
where $ s_j = \lfloor j/m \rfloor +1$ and 
$$o_j=\begin{cases}
j \mod m, & \mbox{if $j \mod m \neq 0$} \\
m, & \mbox{else}.
\end{cases}$$
Note that for any $l \geq 0$ and any $j,k \in\{0,\ldots,m^l\}$ such that $|j-k|\leq 1$ it holds   
\begin{align} \label{grid_granularity}
|\tau_{(l,j)} - \tau_{(l,k)}| \leq m^{-l}.
\end{align}
In particular, for $l \geq 0$ fixed the set $\bigcup_{1 \leq j  \leq m^l } \{\tau_{(l,j)}\}$ corresponds to the grid points of an equidistant decomposition of $\mathcal{M}$ with granularity $m^{-l}.$
%
%
Moreover, the time-dependent grid point sets $(\mathcal{M}_t^{(i)})_{i\in[n]}$ maintained by $\texttt{z-RCUCB}$ are of the form as in \eqref{defi_grid_points}, that is, each $\mathcal{M}_t^{(i)}$ is a subset of $\mathcal{M}^*.$ \\
\medskip

\noindent For sake of convenience, we restate Theorem \ref{theorem_regret_upper_bound_zRCUCB} here again and recall the assumption we make  on the local smoothness of the optimal penalized expected gain (interpreted as a function of $\tau$):
\begin{align} \label{assump_local_smooth}
\nu_{i^*,\tau^*}  - \nu_{i^*,\tau} \leq |\tau^* - \tau |, \qquad \forall \tau \in \mathcal{M}. 
\end{align}
%
\begin{theorem} 
Let $d$ be the $1/3$-near-optimality
of 
$$ \{	(i,\tau) \in [n]\times \mathcal{M} \, | \,     \nu_{i,\tau} \geq \nu_{i^*,\tau^*} -	\epsilon \},$$ 
i.e., the set of all $\epsilon$-best arm/resource-limit pairs,  with corresponding  near optimality constant $C>0.$ 
Then, for any $\delta\in(0,1), \alpha>1$ it holds with probability at least $1-\delta$ that
$$		L_T^{\texttt{z-RCUCB}}   \leq 	\tilde C \big(	\nicefrac{ \log(T^2/\delta)}{T}	\big)^{\frac{1}{d+2}},	$$
where 
%
\begin{align*}
    \tilde C &= \left( \frac{4\alpha  \, C \, m^2 \, (H_T + (1+\lambda(\tau_{\mathrm{max}}))^2)}{3^d (1- m^{-(d+2)})} \right)^{1/(d+2)}, \\
	H_T &= \max_{l\geq 0, 1\leq j \leq m^l-1}  \left( (1+\lambda(\tau_{(l,j)}))^2  - \frac{1}{m^2} \left(1+\lambda(\tau_{(l-1,s_j)})\, c_T   \right)^2 \right), \\
	c_T &= \sqrt{ \frac{2\alpha \log(nT^2/\delta) \left(1+\lambda(\tau_{\mathrm{max}}) \right)^2  }{\tau_{\mathrm{max}}^2 T}}.
\end{align*}
%
%
Moreover, setting $\delta=1/T$ yields 
$$ \exptd [	L_T^{\texttt{z-RCUCB}}] = O\big(\big(	\nicefrac{\log(T)}{T}	\big)^{\frac{1}{d+2}}\big).  $$
\end{theorem}
\newcommand{\vol}{\mathrm{vol}}
For the proof (at the end of this section) we need some preparations.
Thus, in the following we formulate and prove some auxiliary results needed for the proof of Theorem \ref{theorem_regret_upper_bound_zRCUCB}.

\begin{lemma} \label{lemmama_aux_high_prob_event}
Let $\delta\in(0,1).$ 
Define 
$$A_\delta  = \Big\{ \forall	t\in[T], \forall i \in [n], \forall l \geq 0, \forall 1\leq j \leq m^l: |\hat{\nu}_{i,\tau_{(l,j)}}(t)  - \nu_{i,\tau_{(l,j)}}| \leq \mathfrak c_{i,\tautemp}(n \, T^2 /\delta;\alpha)		\Big\} .$$
Then, $\mathbb{P}(A_\delta)\geq 1-\delta.$
\end{lemma}
\begin{proof}
This follows immediately by Proposition \ref{prop_nu_estimates} and the union bound.
\end{proof}

\begin{lemma} \label{lemma_extension_set}
Let $\delta\in(0,1).$ 
%
%
On the event $A_\delta$ as defined in Lemma \ref{lemmama_aux_high_prob_event} it holds that if $\mathcal{M}^{(I_t)}_t$ is extended at $\tilde \tau_t \leq \tau_t$ by $\texttt{z-RCUCB}$ in round $t,$ then 
$$	(I_t,\tilde \tau_t) \in \bigcup_{i \in [n], l \in \N_0, 1\leq j \leq m^l}  
\Big\{	(i,\tau_{(l,j)}) \, \big| \, \exists 0 \leq l'\leq l: \, \sup_{ x \in (\tau_{(l,j-1)},\tau_{\mathrm{max}}] }  \nu_{i,x}	 + 3 m^{-l'} \geq \nu_{i^*,\tau^*}	\Big\} .$$
and
$$	(I_t, \tau_t) \in \bigcup_{i \in [n], l \in \N_0, 1\leq j \leq m^l}   \Big\{	(i,\tau_{(l,j)}) \, \big| \,   \nu_{i,\tau_{(l,j)}}	 + 3 m^{-(l-1)} \geq \nu_{i^*,\tau^*}	\Big\} .$$
In particular, 
$$ (I_t, \tau_t) \in \bigcup_{i \in [n], l \in \N_0, 1\leq j \leq m^l}   \Big\{	(i,\tau_{(l,j)}) \, \big| \, \exists 0 \leq l'\leq l: \, \sup_{ x \in (\tau_{(l,j-1)},\tau_{\mathrm{max}}] }  \nu_{i,x}	 + 3 m^{-l'} \geq \nu_{i^*,\tau^*}	\Big\} ,$$ 
i.e., $\tau_t$ belongs to the set of grid points which will be extended in $\mathcal{M}^{(I_t)}_t$.
\end{lemma}
\begin{proof}
Note that a grid point in $\mathcal{M}_t^{(i)}$ for some $i\in[n]$ can only be extended in round $t$ iff $I_t=i,$ i.e., the corresponding arm was chosen in that round.
Moreover, as all counter variables $ N_{I_t,\tilde \tau}$ such that $\tilde \tau \in \mathcal{M}_t^{(I_t)}$ and $\tilde \tau \leq \tau_t$ hold are incremented in round $t,$ it could be that more than one grid point is extended in one round $t.$
Let for a round $t \in [T]$
\begin{itemize}
%
\item	$ \tilde \tau_t = \tau_{\tilde l_t, \tilde j_t}^{(I_t)}$ be one arbitrary but fixed grid point such that  $\mathcal{M}^{(I_t)}_t$ is extended at $\tilde \tau_t \leq \tau_t;$  
\item $ \tau_t = \tau_{l_t,j_t}^{(I_t)} \in \mathcal{M}_t^{(I_t)}$ be the resource limit chosen in round $t \in [T];$ 
\item $\tau_t^* = \tau_{l_t^*,j_t^*}^{(i^*)}$ be the grid point in $\mathcal{M}_t^{i^*}$ such that $\tau_{l_t^*,j_t^*}^{(i^*)} \geq \tau^* > \low_{i^*}(\tau_{l_t^*,j_t^*}^{(i^*)}).$
\end{itemize}
Write for sake of convenience, $b_i(\tau)=(\tau-\low_{i}(\tau))$ for any $i\in[n]$ and $\tau \in \mathcal{M}_t^{(i)}$ and let $\tilde b$  be $   b_{I_t}(\tau_t) = m^{-l_t}$ if $\mathcal{M}^{(I_t)}_t$ is extended at  $ \tau_t $ and $  m \,  b_{I_t}(\tau_t) = m^{-l_t+1}$ otherwise.
Then, we can infer the following two inequalities by means of the design of  $\texttt{z-RCUCB}$:
\begin{itemize}
%
\item By definition of $\tilde b,$ it must hold that
%
\begin{align} \label{ineq_ext}
\tilde b \geq  b_{I_t}(  \tau_t)   \quad \mbox{and} \quad \mathfrak c_{I_t,\tau_t}(n\, T^2/\delta;\alpha)  \leq \tilde b	.
\end{align}
\item As $(I_t,\tau_t)$ is selected in round $t$ it must hold that
\begin{align}  \label{ineq_choice}
\hat \nu_{I_t,\tau_t}(t) + \mathfrak c_{I_t,\tau_t}(n\, T^2/\delta;\alpha)   + b_{I_t}(\tau_t) > \hat \nu_{i^*,\tau_t^*}(t) + \mathfrak c_{i^*,\tau_t^*}(n\, T^2/\delta;\alpha)   + b_{i^*}(\tau_t^*).
\end{align}
\end{itemize}
We obtain the assertion by means of
\begin{align*}
\sup_{x\in(\low_{I_t}(\tilde \tau_t),\tau_{\mathrm{max}}]} \nu_{I_t,x} 
&\geq \nu_{I_t,\tau_t}   \\
&\stackrel{(a)}{\geq} \hat \nu_{I_t,\tau_t}(t) - \mathfrak c_{I_t,\tau_t}(n\, T^2/\delta;\alpha) \\
&= \hat \nu_{I_t,\tau_t}(t) + \mathfrak c_{I_t,\tau_t}(n\, T^2/\delta;\alpha) + b_{I_t}(\tau_t) \\
&\quad - 2 \mathfrak c_{I_t,\tau_t}(n\, T^2/\delta;\alpha) -  b_{I_t}(\tau_t) \\
&\stackrel{(b)}{\geq} \hat \nu_{I_t,\tau_t}(t) + \mathfrak c_{I_t,\tau_t}(n\, T^2/\delta;\alpha) + b_{I_t}(\tau_t) - 3 \tilde b \\
&\stackrel{(c)}{\geq} \hat \nu_{i^*,\tau_t^*}(t) + \mathfrak c_{i^*,\tau_t^*}(n\, T^2/\delta;\alpha)   + b_{i^*}(\tau_t^*) - 3 \tilde b \\
&\stackrel{(a)}{\geq}  \nu_{i^*,\tau_t^*}  + b_{i^*}(\tau_t^*) - 3 \tilde b \\
&\stackrel{(d)}{\geq}  \nu_{i^*,\tau_*}  - 3 \tilde b,
\end{align*}
where $(a)$ is due to the event $A_\delta,$ $(b)$ follows by \eqref{ineq_ext}, $(c)$ by \eqref{ineq_choice} and $(d)$ by \eqref{assump_local_smooth}.
\end{proof}

\begin{lemma} \label{lemma_card_played_pairs}
Let $d$ be the $1/3$-near-optimality dimension and $C>0$ the corresponding near-optimality constant of $\{	(i,\tau) \in [n]\times \mathcal{M} \, | \,     \nu_{i,\tau} \geq \nu_{i^*,\tau^*} -	\epsilon \}.$ 
Further, let for any $\epsilon>0, i\in[n]$
%
$$ \mathcal{E}^{(i)}(\epsilon) = \Big\{	\{i\} \times (\tilde \tau,\tau] \, \Big| \, \tilde \tau,\tau\in\mathcal{M}, \, \nu_{i,\tau}	 + \epsilon \geq \nu_{i^*,\tau^*}	\Big\},   $$
%
and 
$$ I_l^{(i)} = \Big\{ j\in\{1,\ldots,m^l\} \, \Big| \, \{i\} \times (\tau_{(l,j-1)}, \tau_{(l,j)}] \in 	\mathcal{E}^{(i)}(3/m^{l-1})	\Big\}.  $$ 
Then,
%
%
$$ \sum_{i\in[n]} \sum_{j=1}^{m^l} \IND{\{ j \in I_l^{(i)} \} } = \sum_{i\in[n]} |I_l^{(i)}| \leq C (m^{(l-1)}/3)^{d}. $$
\end{lemma}
\begin{proof}
This follows immediately by the definition of $d,$ as the sets in the condition of $I_l^{(i)}$ are exactly of the form as in Definition \ref{defi_near_opt} for $\epsilon = 3/m^{l-1},$ since $ \tau_{(l,j)} - \tau_{(l,j-1)} \leq 1/m^{l} \leq 1/m^{l-1} = \epsilon$ (cf.\ \eqref{grid_granularity}).
\end{proof}

\begin{proposition} \label{prop_key_zrcucb}
Let $\delta\in(0,1)$ and define $l(T)$ as the smallest integer $l'$ such that
$$			T 
\leq \frac{2\alpha \log(nT^2/\delta) \, C \, (H_T + (1+\lambda(\tau_{\mathrm{max}}))^2) }{(3m)^d} \sum_{l=0}^{l'} (m^{d+2})^l	 .	$$
where 
\begin{align*}
H_T &= \max_{l\geq 0, 1\leq j \leq m^l-1}  \left( (1+\lambda(\tau_{(l,j)}))^2  - \frac{1}{m^2} \left(1+\lambda(\tau_{(l-1,s_j)})\, c_T   \right)^2 \right),  \\
c_T &= \sqrt{ \frac{2\alpha \log(nT^2/\delta) \left(1+\lambda(\tau_{\mathrm{max}}) \right)^2  }{\tau_{\mathrm{max}}^2 T}},
\end{align*}
and $s_j=\lfloor (j+1)/m \rfloor + 1.$ 
Then, with probability at least $1-\delta,$ it holds that
$$		L_T^{\texttt{z-RCUCB}} \leq 3 m^{-(l(T)-1)}. $$
\end{proposition}
\begin{proof}
Denote by $T_{i,\tau_{(l,j)}  }(t) = \sum_{s=1}^{t-1} \IND{ \{ I_s^{\texttt{z-RCUCB}} = i \, \wedge \,   \tau_s^{\texttt{z-RCUCB}} = \tau_{(l,j)} \}}$  the number of times the pair $(i,\tau_{(l,j)})$ has been chosen by $\texttt{z-RCUCB}$ till round $t\in[T].$
Let $l_{max}$ be the finest grid level among the grid sets $(\mathcal{M}_t^{(i)})_{i \in [n]}.$
On the event $A_\delta$ only pairs $(i,\tau_{(l,j)})$ such that $j \in I_l^{(i)}$ are chosen according to the second part of Lemma \ref{lemma_extension_set}.
Such pairs are elements of $\{i\} \times \mathcal{M}_T^{(i)}$ due to the first part of Lemma \ref{lemma_extension_set}.
Thus,
\begin{align} \label{time_horizon_decomp}
\begin{split}
T 	&= \sum_{i \in [n] } \sum_{\tau_{(l,j)} \in \mathcal{M}_T^{(i)}} T_{i,\tau_{(l,j)}}(T^+) \\
&\leq \sum_{i \in [n] } \sum_{l=0}^{l_{max}} \sum_{j=1}^{m^l} T_{i,\tau_{(l,j)}}(T^+) \IND{\{ \tau_{(l,j)} \in \mathcal{M}_T^{(i)} \wedge j \in I_l^{(i)} \}} \\
&= \sum_{i \in [n] } \sum_{l=1}^{l_{max}} \sum_{j=1}^{m^l-1} N_{i,\tau_{(l,j)}}(T^+)  - N_{i,\tau_{(l,j+1)}}(T^+)   \IND{\{ \tau_{(l,j)} \in \mathcal{M}_T^{(i)} \wedge j \in I_l^{(i)} \}} \\
&\quad +  \sum_{i \in [n] } \sum_{l=0}^{l_{max}}  N_{i,\tau_{\mathrm{max}}}(T^+)    \IND{\{  \tau_{(l,m^l)} \in \mathcal{M}_T^{(i)} \wedge m^l \in I_l^{(i)} \}} ,
%
%
\end{split}
\end{align}	
where we used that $T_{i,\tau_{(l,j)}}(T^+) = N_{i,\tau_{(l,j)}}(T^+)  - N_{i,\tau_{(l,j+1)}}(T^+) $ for any $l\geq0, 1\leq j \leq m^l-1$ and $T_{i,\tau_{\mathrm{max}}}(T^+) =  N_{i,\tau_{\mathrm{max}}}(T^+)$ for any $i\in[n]$.

If $\tau_{(l,j)} \in \mathcal{M}_T^{(i)},$ which is by design of $\texttt{z-RCUCB}$ equivalent to that $\mathcal{M}_T^{(i)}$ has not been extended\footnote{Strictly speaking, we have that $\tau_{l,km} = \tau_{l-1,k}$ for any $k=1,\ldots,m^{l-1}$ so that we mean by ``$\mathcal{M}_T^{(i)}$ has not been extended at $\tau_{(l,j)}$'' rather that $\mathcal{M}_T^{(i)}$ has not been expanded at $\tau_{(l,j)}$ \emph{with refinement level} $l,$ which is equivalent to $  (\tau_{(l,j)} - \tau_{(l,j-1)}) < \mathfrak c_{i,\tau_{(l,j)}}(n\, T^2/\delta;\alpha).$ } at $\tau_{(l,j)}$.  
This in turn is equivalent to $  (\tau_{(l,j)} - \tau_{(l,j-1)}) < \mathfrak c_{i,\tau_{(l,j)}}(n\, T^2/\delta;\alpha),$   which implies
\begin{align} \label{ineq_aux_upper_bound}
\begin{split}
N_{i,\tau_{(l,j)}}(T^+)   
&\leq \frac{2\alpha \log(nT^2/\delta) \left(1+\lambda(\tau_{(l,j)})  \sqrt{\frac{N_{i,\tau_{(l,j)}}(T^+)}{N_{i,0}(T^+)}} \right)^2  }{(\tau_{(l,j)} - \tau_{(l,j-1)})^2} .
%
%
\end{split}
\end{align}
Since $N_{i,0}\geq N_{i,\tau}$ for any $(i,\tau) \in [n] \times \mathcal{M},$ we can infer that
\begin{align} \label{upper_bound_N_terms}
\begin{split}
N_{i,\tau_{(l,j)}}(T^+)   
\leq \frac{2\alpha \log(nT^2/\delta) (1+\lambda(\tau_{(l,j)}))^2  }{(\tau_{(l,j)} - \tau_{(l,j-1)})^2} 
&= 2\alpha \log(nT^2/\delta) (1+\lambda(\tau_{(l,j)}))^2  m^{2l} ,
\end{split}
\end{align}
It holds that $ N_{i,\tau_{(l,j)}}(t) \geq N_{i,\tau_{\mathrm{max}}}(t) $ for any $l\geq0, 1\leq j \leq m^l$ and if $|\mathcal{M}_T^{(i)}|>1,$ then this implies that $\mathcal{M}_T^{(i)}$ was extended at $\tau_{\mathrm{max}}.$
This in turn is equivalent to $ (\tau_{\mathrm{max}}) \geq  \mathfrak c_{i,\tau_{\mathrm{max}}}(n\, T^2/\delta;\alpha) $ which similarly as for \eqref{ineq_aux_upper_bound} with reversed inequality yields
\begin{align*}
N_{i,\tau_{\mathrm{max}}}(\rho^{(i)})  
%
%
&\geq \frac{2\alpha \log(nT^2/\delta) \left(1+\lambda(\tau_{\mathrm{max}})  \sqrt{\frac{N_{i,\tau_{\mathrm{max}}}(\rho^{(i)})}{N_{i,0}(\rho^{(i)})}} \right)^2  }{\tau_{\mathrm{max}}^2},
%
%
\end{align*}
where we denote by $\rho^{(i)} \in [T]$ the round, in which the extension took place.
However, for any $t \in [\rho^{(i)}]$ we have $N_{i,\tau_{\mathrm{max}}}(t) =  N_{i,0}(t),$ so that we can infer
\begin{align} \label{lower_bound_tau_max}
N_{i,\tau_{(l,j)}}(T^+) \geq N_{i,\tau_{\mathrm{max}}}(\rho^{(i)})  
%
%
&\geq \frac{2\alpha \log(nT^2/\delta) \left(1+\lambda(\tau_{\mathrm{max}}) \right)^2  }{\tau_{\mathrm{max}}^2} = c_T^2 \, T.
%
%
\end{align}
Moreover, for any $l\geq 1$ it holds that $N_{i,\tau_{(l,j+1)}}(T^+) \geq N_{i,\tau_{(l-1,s)}}(T^+),$ where $s_j=\lfloor (j+1)/m \rfloor + 1.$ 
Further, $  (\tau_{(l-1,s_j)} - \tau_{(l-1,s_j-1)}) \geq \mathfrak c_{i,\tau_{(l-1,s_j)}}(n\, T^2/\delta;\alpha)$ holds, since $\tau_{(l-1,s_j)}$ is the grid point which was extended in order to generate $\tau_{(l,j+1)}.$
Thus, if $l\geq 1$ similarly as for \eqref{ineq_aux_upper_bound} with reversed inequality, we obtain
\begin{align*}
N_{i,\tau_{(l,j+1)}}(T^+) 
&\geq N_{i,\tau_{(l-1,s_j)}}(T^+) \\
%
%
&\geq \frac{2\alpha \log(nT^2/\delta) \left(1+\lambda(\tau_{(l-1,s_j)})  \sqrt{\frac{N_{i,\tau_{(l-1,s_j)}}(T^+)}{N_{i,0}(T^+)}} \right)^2  }{(\tau_{(l-1,s_j)} - \tau_{(l-1,s_j-1)})^2}.
%
%
\end{align*}
Using the (crude) estimate $N_{i,0}\leq T$ and \eqref{lower_bound_tau_max} yields 
\begin{align} \label{lower_bound_N_terms}
N_{i,\tau_{(l,j+1)}}(T^+) 
\geq	2\alpha \log(nT^2/\delta) \left(1+\lambda(\tau_{(l-1,s_j)})\, c_T   \right)^2 m^{2(l-1)},
\end{align}
%
%
Recall that
$$	H_T = \max_{l\geq 0, 1\leq j \leq m^l-1}  \left( (1+\lambda(\tau_{(l,j)}))^2  - \frac{1}{m^2} \left(1+\lambda(\tau_{(l-1,s_j)})\, c_T   \right)^2 \right), $$
then by using \eqref{upper_bound_N_terms}, \eqref{lower_bound_N_terms} together with Lemma \ref{lemma_card_played_pairs} to bound the right-hand side in \eqref{time_horizon_decomp}, we obtain
\begin{align}
\begin{split}
T 
&\leq 		\frac{2\alpha \log(nT^2/\delta) \, C}{(3m)^d} \left(	H_T	\Big(-1 + \sum_{l=0}^{l_{max}} (m^{d+2})^l\Big) + (1+\lambda(\tau_{\mathrm{max}}))^2	\sum_{l=0}^{l_{max}} (m^{d+2})^l	\right) \\
&\leq \frac{2\alpha \log(nT^2/\delta) \, C \, (H_T + (1+\lambda(\tau_{\mathrm{max}}))^2) }{(3m)^d} \sum_{l=0}^{l_{max}} (m^{d+2})^l	 .
\end{split}
\end{align}
By definition of $l'$ we have that $l_{max}\geq l',$ so that the grid point $\hat \tau$ which has the finest grid level among the grid sets $(\mathcal{M}_t^{(i)})_{i \in [n]},$ fulfills $ \nu_{\hat i,\hat \tau}	 + 3 m^{-(l(T)-1)} \geq \nu_{i^*,\tau^*} $ on the event $A_\delta$ according to Lemma \ref{lemma_extension_set}.
The latter event has probability at least $1-\delta$ by Lemma \ref{lemmama_aux_high_prob_event}.
\end{proof}

\begin{proof} [\sl Proof of Theorem \ref{theorem_regret_upper_bound_zRCUCB} ]
We intend to use Proposition \ref{prop_key_zrcucb} by providing an upper bound for $m^{-l(T)+1}.$ 
By definition of $l(T)$ it holds
\begin{align*}
T 
&\leq \frac{2\alpha \log(nT^2/\delta) \, C \, (H_T + (1+\lambda(\tau_{\mathrm{max}}))^2) }{(3m)^d} \sum_{l=0}^{l(T)} (m^{d+2})^l \\
&= 	\frac{2\alpha \log(nT^2/\delta) \, C \, (H_T + (1+\lambda(\tau_{\mathrm{max}}))^2) }{(3m)^d} \frac{m^{(d+2)(l(T)+1)}-1}{m^{(d+2)}-1} \\
&\leq 	\frac{2\alpha \log(nT^2/\delta) \, C \, (H_T + (1+\lambda(\tau_{\mathrm{max}}))^2) }{(3m)^d} \frac{m^{(d+2)l(T)}}{1- m^{-(d+2)}}.
\end{align*}
By rearranging we obtain
\begin{align*}
m^{-l(T)+1}  
&\leq 	\left( \frac{2\alpha  \, C \, m^2 \, (H_T + (1+\lambda(\tau_{\mathrm{max}}))^2)}{3^d (1- m^{-(d+2)})} \right)^{1/(d+2)} \left(\frac{\log(nT^2/\delta)}{T}\right)^{1/(d+2)} \\
&\leq 	 \left( \frac{4\alpha  \, C \, m^2 \, (H_T + (1+\lambda(\tau_{\mathrm{max}}))^2)}{3^d (1- m^{-(d+2)})} \right)^{1/(d+2)} \left( \frac{\log(T^2/\delta)}{T}\right)^{1/(d+2)},
\end{align*}
where we used that $\log(n)/T \leq \frac{\log(T^2/\delta)}{T}.$
Setting $$\tilde C= \left( \frac{4\alpha  \, C \, m^2 \, (H_T + (1+\lambda(\tau_{\mathrm{max}}))^2)}{3^d (1- m^{-(d+2)})} \right)^{1/(d+2)}, $$ yields 	$L_T^{\texttt{z-RCUCB}}   \leq 	\tilde C \big(	\nicefrac{ \log(T^2/\delta)}{T}	\big)^{\frac{1}{d+2}},$ with probability at least $1-\delta$ due to Proposition \ref{prop_key_zrcucb}.
Finally, by using $\delta=1/T$ 
\begin{align*}
\exptd [L_T^{\texttt{z-RCUCB}} ] \leq (1-\delta) \tilde C \big(	\nicefrac{ \log(T^2/\delta)}{T}	\big)^{\frac{1}{d+2}}  + \delta (1+\lambda(\tau_{\mathrm{max}}))  = O\big(\big(	\nicefrac{\log(T)}{T}	\big)^{\frac{1}{d+2}}\big),
\end{align*}
since the loss is always bounded by $(1+\lambda(\tau_{\mathrm{max}})).$
\end{proof}

\section{Update complexities: Proofs of Proposition \ref{prop_update_complexity_RCUCB} and \ref{prop_update_complexity_zRCUCB}} \label{sec_proof_updates}

\setcounter{proposition}{1}
\begin{proposition}
	$\texttt{RCUCB}$ has a worst case update complexity of order $O(n|\mathcal{M}|).$
\end{proposition}
\begin{proof}
%
	The update step of $\texttt{RCUCB}$ consists in updating the estimates $\hat{\nu}_{i,\tautemp}$ and the corresponding confidence lengths $\mathfrak c_{i,\tautemp}.$
	For the former we need to update in the worst case after choosing the pair $(I_t,\tau_t)$ all $\hat{\nu}_{I_t,\tautemp}$ with $\tautemp \leq \tau_t$ (cf.\ Section \ref{sec_estimates}).
	For the latter we have to update all confidence lengths, i.e., for any $(i,\tautemp) \in [n]\times \mathcal{M}$, as each $\mathfrak c_{i,\tautemp}$ depends on the current learning round $t$ due to the logarithmic term occurring in each of them.
	There are $n\cdot |\mathcal{M}|$ many confidence lengths, so that the worst case update complexity is of order $O(n|\mathcal{M}|).$
\end{proof}

\begin{proposition}  
	$\texttt{z-RCUCB}$'s  update complexity is in $O(|\mathcal{M}^{(I_t)}_t|+(m-1)).$
\end{proposition}
\begin{proof}
	%
	Similarly as in the update step of $\texttt{RCUCB},$ one needs to update $O(|\mathcal{M}^{(I_t)}_t|)$ many penalized expected gain estimates $\hat{\nu}_{i,\tautemp}$ in the worst case.
	However, for $\texttt{z-RCUCB}$ the corresponding confidence lengths $\mathfrak c_{i,\tautemp}$ depend only on the current learning round $t$ due to counter variables $N_{i,\tautemp}$ and  $N_{i,0},$ which need to be updated only for the chosen arm $I_t.$ 
	This leads to $O(|\mathcal{M}^{(I_t)}_t|)$ many updates.
	Finally, as $\mathcal{M}^{(I_t)}_t$ might be extended at some point $\tau$ in the learning round $t,$ one needs to initialize $O(m-1)$ many penalized expected gain estimates as well as $O(m-1)$ many confidence lengths.
	Having all things considered, we see that $\texttt{z-RCUCB}$ has indeed a worst case update complexity of order  $O(|\mathcal{M}^{(I_t)}_t|+(m-1)).$
\end{proof}

\section{Simulation Details} \label{sec_exp_appendix}

In this section we explain all details of the simulations carried out in Section \ref{sec_exp}.
In Subsection \ref{subsec_baselines} we explain the variants of UCB and Thompson Sampling, while in Subsection \ref{subsec_exp_synth_details} the distributions used in the synthetic data experiment are described.

\subsection{Compared Algorithms} \label{subsec_baselines}

The modifications of UCB and Thompson Sampling used in the experimental study are as follows.

\textbf{Upper Confidence Bound.}
The modified $\texttt{UCB}$ algorithm chooses, after sampling each possible pair once, its pair $(I_t^{\texttt{UCB}},\tau_t^{\texttt{UCB}}) \in [n]\times \mathcal{M}$ in round $t \in \{ n| \mathcal{M}|+1,\ldots,T \}$ according to
$$  (I_t^{\texttt{UCB}},\tau_t^{\texttt{UCB}}) \in \argmax_{ (i,   \tautemp) \in [n] \times \mathcal{M}}  \big(\tilde \nu_{i,\tautemp}(t) + \tilde{\mathfrak{c}}_{i,\tautemp}(t;\alpha)\big) \, , $$
where 
\begin{align}\label{def_standard_estimate}
\begin{split}
\tilde \nu_{i,\tautemp}(t) &= \frac{\tilde g_{i,\tautemp}(t) - \tilde \Lambda_{i,\tautemp}(t) + \lambda(\tau_{\mathrm{max}})}{(1+\lambda(\tau_{\mathrm{max}}))} \\
\tilde g_{i,\tautemp}(t) &=   \frac{\sum_{s=1}^{t-1}  ( R_{i,s} - c(C_{i,s})) \cdot \IND{\{ C_{i,s}\leq \tau_s^{\texttt{UCB}}  \} } \cdot \IND{ \{ I_s^{\texttt{UCB}} = i \, \wedge \,   \tau_s^{\texttt{UCB}} = \tautemp \}}}{T_{i,\tautemp}^{\texttt{UCB}}(t)}, \\
\tilde \Lambda_{i,\tautemp}(t) &=  \lambda(\tautemp) \frac{\sum_{s=1}^{t-1}  \IND{\{ C_{i,s} > \tau_s^{\texttt{UCB}}  \} } \cdot \IND{ \{ I_s^{\texttt{UCB}} = i \, \wedge \,   \tau_s^{\texttt{UCB}} = \tautemp \}}}{T_{i,\tautemp}^{\texttt{UCB}}(t)}, \\
T_{i,\tautemp}^{\texttt{UCB}}(t) &= \sum_{s=1}^{t-1} \IND{ \{ I_s^{\texttt{UCB}} = i \, \wedge \,   \tau_s^{\texttt{UCB}} = \tautemp \}},
\end{split}
\end{align}
and 
$$ \tilde{\mathfrak{c}}_{i,\tautemp}(t;\alpha) = \sqrt{  \frac{\alpha   \, \log(t)}{2 \, T_{i,\tautemp}^{\texttt{UCB}} }}. $$  
Note that adding $\lambda(\tau_{\mathrm{max}})$ and then dividing by $1+\lambda(\tau_{\mathrm{max}})$ in $\tilde \nu_{i,\tautemp}(t) $  is equivalent to assuming that the reward sequence in \eqref{def_rewards_sequence} is in $[0,1],$ so that the standard confidence lengths $\tilde{\mathfrak{c}}_{i,\tautemp}(t;\alpha)$ can be used.
It is worth mentioning that this rescaling does not alter the ranking among the pairs $(i,\tau) \in [n] \times \mathcal{M}$ with respect to their (sub-)optimality.

\textbf{Thompson Sampling.} Let 	$	T_{i,\tautemp}^{\texttt{TS}}(t) = \sum_{s=1}^{t-1} \IND{ \{ I_s^{\texttt{TS}} = i \, \wedge \,   \tau_s^{\texttt{TS}} = \tautemp \}},$ then the modified $\texttt{TS}$ algorithm chooses, after sampling each possible pair once, its pair $(I_t^{\texttt{TS}},\tau_t^{\texttt{TS}})\in [n]\times \mathcal{M}$ in round $t \in \{ n| \mathcal{M}|+1,\ldots,T \}$  according to 
$$  (I_t^{\texttt{TS}},\tau_t^{\texttt{TS}}) \in \argmax_{ (i,   \tautemp) \in [n] \times \mathcal{M}} \,  \hat \theta_{i,\tautemp}(t), $$
where $\hat \theta_{i,\tautemp}(t) $ is a sample from a $\mathrm{Beta}(1 + S_{i,\tautemp}(t),1 + F_{i,\tautemp}(t))$ distribution.
Here, $ S_{i,\tautemp}(t)$ and $ F_{i,\tautemp}(t)$ are the parameters of the posterior distribution, which are updated if $i=I_t^{\texttt{TS}}$ and $\tautemp \leq \tau_t^{\texttt{TS}}$   in round $t$ after receiving the feedback by conducting a Bernoulli trial with success probability 
\begin{align} \label{def_pseudo_reward_TS}
\frac{( R_{I_t^{\texttt{TS}},t}   - c(C_{I_t^{\texttt{TS}},t})  ) \IND{\{C_{I_t^{\texttt{TS}},t}\leq \tautemp \} } - \lambda(\tautemp) \IND{\{C_{I_t^{\texttt{TS}},t}> \tautemp \}  }   + \lambda(\tau_{\mathrm{max}}) }{1+\lambda(\tau_{\mathrm{max}})} , 
\end{align}
incrementing  $ S_{i,\tautemp}(t)$ by 1 in case of success, and incrementing $ F_{i,\tautemp}(t)$ by $1$  otherwise.
Note that adding $\lambda(\tau_{\mathrm{max}})$ and then dividing by $1+\lambda(\tau_{\mathrm{max}})$ in \eqref{def_pseudo_reward_TS}  is equivalent to assuming that the reward sequence in \eqref{def_rewards_sequence} is in $[0,1].$
Strictly speaking, one would conduct the latter Bernoulli experiment only for deciding on the update of $ S_{I_t^{\texttt{TS}}, \tau_t^{\texttt{TS}}}(t)$ or $ F_{I_t^{\texttt{TS}}, \tau_t^{\texttt{TS}}}(t).$
However, by updating all parameters for which $\tautemp \leq \tau_t^{\texttt{TS}}$ holds, the resulting $\texttt{TS}$ algorithm performs significantly better\footnote{Note that we use the beta-Bernoulli  $\texttt{TS}$ variant \cite{agrawal2012analysis} as the scaled reward sequence \eqref{def_pseudo_reward_TS} is taking values in $[0,1]$, for which this instantiation of  $\texttt{TS}$ is known to achieve good empirical performance.}.
Note that this resembles the idea underlying the suggested penalized expected gain estimates $\hat \nu_{i,\tautemp}$ in \eqref{def_exp_gain_estimate}.

\subsection{Synthetic Data Details} \label{subsec_exp_synth_details}

\textbf{PosCorr.}
Each arm $i\in [n] $ is associated with a truncated bivariate Gaussian distribution supported on $[0,1]^2$ for $P^{(r,c)}_i$, with mean vector $\mu_i\in \mathbb R^2$ and covariance matrix $\Sigma_i \in \mathbb{R}^{2\times 2}.$   
We set $\mu_1=(0.6,0.45)^\top$ and $\mu_k=(0.5,0.5)^\top,$ $k=\{2,\ldots,10\}$, and define the covariance matrix for an arm $i$ by means of
\begin{align} \label{cov_matrix}
\Sigma_i = 	\sigma \, \left( \begin{array} {cc} 1 & 2 x_i \sqrt{1-x_i^2} \\
2 x_i \sqrt{1-x_i^2} &  1
\end{array}\right),   
\end{align}
for some $x_i \in [0,1]$ and $\sigma>0.$
Note that by defining the covariance matrices in this way, we can ensure that $\Sigma_i$  is positive-definite and also steer the correlation between the reward and resource consumption distribution of an arm $i$ effectively by the choice of $x_i.$
More specifically, we choose $\sigma=0.2,$ $x_1=0.2, x_2=0.3, x_3=0.4, x_4=0.4, x_5=\ldots=x_{10}=0.6.$
%
In this way, arm 1 with resource limit $\tau=0.4$ corresponds to the optimal arm/resource-limit pair $(i^*,\tau^*)$.

\textbf{NegCorr.}
As in the problem instance \textbf{PosCorr}, each arm $i\in [n] $ is once again associated with a truncated bivariate Gaussian distributions supported on $[0,1]^2$ for $P^{(r,c)}_i$, with mean vector $\mu_i\in \mathbb R^2$ and covariance matrix $\Sigma_i \in \mathbb{R}^{2\times 2}.$   
We set $\mu_1=(0.9,0.8)^\top$ and $\mu_k=(0.8,0.8)^\top,$ $k=\{2,\ldots,10\}$, while the covariance matrix is as in \eqref{cov_matrix} with $\sigma=0.2,$ $x_1=\ldots=x_{10}=-0.2,$ i.e., all arms have a negative correlation between the reward and resource consumption distribution.
In light of this, arm 1 with resource limit $\tau=0.4$ corresponds to the optimal arm/resource-limit pair $(i^*,\tau^*)$. 

\textbf{Indep.}
Each arm $i\in [n] $ is associated with a $\mathrm{Beta}(a_i,b_i)$ distribution for its reward distribution $P^{(r)}_i$, and an $\mathrm{Exp}(\lambda_i)$ distribution for its consumption of resources distribution $P_i^{(c)}.$ 
For any $i\in[n]$ the joint distribution $P_i^{(r,c)}$ is the product of $P^{(r)}_i$ and $P_i^{(c)},$ i.e., both distributions are independent and thus have zero correlation.
We set $(a_1,b_1)=(0.8,0.2)$ and $(a_2,b_2)= \ldots = (a_{10},b_{10}) = (0.8,0.3),$ while $\lambda_i=a_i/(a_i+b_i)+1.$
In this way, again arm 1 with resource limit $\tau=0.4$ corresponds to the optimal arm/resource-limit pair $(i^*,\tau^*)$.

\section{Further Experiments} \label{sec_exp_part_two}

In this section we provide further experiments complementing the setting considered in Section \ref{subsec_syn_exp} by varying the number of grid points and the number of arms. 
We also consider a problem instance, for which the probability of observing censored feedback is low.

\subsection{Varying Sizes of the Admissible Resource Range} \label{subsec_exp_varygrids}

We consider once again the problem instances \textbf{PosCorr}, \textbf{NegCorr} and \textbf{Indep} as described in Section \ref{subsec_exp_synth_details}.
For all problem instances we consider two variants for the admissible resource range $\mathcal{M}=(0,1],$ (i.e., $\tau_{\mathrm{max}}=1$):
\begin{itemize}
\item   $\mathcal{M}_1=\{0.5,0.9\};$ 
\item  $\mathcal{M}_2$ corresponds to an equidistant grid point set of $(0,1]$ of size $20.$ 
\end{itemize}
The choice of the cost and penalty function is the same as in  Section \ref{subsec_syn_exp}, i.e., $c(x)=x/10$ and $\lambda(x)=c(x)\IND{\{x\leq 0.5\} }  + 10x \IND{\{x> 0.5\}}. $ 
The same goes for the total number of rounds $T$ (i.e., $T=100,000$) and the number of repetitions (set to $100$).

\begin{figure*}
\centering
\subfigure
{
\includegraphics[width=0.3\linewidth]{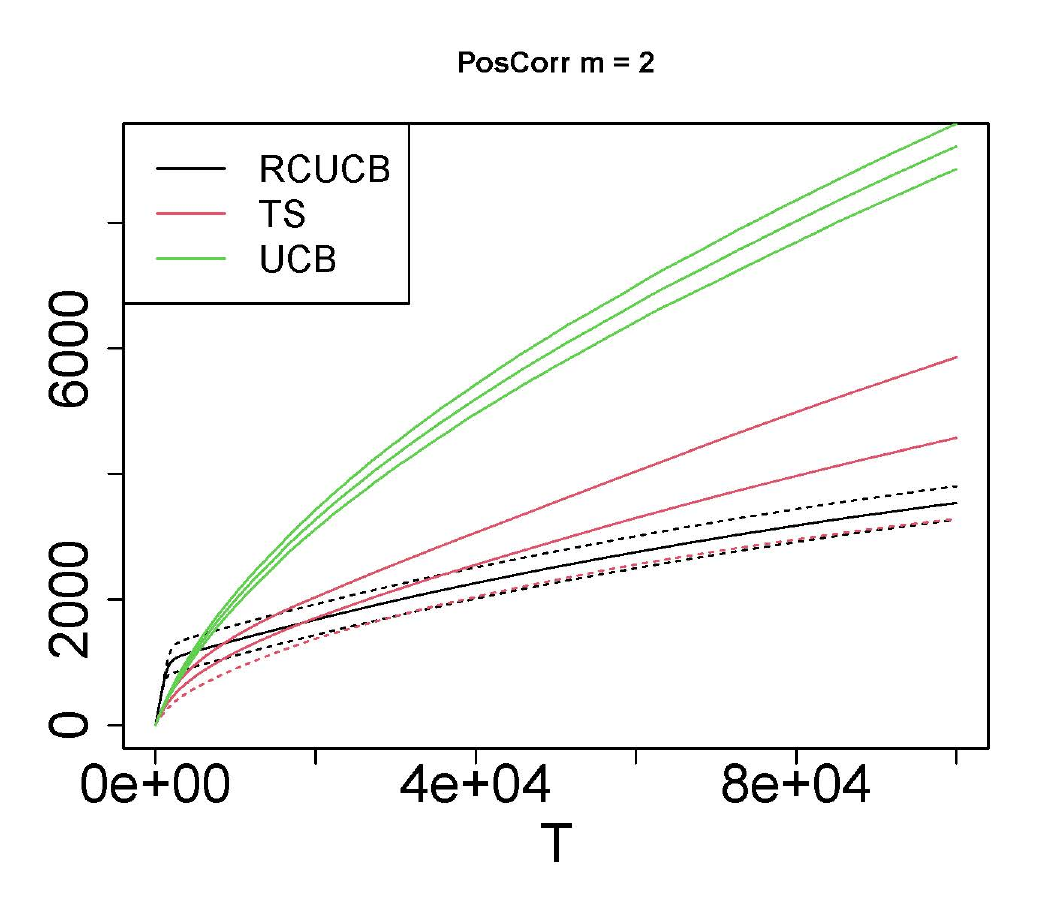}}
\subfigure
{  
\includegraphics[width=0.3\linewidth]{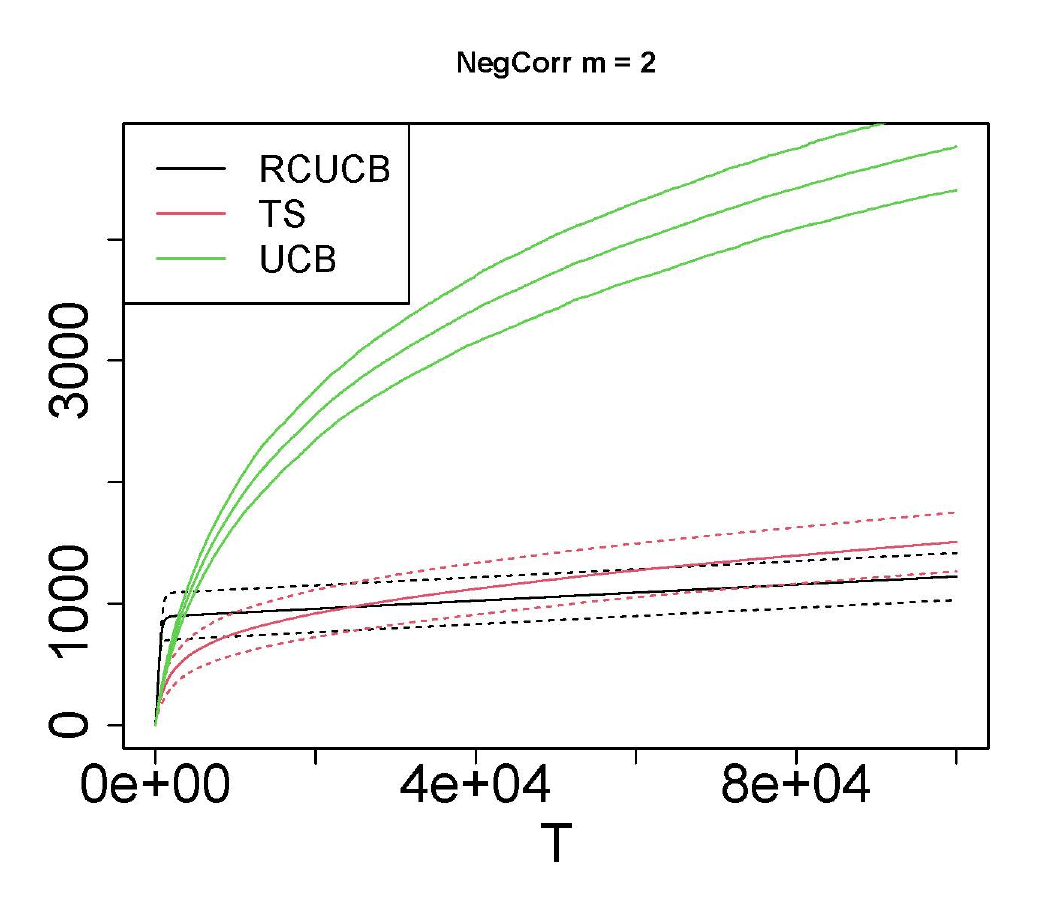}}
\subfigure
{  
\includegraphics[width=0.3\linewidth]{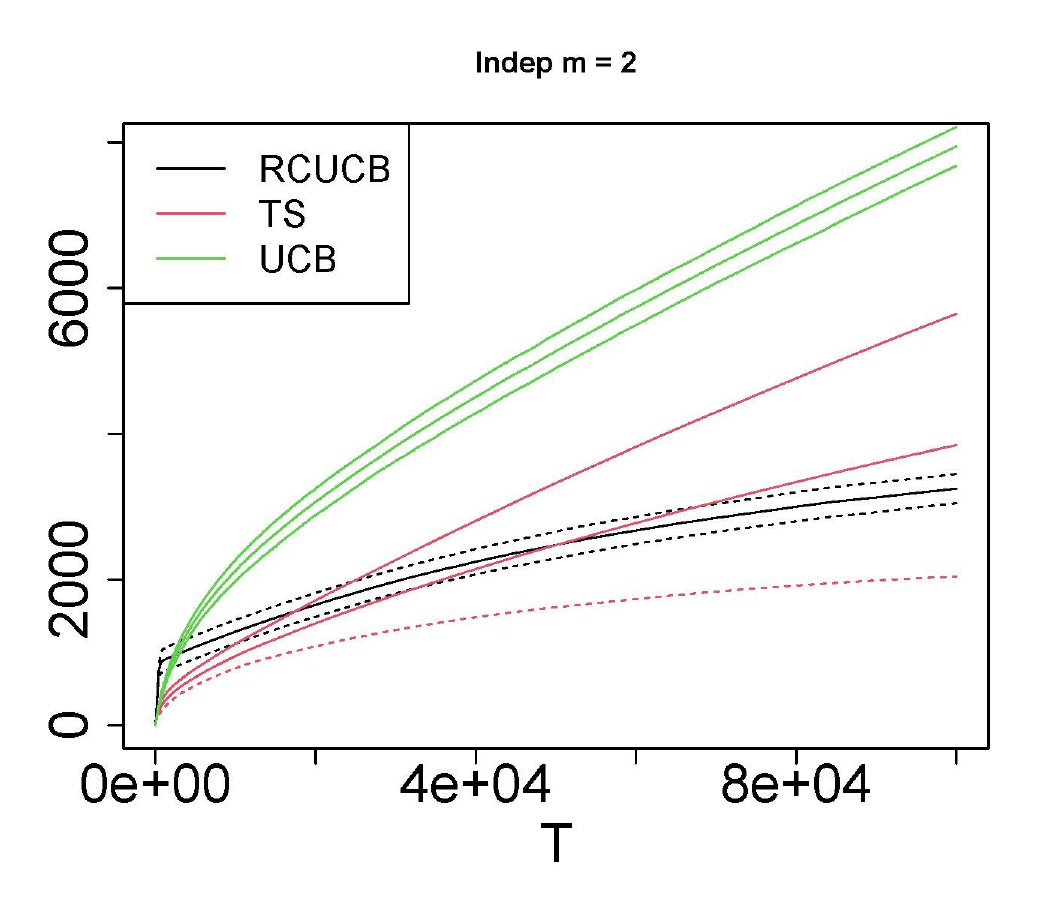}
}
\caption{Mean cumulative regret (solid lines) for UCB ($\alpha=1$), TS and RCUCB ($\alpha=1$) for the \textbf{PosCorr}, \textbf{NegCorr}, and \textbf{Indep} problem instances used with $\mathcal{M}_1$ as the admissible resource limit set. The dashed lines depict the empirical confidence intervals, using the standard error.}
\label{fig:regretanalysisproblem_two_grids}
\end{figure*}

\begin{figure*}
\centering
\subfigure
{
\includegraphics[width=0.3\linewidth]{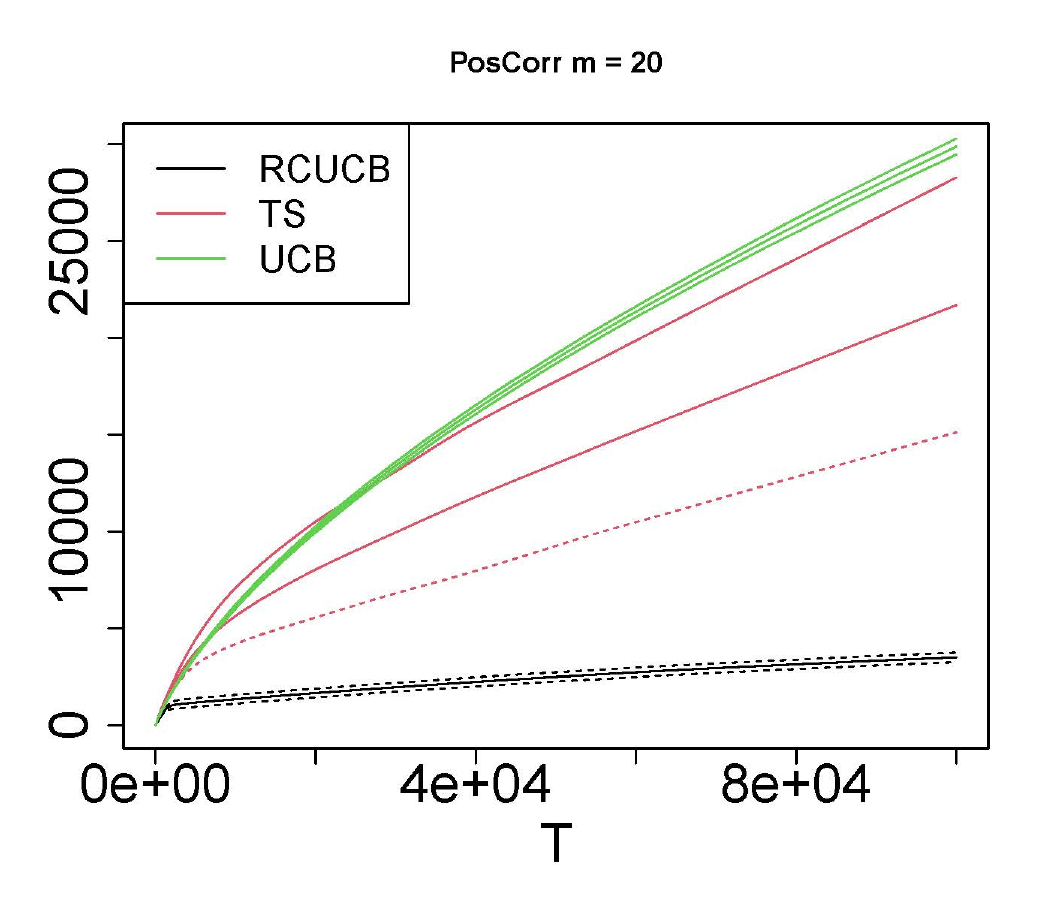}}
\subfigure
{  
\includegraphics[width=0.3\linewidth]{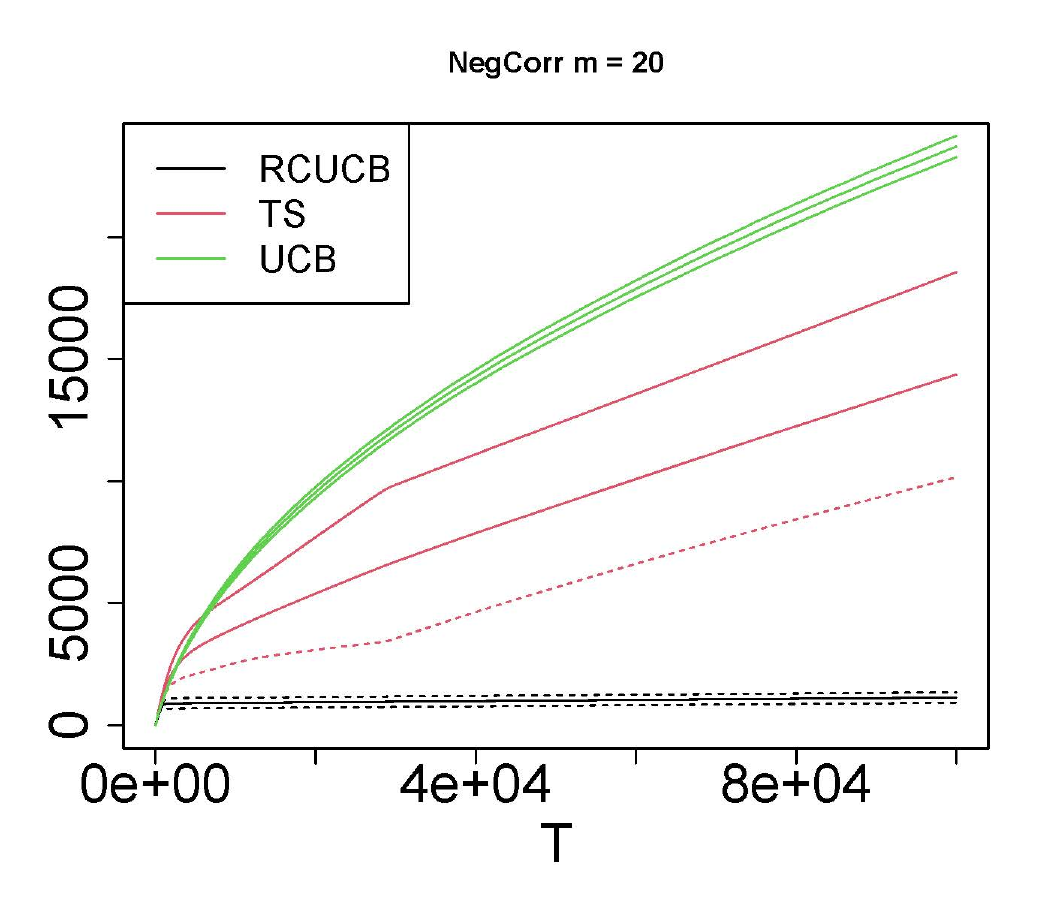}}
\subfigure
{  
\includegraphics[width=0.3\linewidth]{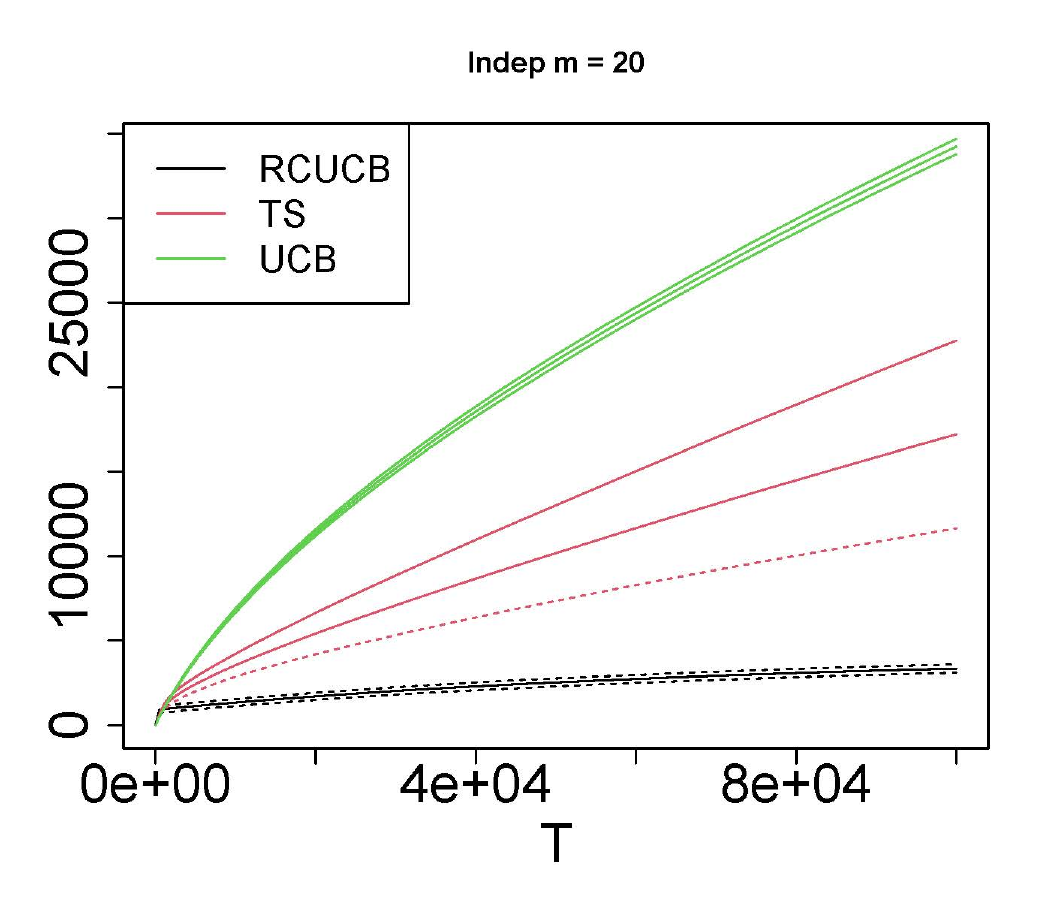}
}
\caption{Mean cumulative regret (solid lines) for UCB ($\alpha=1$), TS and RCUCB ($\alpha=1$) for the \textbf{PosCorr}, \textbf{NegCorr}, and \textbf{Indep} problem instances used with $\mathcal{M}_2$ as the admissible resource limit set. The dashed lines depict the empirical confidence intervals, using the standard error.}
\label{fig:regretanalysisproblem_twenty_grids}
\end{figure*}

The results for $\mathcal{M}_1$ are illustrated in Figure \ref{fig:regretanalysisproblem_two_grids}, while Figure \ref{fig:regretanalysisproblem_twenty_grids} shows the results for $\mathcal{M}_2.$
It is clearly visible that the superiority of $\texttt{RCUCB}$ is more distinct, the larger the number of elements in the admissible resource range.
Although  $\texttt{TS}$ performs well for $\mathcal{M}_1$ on average\footnote{Performance is to be understood with respect to the cumulative regret.}, it reveals a much higher variation than the other two algorithms.
This high variation is also visible for $\mathcal{M}_2,$ but with a much worse average performance.


The following table reports the mean proportion of censored rounds, i.e., where the resource limit was exceeded or equivalently no reward was observed, the corresponding empirical standard deviation (in squared brackets), as well as the probability of observing a censored observation for the optimal arm/limit pair $(i^*,\tau^*)$.
\begin{center}
\begin{tabular}{c|c|c|c||c}
& $\texttt{RCUCB}$ & $\texttt{TS}$ & $\texttt{UCB}$ & $1-P_{i^*}^{(c)}(\tau^*)$ \\
\hline
\multicolumn{5}{c}{$ \mathcal{M}_1$} \\
\hline
\textbf{PosCorr}  & 0.5876 [0.0029] & 0.5688 [0.0166] & 0.5919 [0.0038] & 0.5584 \\
\hline
\textbf{NegCorr}  & 0.7266 [0.0018] & 0.7081 [0.0030] & 0.7246 [0.0030] & 0.7189 \\
\hline
\textbf{Indep}    & 0.4122 [0.0017] & 0.4158 [0.0052] & 0.4142 [0.0016] & 0.4066 \\
\hline
\multicolumn{5}{c}{$\mathcal{M}_2 $} \\
\hline
\textbf{PosCorr} & 0.6126 [0.0027]  & 0.7383 [0.0935] & 0.8157 [0.0027] &  0.5851 \\
\hline
\textbf{NegCorr}  & 0.7477 [0.0016]  & 0.8396 [0.0432] & 0.9185 [0.0021] &  0.7559 \\
\hline
\textbf{Indep}    & 0.4282 [0.0017]  & 0.5639 [0.0019] & 0.5870 [0.0736] &  0.4222 \\
\end{tabular}
\end{center}
For $\mathcal{M}_1$ we see that all three algorithms have an empirical probability of obtaining censored rewards which is close to the actual ground truth probability of obtaining censored rewards for the optimal arm/limit pair.
For $\mathcal{M}_2$ the results are similar as in Section \ref{subsec_syn_exp}, that is, $\texttt{UCB}$ and $\texttt{TS}$ have a higher proportion of censored rounds than $\texttt{RCUCB},$ while the latter obtains empirically censored feedback close to the actual ground truth probability of obtaining censored rewards for the optimal arm/limit pair.

In summary, these results confirm firstly the findings in Section \ref{sec_exp}, that is,  $\texttt{RCUCB}$ is in general preferable over a straightforward application of a bandit algorithm for the considered bandit problem (i.e.,  $\texttt{UCB}$) and also over a straightforward modification of  $\texttt{TS}$ in order to incorporate the structural property of the bandit problem (see the discussion below \eqref{def_pseudo_reward_TS}).
Secondly, these results allow to conclude that $\texttt{RCUCB}$'s superiority is increasing with the size of the admissible range of resource limits.

\subsection{Larger Number of Arms}

We consider a modification of the problem instance \textbf{PosCorr} as follows.
We set 
$$\mu_i= \Big( (1 - (i-1)/n) 0.9  ), 0.3 + 0.7 (i-1)/n  \Big)^\top, \quad i=1,\ldots,n, $$   
while all covariance matrices are still of the form as in \eqref{cov_matrix}, where we set  $x_1 = \ldots = x_n = 0.2 $ and $\sigma=0.2.$
The admissible resource range $\mathcal{M},$  is defined as the equidistant grid point set of $(0,1]$ of size $5.$ 
Again the choice of the cost and penalty function is the same as in Section \ref{subsec_syn_exp} and \ref{subsec_exp_varygrids}, i.e., $c(x)=x/10$ and $\lambda(x)=c(x)\IND{\{x\leq 0.5\} }  + 10x \IND{\{x> 0.5\}}. $ 
The same goes for the total number of rounds $T$ (i.e., $T=100,000$) and the number of repetitions (set to $100$).

\begin{figure*}[htb]
\centering
\subfigure
{
\includegraphics[width=0.3\linewidth]{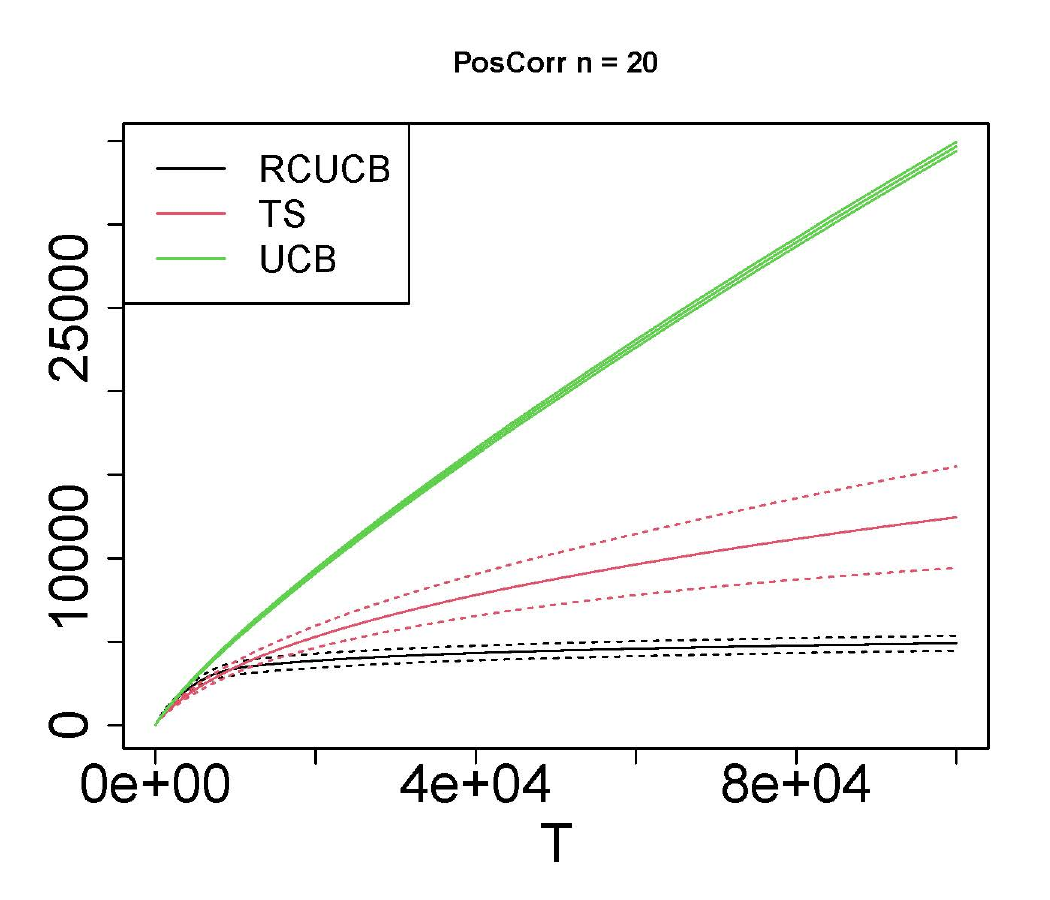}}
\subfigure
{  
\includegraphics[width=0.3\linewidth]{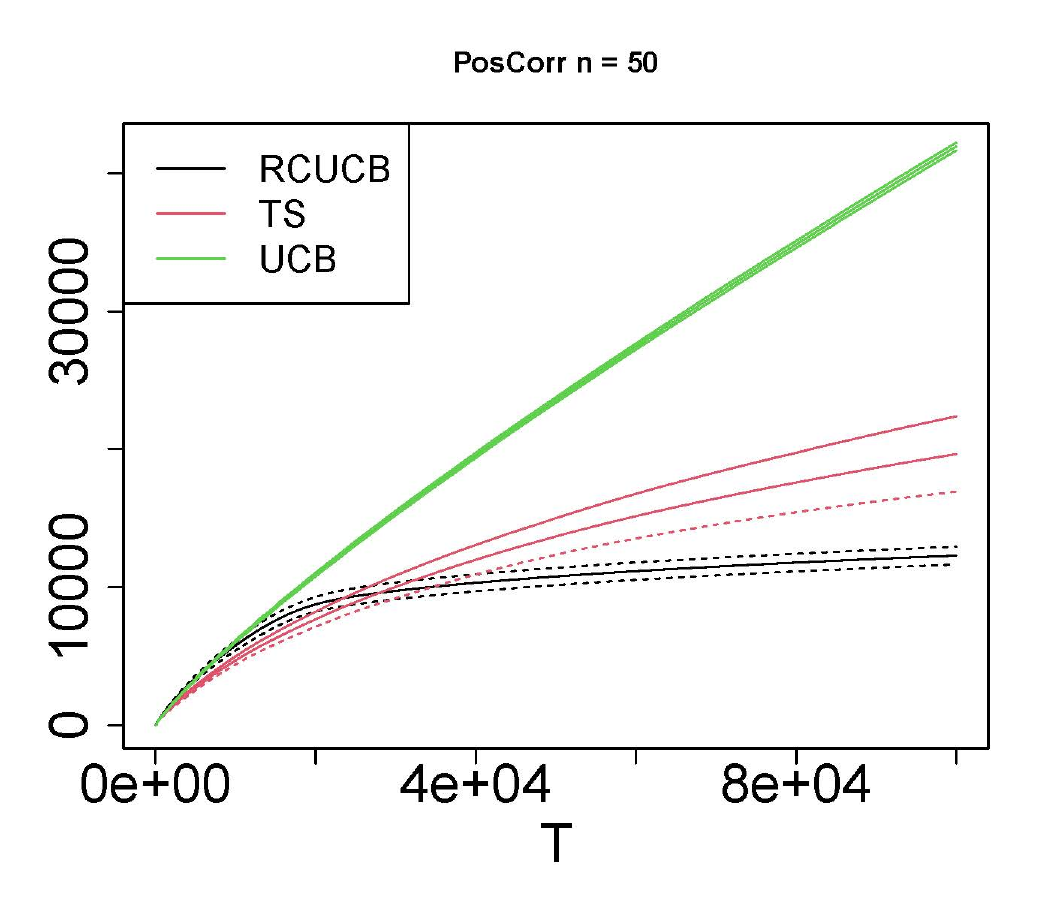}}
\subfigure
{  
\includegraphics[width=0.3\linewidth]{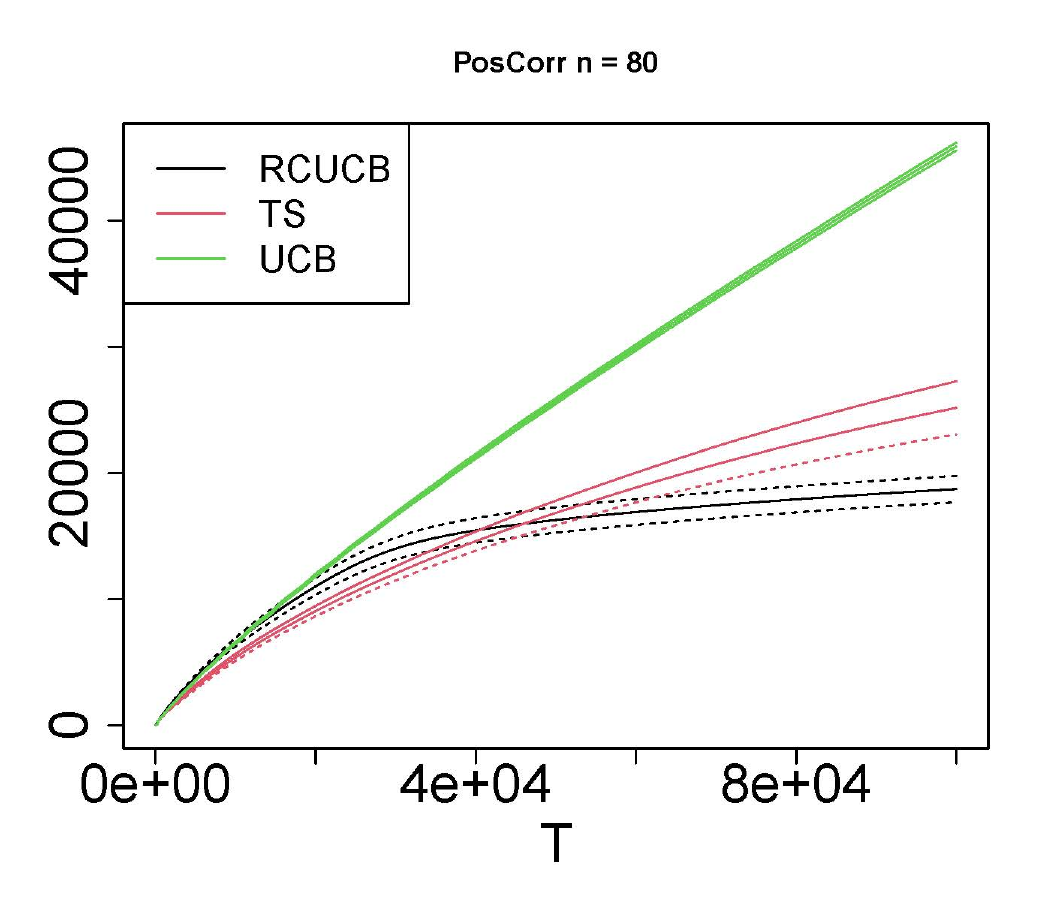}}	
\caption{Mean cumulative regret (solid lines) for UCB ($\alpha=1$), TS and RCUCB ($\alpha=1$) for the modified \textbf{PosCorr} problem instance with $n=20$ (left), $n=50$ (middle) and $n=80$ (right) arms. The dashed lines depict the empirical confidence intervals, using the standard error.}
\label{fig:regretanalysisproblem_large_arms}
\end{figure*}

The results for $n=20,$ $n=50$ and $n=80$ are illustrated in Figure \ref{fig:regretanalysisproblem_large_arms}, which reveal that $\texttt{RCUCB}$ is also well suited for the case of larger number of arms.

\subsection{Low Censoring Scenario}

Finally, we modify the problem instance \textbf{PosCorr} such that the probability of observing censored rewards is low for the optimal arm/resource-limit pair.
To this end, we set 
$$\mu_i= \Big( (1 - (i-1)/n) 0.9  ), 0 \Big)^\top, \quad i=1,\ldots,n, $$   
while all covariance matrices are still of the form as in \eqref{cov_matrix}, where we set  $x_1 = \ldots = x_n = 0.2 $ and $\sigma=0.1.$
The admissible resource range $\mathcal{M},$  is defined as the equidistant grid point set of $(0,0.6]$ of size $10.$ 
The cost function, the penalty function, the total number of rounds and the number of repetitions are the same as in Section \ref{subsec_syn_exp} and \ref{subsec_exp_varygrids}.
With this configuration, the probability of observing a censored reward for the optimal arm/resource-limit pair is about $0.075.$

\begin{figure*}[htb]
\centering
\subfigure
{
\includegraphics[width=0.42\linewidth]{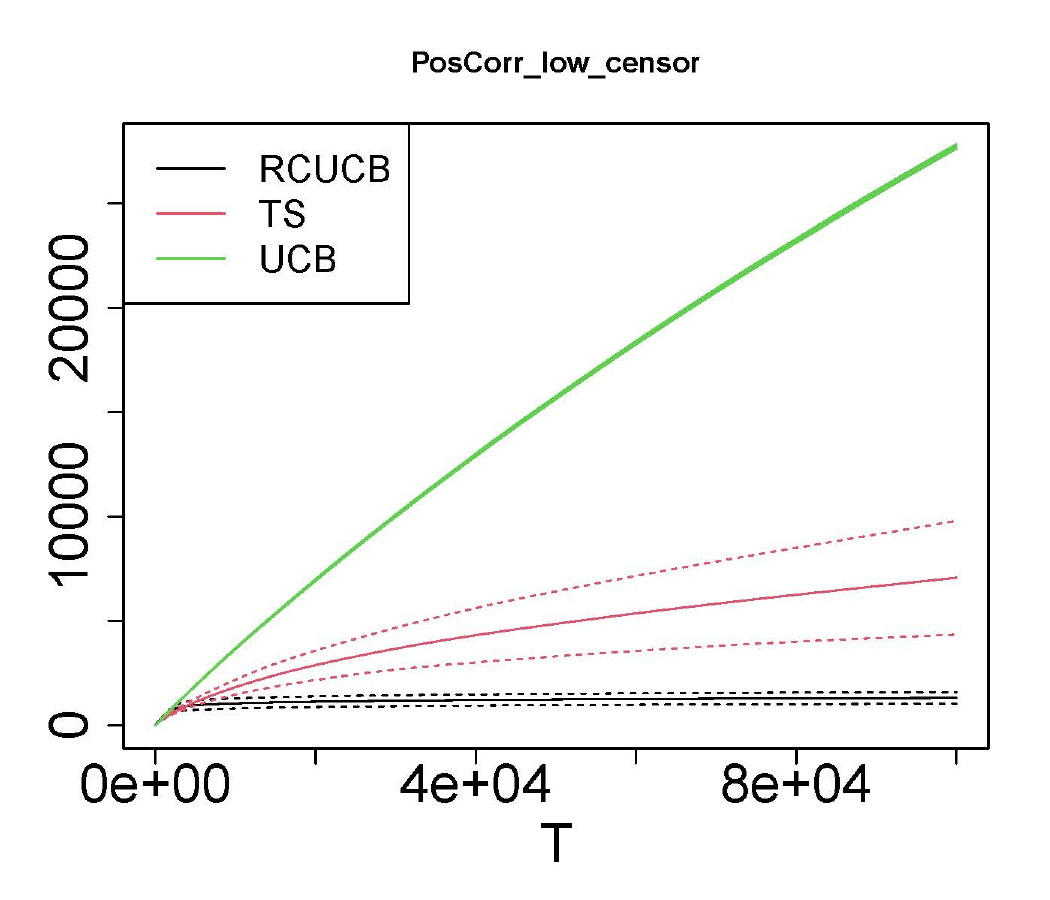}}
\caption{Mean cumulative regret (solid lines) for UCB ($\alpha=1$), TS and RCUCB ($\alpha=1$) for the modified \textbf{PosCorr} problem instance with low censoring probability for $n=5$ arms. The dashed lines depict the empirical confidence intervals, using the standard error.}
\label{fig:regretanalysisproblem_low_censor}
\end{figure*}

Figure \ref{fig:regretanalysisproblem_low_censor} illustrates the result obtained for the mean cumulative regret of the algorithms considered, if the number of arms $n$ is five.
Once again $\texttt{RCUCB}$ shows the best performance and compared to  $\texttt{TS}$ has a much smaller variation.

\section{Theoretical Guarantee of UCB} \label{sec_UCB_regret}

UCB as specified in Section \ref{subsec_baselines} satisfies the following regret upper bound.

\begin{corollary}\label{coroallary_regret_upper_bound_ucb}
Let $\alpha>1$. Then, for any number of rounds $T,$ the cumulative regret of the modified $\texttt{UCB}$ is bounded as follows:
\begin{align*}
\mathcal R_T^{\texttt{UCB}}  
\leq \sum\limits_{(i,   \tautemp) \in [n] \times \mathcal{M}: \Delta_{i,\tautemp}>0}  \frac{  2 \alpha  (1+\lambda(\tau_{\mathrm{max}}))^2 \log(T)  }{\Delta_{i,\tautemp}}
&  + \Delta_{i,\tautemp} \left(  1 + \frac{4 \left( \frac{\alpha+1}{\alpha-1} \right)^2}{\log\left(\frac{\alpha+1}{2}\right)}  \right).
\end{align*}
\end{corollary}
The proof is essentially corresponding to the first part of the proof of Theorem \ref{theorem_regret_upper_bound}.
The only difference is that $\texttt{UCB}$ revolves around the number of times a pair $(i,\tau)$ is chosen, i.e., $T_{i,\tau},$ so that the coarse regret decomposition in \eqref{regret_decomposition} is used to bound the cumulative regret instead of the refined regret decomposition in \eqref{eq_regret_decomp_refined}.
\begin{proof}
Due to the rescaling used in the definition of the estimates in \eqref{def_standard_estimate}, it holds that these estimates converge against 
$ \nu_{i,\tautemp}^{(0,1)} :=  \frac{\nu_{i,\tautemp} + \lambda(\tau_{\mathrm{max}})}{(1+\lambda(\tau_{\mathrm{max}}))}.$
However, this rescaling does not alter the ranking among the pairs $(i,\tau) \in [n] \times \mathcal{M}$ with respect to their (sub-)optimality.

The pair $(i,\tautemp) \in [n] \times \mathcal{M}$ is chosen in round $t\in\{n |\mathcal{M}| +1,\ldots,T\}$ if
\begin{align} \label{ineq_regret_upper_aux_UCB}
(i,\tautemp) \in \argmax_{ (j,   \tautemp') \in [n] \times \mathcal{M}} \tilde \nu_{j,\tautemp'}(t) + \tilde{\mathfrak{c}}_{j,\tautemp'}(t;\alpha).
%
\end{align}
Let us define the following ``bad'' events:
\begin{align*}
B_{t,1} := \{  \tilde \nu_{i,\tautemp}(t) - \tilde{\mathfrak{c}}_{i,\tautemp}(t;\alpha)  > \nu_{i,\tautemp}^{(0,1)}    \}, \quad
B_{t,2} := \{  \tilde \nu_{i^*,\tau^*}(t) + \tilde{\mathfrak{c}}_{i^*,\tau^*}(t;\alpha)  < \nu_{i^*,\tau^*}^{(0,1)}    \},
\end{align*}
where $(i^*,\tau^*)$ is the optimal pair (cf.\ discussion below \eqref{regret_def}).
Now, if $B_{t,1}^\complement$ holds, then \eqref{ineq_regret_upper_aux_UCB} implies
\begin{align*}
\nu_{i,\tautemp}^{(0,1)} + 2 \, \tilde{\mathfrak{c}}_{i,\tautemp}(t;\alpha) \geq \tilde \nu_{i^*,\tau^*}(t) + \tilde{\mathfrak{c}}_{i^*,\tau^*}(t;\alpha).
\end{align*}
If additionally $B_{t,2}^\complement$ holds, then the latter implies
\begin{align*}
\nu_{i,\tautemp}^{(0,1)}  + 2 \, \tilde{\mathfrak{c}}_{i,\tautemp}(t;\alpha) \geq  \nu_{i^*,\tau^*}^{(0,1)}.
\end{align*}
This in turn  implies $T_{i,\tautemp}(t) \leq \frac{  2 \alpha  (1+\lambda(\tau_{\mathrm{max}}))^2 \log(t)  }{ \Delta_{i,\tautemp}^2}.$
Thus, if $i\neq i^*$ and $\tautemp \neq \tau^*$ as well as $B_{t,1}^\complement \cap B_{t,2}^\complement$ holds, then it follows that
$T_{i,\tautemp}(T) \leq \frac{  2 \alpha  (1+\lambda(\tau_{\mathrm{max}}))^2 \log(T)  }{\Delta_{i,\tautemp}^2}.$
Next, by using a peeling argument and Hoeffding's maximal inequality similarly as for the proof of Proposition \ref{prop_nu_estimates} (cf.\ the proof of Theorem 2.2 in \cite{bubeck2010bandits}) we obtain
$$\mathbb{P}(	B_{t,1} \cup B_{t,2}) \leq  2\left(1+\frac{\log(t)}{\log(\frac{\alpha+1}{2})} \right) t^{-\frac{2\alpha}{\alpha+1}}.$$
With these considerations, we can infer that
\begin{align*}
\mathbb{E}&(T_{i,\tautemp}(T^+)) \\
%
%
&\leq \frac{  2 \alpha  (1+\lambda(\tau_{\mathrm{max}}))^2 \log(T)  }{\Delta_{i,\tautemp}^2} +1 + \sum_{t = \lceil  \frac{  2 \alpha  (1+\lambda(\tau_{\mathrm{max}}))^2 \log(T)  }{\Delta_{i,\tautemp}^2} \rceil}^{T} \mathbb{P} (	B_{t,1} \cup B_{t,2}) \\
&\leq \frac{  2 \alpha  (1+\lambda(\tau_{\mathrm{max}}))^2 \log(T)  }{\Delta_{i,\tautemp}^2} +1 + \frac{2}{\log(\frac{\alpha+1}{2})}  \sum_{t = 2}^{T} \left( \log\left(\frac{\alpha+1}{2}\right) +\log(t)\right) t^{-\frac{2\alpha}{\alpha+1}}\\
&\leq \frac{  2 \alpha  (1+\lambda(\tau_{\mathrm{max}}))^2 \log(T)  }{\Delta_{i,\tautemp}^2} +1 + \frac{2}{\log(\frac{\alpha+1}{2})}  \int_{1}^\infty \left( \log\left(\frac{\alpha+1}{2}\right) +\log(x)\right) x^{-\frac{2\alpha}{\alpha+1}} \, \mathrm{d}x \\
&\leq \frac{  2 \alpha  (1+\lambda(\tau_{\mathrm{max}}))^2 \log(T)  }{\Delta_{i,\tautemp}^2} + 1 + \frac{4}{\log\left(\frac{\alpha+1}{2}\right)} \left( \frac{\alpha+1}{\alpha-1} \right)^2,
%
%
\end{align*}		
where we used in the second last line that $\sum_{t=2}^{T} \log(t)/t^{c} \leq \int_{1}^{\infty} \log(x)/ x^{c} \, \mathrm{d}x$ for any $c>1,$ and for the last line that
$$  \int_{1}^{\infty} \frac{\log(x)}{x^{c}}  \, \mathrm{d}x = \frac{1}{(c-1)^2},$$
which can be seen by integration by parts.
Using the regret decomposition in \eqref{regret_decomposition} completes the proof.
\end{proof}

\end{document}